\documentclass{article}

\PassOptionsToPackage{numbers,sort&compress}{natbib}
\usepackage[preprint]{neurips_2026}

\usepackage[T1]{fontenc}
\usepackage[utf8]{inputenc}
\usepackage{microtype}
\usepackage{amsmath,amssymb,amsthm,mathtools}
\usepackage{booktabs,tabularx,array,multirow}
\usepackage{float}
\usepackage{placeins}
\usepackage{graphicx}
\usepackage{adjustbox}
\usepackage[figuresright]{rotating}
\usepackage{subcaption}
\usepackage{xcolor}
\usepackage{enumitem}
\usepackage{tikz}
\usetikzlibrary{arrows.meta, positioning}
\definecolor{observedcolor}{HTML}{DCE6F2} % muted blue for the prompt
\definecolor{suffixcolor}{HTML}{FCE4D6}   % muted orange for rollout suffixes
\usepackage[colorlinks=true,linkcolor=blue,citecolor=blue,urlcolor=blue]{hyperref}

\newtheorem{definition}{Definition}

\newcommand{\E}{\mathbb{E}}

\newcommand{\mem}{\mathrm{mem}}
\newcommand{\gen}{\mathrm{gen}}

\newcommand{\taskdiv}{M}
\newcommand{\metatrainlen}{n}
\newcommand{\promptlen}{\ell}
\newcommand{\rolllen}{N}
\newcommand{\pmcsamples}{R}

\title{What does a Bayes-filtered transformer believe? A predictive Monte Carlo approach}
\author{%
  Afiq Abdillah Effiezal Aswadi \quad Haotong Ma \quad Susan Wei \\
  Department of Econometrics and Business Statistics \\
  Monash University \\
}

\begin{document}
\raggedbottom
\maketitle

\begin{abstract}
  A \emph{Bayes-filtered transformer} (BFT) is a transformer trained on sequences that are generated in two steps: first a latent task is drawn from a prior, then observations are drawn conditional on that task. Trained under autoregressive log loss, the BFT's next-token prediction, in the idealized limit, \emph{is} the Bayesian posterior predictive distribution (PPD) induced by that prior and that conditional law.
In practice the trained BFT is only an approximation of this ideal PPD, raising an interpretive question: what prior and posterior over the latent task has the trained BFT actually internalized? Existing work answers this question by comparing the trained BFT's predictions against the predictions of various ``reference'' posteriors, each standing in for a different candidate algorithm or computation the BFT might be implementing. This prediction-space comparison is fragile: different posteriors can share the same posterior-mean predictions. We use \emph{predictive Monte Carlo} (PMC) as a general interpretability tool for any BFT: using only next-token generation, PMC returns an approximation to the implicit prior and posterior over the latent task, answering the interpretive question directly in latent space.
We apply PMC to three stylized task families spanning 0-Markov and 1-Markov exchangeability. The phenomena previously reported in these settings remain visible in latent space.
Code is available at \url{https://github.com/afiq-aswadi/bft-pmc}.

\end{abstract}

\section{Introduction}

A \emph{Bayes-filtered transformer} (BFT) is one pretrained on sequences from a
hierarchical pretraining distribution: a latent task is sampled from a prior, and
observations are generated conditional on the task. Autoregressive log-loss
training under such a distribution, also known as \emph{Bayes-filtered} meta-learning
\citep{Ortega2019}, is minimized by the Bayesian posterior predictive
distribution (PPD). In the idealized limit of sufficient capacity, infinite
data, and perfect optimization, a BFT \emph{is} this PPD. In practice these
conditions fail, and the trained BFT realizes only some approximation of the
ideal object.

An interpretive question follows: given a trained BFT, what prior and
posterior belief over the latent task has it actually internalized?
\emph{Predictive Monte Carlo} (PMC) \citep{Fortini2020}, a technique from
Bayesian predictive inference (BPI) \citep{Fortini2025}, answers this question using only
next-token generation. It samples from the BFT's implicit prior over the latent
task by rolling out from an empty prompt, and from its implicit posterior by
rolling out from a given prompt. In each case, the draw of the latent task is
computed from the completed rollout, for instance as an empirical frequency or
a regression fit.

\begin{figure}[!htbp]
  \centering
  \begin{subfigure}[c]{0.55\linewidth}
    \centering
    \raisebox{15pt}{%
      \resizebox{\linewidth}{!}{%
        \begin{tikzpicture}[
            rollout/.style={font=},
            shared/.style={text=black, fill=observedcolor, rounded corners, inner sep=3pt, minimum width=0.9cm, align=center},
            suffix/.style={text=black, fill=suffixcolor, rounded corners, inner sep=3pt, minimum width=0.9cm, align=center},
            arrow/.style={->, thick, >=Stealth},
            node distance=0.6cm and 0.3cm
          ]

          % ==============================
          % ROW 1
          % ==============================
          \node[font=\sffamily] (r1label) at (0, 0) {Rollout 1};
          \node[rollout, shared, right=0.3cm of r1label] (y1_1) {$y_1$};
          \node[rollout, shared, right=0.1cm of y1_1] (y1_dots) {$\dots$};
          \node[rollout, shared, right=0.1cm of y1_dots] (y1_n) {$y_{\promptlen}$};
          \node[rollout, suffix, right=0.5cm of y1_n] (s1_1) {$y_{\promptlen+1}^{(1)}$};
          \node[rollout, suffix, right=0.1cm of s1_1] (s1_dots) {$\dots$};
          \node[rollout, suffix, right=0.1cm of s1_dots] (s1_N) {$y_{\promptlen+N}^{(1)}$};
          \node[font=\footnotesize, right=0.5cm of s1_N] (theta1) {$\hat\theta^{(1)}$};
          \draw[arrow] (s1_N) -- (theta1);

          % HEADERS
          \node[align=center, font=\scriptsize\sffamily, above=0.2cm of y1_dots] {Prompt};
          \node[align=center, font=\scriptsize\sffamily, above=0.2cm of s1_dots] {Sample via Beta-Bernoulli BFT};

          % ==============================
          % ROW 2
          % ==============================
          \node[font=\sffamily, below=0.8cm of r1label] (r2label) {Rollout 2};
          \node[rollout, shared, right=0.3cm of r2label] (y2_1) {$y_1$};
          \node[rollout, shared, right=0.1cm of y2_1] (y2_dots) {$\dots$};
          \node[rollout, shared, right=0.1cm of y2_dots] (y2_n) {$y_{\promptlen}$};
          \node[rollout, suffix, right=0.5cm of y2_n] (s2_1) {$y_{\promptlen+1}^{(2)}$};
          \node[rollout, suffix, right=0.1cm of s2_1] (s2_dots) {$\dots$};
          \node[rollout, suffix, right=0.1cm of s2_dots] (s2_N) {$y_{\promptlen+N}^{(2)}$};
          \node[font=\footnotesize, right=0.5cm of s2_N] (theta2) {$\hat\theta^{(2)}$};
          \draw[arrow] (s2_N) -- (theta2);

          % ==============================
          % DOTS & ROW L
          % ==============================
          \node[below=0.2cm of r2label, font=\large] (vdots_label) {$\vdots$};
          \node[below=0.2cm of y2_dots, font=\large] {$\vdots$};
          \node[below=0.2cm of s2_dots, font=\large] {$\vdots$};
          \node[below=0.2cm of theta2, font=\large] {$\vdots$};

          \node[font=\sffamily, below=0.8cm of vdots_label] (rLlabel) {Rollout $R$};
          \node[rollout, shared, right=0.3cm of rLlabel] (yL_1) {$y_1$};
          \node[rollout, shared, right=0.1cm of yL_1] (yL_dots) {$\dots$};
          \node[rollout, shared, right=0.1cm of yL_dots] (yL_n) {$y_{\promptlen}$};
          \node[rollout, suffix, right=0.5cm of yL_n] (sL_1) {$y_{\promptlen+1}^{(R)}$};
          \node[rollout, suffix, right=0.1cm of sL_1] (sL_dots) {$\dots$};
          \node[rollout, suffix, right=0.1cm of sL_dots] (sL_N) {$y_{\promptlen+N}^{(R)}$};
          \node[font=\footnotesize, right=0.5cm of sL_N] (thetaL) {$\hat\theta^{(R)}$};
          \draw[arrow] (sL_N) -- (thetaL);

        \end{tikzpicture}%
      }%
    }%
    \caption{PMC procedure requires only forward passes.}
    \label{fig:pmc-schematic-rollouts}
  \end{subfigure}
  \hfill
  \begin{subfigure}[c]{0.42\linewidth}
    \centering
    \includegraphics[width=\linewidth]{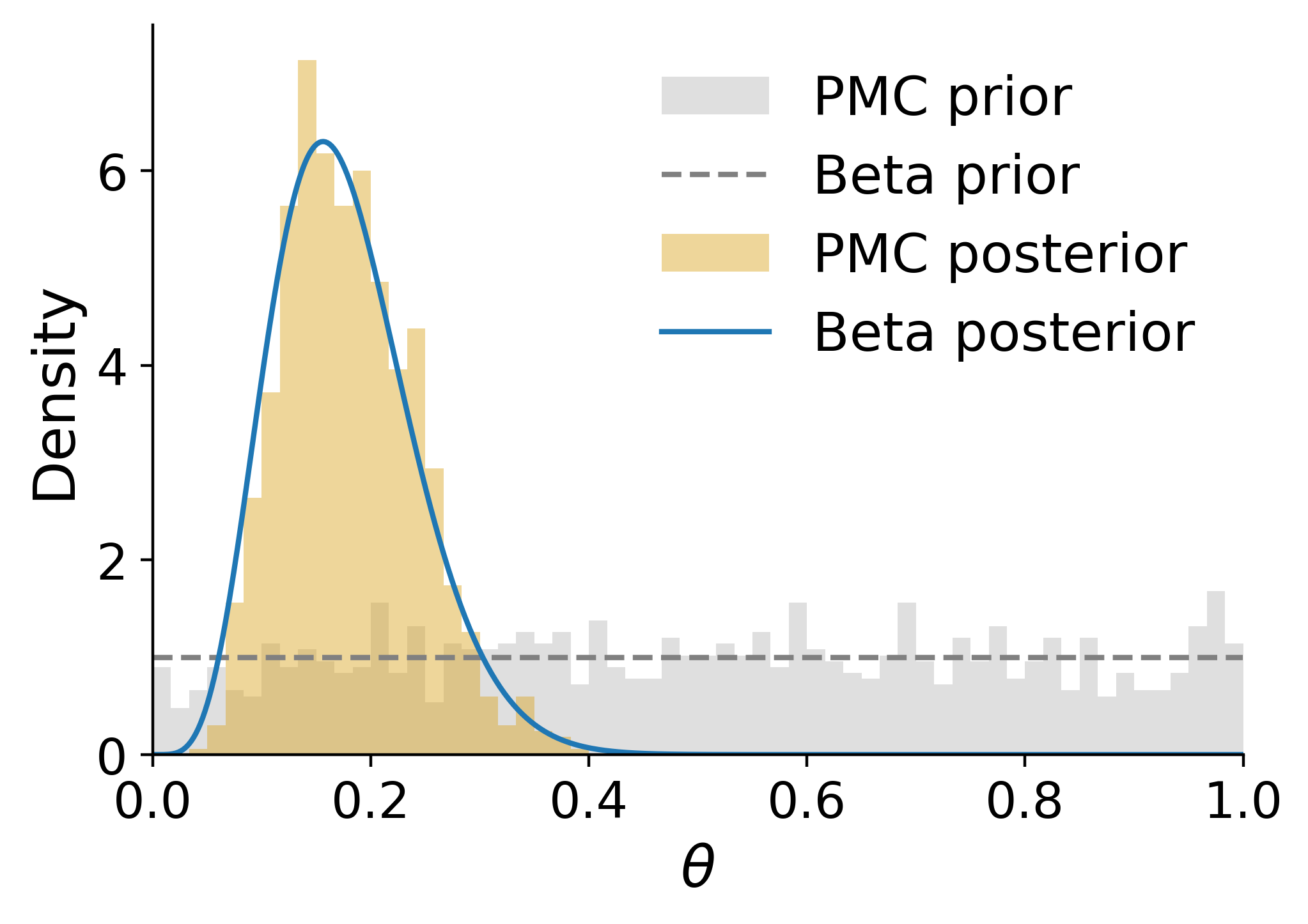}
    \caption{PMC prior and posterior samples vs the analytic
      $\operatorname{Beta}$ prior and conjugate posterior.}
    \label{fig:pmc-schematic-hist}
  \end{subfigure}
  \caption{\textbf{PMC schematic.} Given a (possibly empty) prompt $y_{1:\promptlen}$, the Beta-Bernoulli BFT autoregressively generates $R$ rollouts $y_{\promptlen+1:\promptlen+N}^{(r)}$ of length $N$ (panel~a). For each rollout $r$, $\hat\theta^{(r)}=\frac{1}{\promptlen+N}\bigl(\sum_{i=1}^{\promptlen} y_i + \sum_{i=\promptlen+1}^{\promptlen+N} y_i^{(r)}\bigr)$ is the sample mean of the full rollout (including the initial prompt). Panel~(b) shows one histogram of $\{\hat\theta^{(r)}\}_{r=1}^R$ obtained with an empty prompt and one obtained with a non-empty prompt. With the \emph{empty} prompt, the $\hat\theta^{(r)}$ are samples from the BFT's implicit \emph{prior} over $\theta$, which tracks the analytic $\operatorname{Beta}(1,1)$ prior used during pretraining. With the \emph{non-empty} prompt (here $\promptlen=32$ with 5 ones and 27 zeros), they are samples from the BFT's implicit \emph{posterior}, which tracks the analytic conjugate posterior $\operatorname{Beta}(6,28)$. Panel~(a) is adapted from \citet{Ng2026}. Training and rollout settings for this model are in Appendix~\ref{app:beta-bernoulli}.}
  \label{fig:pmc-schematic}
\end{figure}
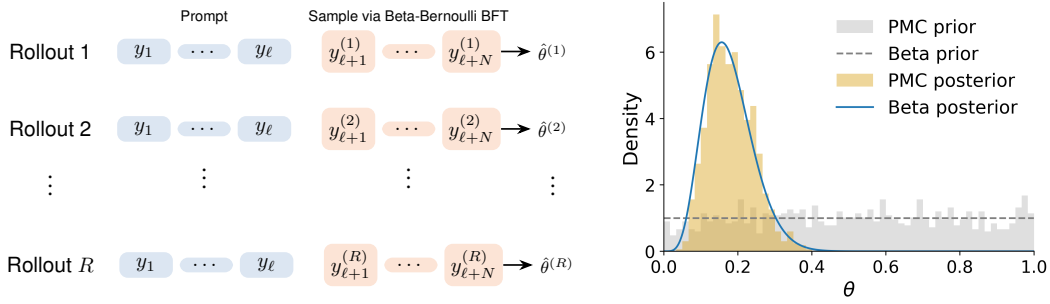

As a first illustration, consider a BFT pretrained on binary sequences
generated from a Bernoulli likelihood with a uniform $\operatorname{Beta}(1,1)$
prior on the success probability $\theta$, a \emph{Beta-Bernoulli BFT}.
Figure~\ref{fig:pmc-schematic} shows PMC applied to such a model. Given a
prompt $y_{1:\promptlen}$, the PMC procedure (panel~a) draws $R$ rollouts from
the Beta-Bernoulli BFT and, for each rollout $r$, computes the sample mean
$\hat\theta^{(r)}$ of the completed sequence. The empirical distribution of
$\{\hat\theta^{(r)}\}_{r=1}^R$ is the BFT's PMC-recovered posterior
over $\theta$. Running the same procedure with an empty prompt recovers its
implicit prior. Panel~(b) compares both against their analytic counterparts:
the PMC prior samples against the $\operatorname{Beta}(1,1)$ prior used during
pretraining, and the PMC posterior samples against the conjugate
$\operatorname{Beta}$ posterior. Despite being only an
approximation to the Beta-Bernoulli PPD, the trained BFT's implicit
prior and posterior over $\theta$ both track their analytic counterparts
closely. Simultaneous agreement of prior and posterior in latent space
indicates that the model both encodes the pretraining prior and updates it via
Bayes' rule. A prediction-space comparison can establish neither, since
distinct priors and posteriors can produce identical predictions.

The class of BFTs includes practically important foundation models.
Prior-data fitted networks (PFNs) \citep{Muller2024} are BFTs in the
exchangeable setting, and underpin TabPFN \citep{Hollmann2023,Hollmann2025},
which is currently state of the art on real-world small tabular classification
and regression, purely via \textbf{in-context learning} (ICL) at inference
time.

A parallel line of work uses BFTs as a controlled testbed for studying ICL in
transformers: by adopting pretraining distributions whose posteriors admit
closed-form expressions, the transformer's behavior can be compared directly
against a known Bayesian reference. \citet{Raventos2023} trained transformers
on in-context linear regression, \citet{Carroll2025} studied the training
dynamics in that setting, \citet{NEURIPS2025_abb22b1e} extended this line to
additional exchangeable data-generating mechanisms including balls-and-urns,
and \citet{edelman2024the,Park2025} pretrained on mixtures of Markov chains.
\citet{Raventos2023,Carroll2025,NEURIPS2025_abb22b1e,Park2025} fix two
reference priors: a continuous \textbf{generalizing} prior $\Pi_\infty$, and an
empirical \textbf{memorizing} prior $\Pi_{\taskdiv}$ consisting of a finite set
of pretraining tasks drawn from the generalizing prior. The transformer is then
characterized by which of the two induced posteriors more closely describes it.
These three settings, \emph{balls-and-urns}, \emph{linear regression}, and
\emph{Markov chains} (Table~\ref{tab:families}), are the task families of this
paper. Two empirical phenomena have emerged from this line of work:
\begin{itemize}[leftmargin=1.8em, itemsep=2pt, topsep=2pt]
  \item \textbf{Task-diversity threshold} \citep{Raventos2023}: as the number of pretraining tasks grows, the transformer shifts from the memorizing scheme to the generalizing scheme, despite the memorizing scheme being optimal under the pretraining distribution.
  \item \textbf{Transient generalization} \citep{Carroll2025}: at intermediate task diversity, the transformer first approximates the generalizing scheme during training before later specializing to the memorizing scheme.
\end{itemize}

\begin{figure}[!htbp]
  \centering
  \includegraphics[width=\linewidth]{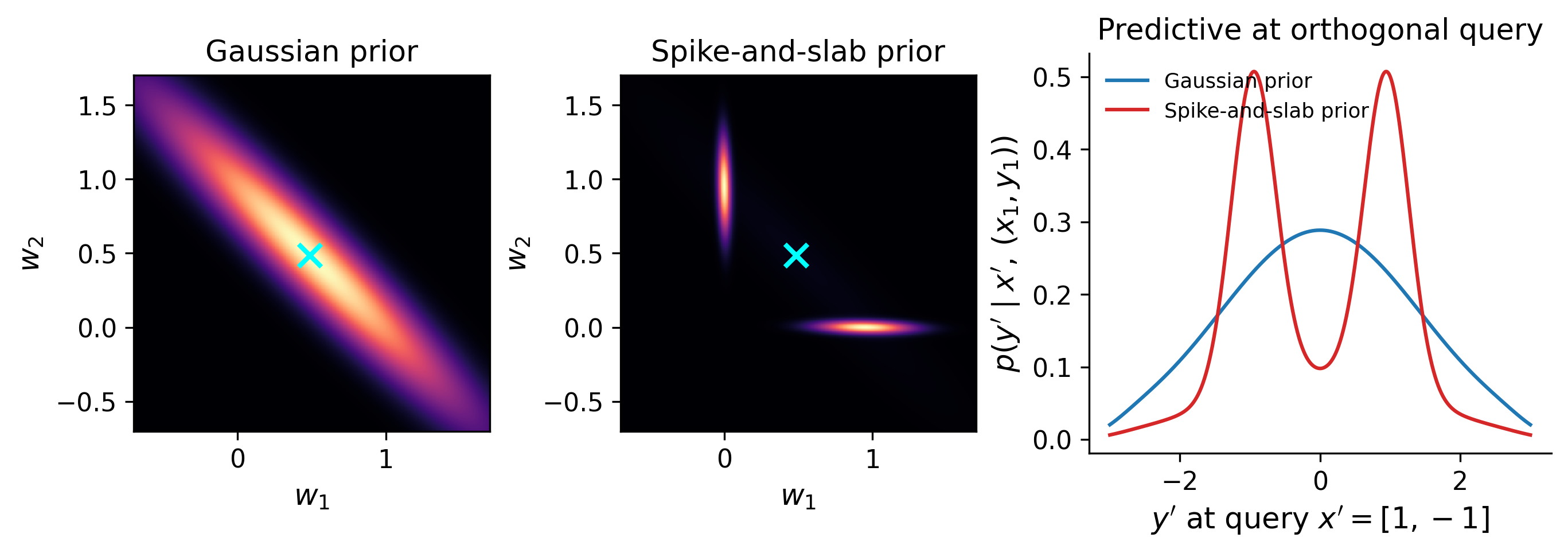}
  \caption{\textbf{Prediction-space measurement is fragile.} 2-D Bayesian linear regression: posterior over the coefficient $w \in \mathbb{R}^2$ given a single observation $(x_1,y_1)=([1,1],1)$, computed under two different priors. The Gaussian and spike-and-slab priors yield posteriors that share the mean $(0.5, 0.5)$ (marked $\times$) but differ in shape (left, middle). At the test query $x'=[1,-1]$, the two predictive distributions over $y'$ are unimodal vs bimodal (right), yet the posterior-mean predictions coincide. A comparison of posterior means therefore cannot distinguish the two posteriors.}
  \label{fig:intro-identifiability}
\end{figure}

The papers above measure the transformer in prediction space: they compare its
predictions against those of each scheme and read off which scheme the
transformer tracks, either at the end of training as task diversity $\taskdiv$
varies, or throughout training at a fixed $\taskdiv$. The discrepancy measure
varies with the setting, from MSE between point predictions
\citep{Raventos2023,Carroll2025} to KL-type divergences between next-token
predictives \citep{NEURIPS2025_abb22b1e,edelman2024the,Park2025}. The two
schemes, however, differ in their posteriors over the latent task, and distinct
posteriors can produce identical predictions, so prediction-space comparisons
cannot always tell the schemes apart. Figure~\ref{fig:intro-identifiability}
gives a concrete example for MSE: two priors with the same posterior mean give
identical point predictions even when their posterior shapes are markedly
different. Divergences between next-token predictives are more discriminating
but still a projection: in the categorical families, the predictive at each
position is determined by the posterior mean of the latent parameter, so
distinct posteriors with the same posterior-mean trajectory remain
indistinguishable in prediction space.

\noindent\textbf{Contributions.} \textbf{(i)} We unify previously disparate synthetic ICL studies (balls-and-urns \citep{NEURIPS2025_abb22b1e}, linear regression \citep{Raventos2023,Carroll2025}, and Markov chains \citep{edelman2024the,Park2025}) as \emph{$k$-Markov exchangeable} BFTs. \textbf{(ii)} We move ICL evaluation from prediction space into latent space, repurposing PMC as a black-box interpretability tool for $k$-Markov exchangeable BFTs that uses only next-token generation.
\textbf{(iii)} To our knowledge, this is the first application of PMC to BFTs.
\textbf{(iv)} On the three families, we reproduce two previously reported ICL phenomena (the task-diversity threshold and transient generalization) directly in latent space.

\noindent\textbf{Outline.} Section~\ref{sec:in-context-kme} defines the exchangeability properties ($0$-Markov and $1$-Markov) shared by our task families. Section~\ref{sec:framework} defines meta-learning on Bayes-filtered data, whose idealized limit is the Bayesian PPD that a trained BFT approximates. Section~\ref{sec:pmc} presents PMC and explains how it recovers the prior and posterior the trained BFT has actually internalized.
Section~\ref{sec:experiments} defines the memorizing and generalizing baselines and the prediction-space and latent-space metrics, and applies PMC across balls-and-urns, linear regression, and Markov chains, where the task-diversity threshold and transient generalization persist in latent space.
Section~\ref{sec:related} situates the paper in the ICL phenomenology and Bayesian predictive inference literatures.

\section{Markov exchangeability}
\label{sec:in-context-kme}

Recent controlled ICL studies pretrain transformers on sequences drawn from
specific pretraining distributions: balls-and-urns
\citep{NEURIPS2025_abb22b1e}, linear regression tasks
\citep{Raventos2023,Carroll2025}, and finite-state Markov chains
\citep{edelman2024the,Park2025}.
Table~\ref{tab:families} summarizes these three task families side by side.
They all sit inside the broader class of $k$-Markov exchangeable processes. We
treat $k=0$ (exchangeable, Section~\ref{sec:exch}) and $k=1$ (Markov
exchangeable, Section~\ref{sec:1markov-exch}). The general case is in
Appendix~\ref{app:kmarkov}.

Throughout, $\mathcal Y$ denotes a finite token alphabet. The continuous
response in linear regression is discretized
(Appendix~\ref{app:linear-baselines}). We refer to Appendix~\ref{app:notation}
for a full list of notation used. Write $\mathcal Y^m$ for the set of length-$m$ sequences over $\mathcal Y$, 
with $\mathcal Y^0:=\{\varnothing\}$, and $\mathcal Y^* := \bigcup_{m\ge 0}
\mathcal Y^m$ for the set of finite sequences. We use $\mathcal Y^\infty$ to denote
the set of infinite sequences. We work with a stochastic
process $(Y_n)_{n\ge 1}$ on $\mathcal Y$ with joint law $\mathbb P$, and with
the associated sequence of one-step predictive distributions
\begin{equation}
P_0(A) := \mathbb P(Y_1\in A),\qquad P_n(A) := \mathbb P(Y_{n+1}\in A\mid Y_{1:n}),\qquad A\subseteq\mathcal Y,\ n\ge 1.
\label{eq:predictive-def}
\end{equation}

The controlled ICL setups we consider all draw sequences from a pretraining distribution of the same form: a
finite-dimensional random \textbf{latent task} $\tilde\theta\in\Theta$ is
drawn from a prior $\Pi$, and tokens are then generated sequentially via a
probability kernel $P_{\tilde\theta}$ from $\mathcal Y^*$ to $\mathcal Y$,
meaning that after history $h\in\mathcal Y^*$ the next token is drawn from
$P_{\tilde\theta}(\cdot\mid h)$. The history argument lets the next token
depend on earlier tokens, as in the Markov chains family. The resulting joint
law is
\begin{equation}
\mathbb P(Y_{1:n}=y_{1:n}) = \mathbb E_{\tilde\theta\sim \Pi}\left[\prod_{i=1}^n P_{\tilde\theta}(y_i\mid y_{1:i-1})\right].
\label{eq:icb-joint}
\end{equation}
The corresponding one-step predictive, written in terms of the random prefix
$Y_{1:n}$ rather than a fixed value $y_{1:n}$, is
\begin{equation}
P_n(A) = \mathbb E_{\tilde\theta\sim \Pi(\cdot\mid Y_{1:n})}\bigl[P_{\tilde\theta}(A\mid Y_{1:n})\bigr], \qquad A\subseteq\mathcal Y,
\label{eq:icb-general}
\end{equation}
where $\Pi(\cdot\mid Y_{1:n})$ is the posterior on $\Theta$ induced by $\Pi$ and the observations. The form of $\tilde\theta$ and $P_{\tilde\theta}$ in each family is summarized in Table~\ref{tab:families}.

\subsection{Exchangeability ($0$-Markov exchangeability)}\label{sec:exch}
A process $(Y_n)_{n\ge 1}$ on $\mathcal Y$ with joint law $\mathbb P$ is
\emph{exchangeable} if
$(Y_1,\dots,Y_n)\stackrel{d}{=}(Y_{\sigma(1)},\dots,Y_{\sigma(n)})$ for every
$n$ and every permutation $\sigma$ of $\{1,\dots,n\}$. Because the quantifier
runs over every $n$, this is exchangeability of the infinite process, which is
what the representation theorem below requires. By \textbf{de Finetti's
  representation theorem} \citep{deFinetti1937,Fortini2025}, exchangeability is equivalent to
the existence of a random probability measure $\tilde P$ on $\mathcal Y$ such
that conditional on $\tilde P$ the tokens $Y_1,Y_2,\dots$ are i.i.d.\ with
distribution $\tilde P$. We call $\tilde P$ the \textbf{directing random
measure}.

For balls-and-urns, the latent task is a probability vector $\tilde
p\in\Delta^{|\mathcal Y|-1}$, the probability simplex on $\mathcal Y$, with prior $\Pi(\tilde
p)=\operatorname{Dirichlet}(1,\dots,1)$, and the directing random measure is
$\tilde P=\operatorname{Categorical}(\tilde p)$: conditional on $\tilde p$,
the tokens are i.i.d.\ $\operatorname{Categorical}(\tilde p)$.

For linear regression, the exchangeable observations are pairs
$Z_i=(X_i,Y_i)$. The latent task is the regression vector $\tilde w\in\mathbb
R^d$, with prior $\Pi(\tilde w)=\mathcal N(0,I_d)$. Conditional on $\tilde w$,
the pairs are i.i.d.\ with $X_i \sim \mathcal N(0, I_d)$ and $Y_i \mid X_i \sim
P_{\tilde w}(\cdot\mid X_i)$, where $P_{\tilde w}(\cdot\mid x)$ is the
discretization of $\mathcal N(\tilde w^\top x,\sigma^2)$ to $\mathcal Y$
(Appendix~\ref{app:linear-baselines}).

\subsection{$1$-Markov exchangeability}\label{sec:1markov-exch}
This subsection treats the $1$-Markov case. The general $k$-Markov case
($k\ge 2$) is recorded in Appendix~\ref{app:kmarkov}. For an exchangeable
process, the law of a path depends only on how many times each token value
appears in it. The Markov analog lets the law depend on the path's initial
state and its transition counts.
Formally, a process
$(Y_n)_{n\ge 1}$ on $\mathcal Y$ with joint law $\mathbb P$ is \emph{1-Markov
exchangeable} \citep{Fortini2025} if for every $n$ and any two paths
$y_{1:n},y'_{1:n}\in\mathcal Y^n$,
\[
  y_1=y'_1
  \ \text{and}\
  N_{a,b}(y_{1:n})=N_{a,b}(y'_{1:n})\ \forall a,b\in\mathcal Y
  \ \Longrightarrow\
  \mathbb P(Y_{1:n}=y_{1:n})=\mathbb P(Y_{1:n}=y'_{1:n}),
\]
where $N_{a,b}(y_{1:n}) := \sum_{i=1}^{n-1}\mathbf 1\{y_i=a,\ y_{i+1}=b\}$ are
the transition counts.

Let $\mathcal Q := \{Q=(Q_{a,b})_{a,b\in\mathcal Y}:\ Q_{a,b}\in[0,1],\
  \sum_{b\in\mathcal Y}Q_{a,b}=1\ \forall a\in\mathcal Y\}$ denote the set of
transition matrices on $\mathcal Y$. Under a recurrence condition (every state
in $\mathcal Y$ is visited infinitely often $\mathbb P$-almost surely; see
\citet{Fortini2025} for a precise statement), the \textbf{Diaconis--Freedman
  representation theorem} \citep{DiaconisFreedman1980} says that Markov exchangeability is equivalent to the
existence of a \textbf{random transition matrix} $\tilde Q\in\mathcal Q$, with
prior distribution $\Pi$ on $\mathcal Q$, such that conditional on $\tilde Q$,
$(Y_n)$ is a Markov chain governed by the same transition matrix $\tilde Q$ at
every step, i.e.\ a time-homogeneous chain.

For a transition matrix $Q\in\mathcal Q$, the entry $Q_{a,b}$ is the
probability of moving from state $a$ to state $b$, and $Q_a(A):=\sum_{b\in
A}Q_{a,b}$ is the probability that the next state lies in $A$ given current
state $a$. The one-step predictive is the PPD $P_n(A) = \mathbb E_{\tilde
Q\sim \Pi(\cdot\mid Y_{1:n})}[\tilde Q_{Y_n}(A)]$ for $A\subseteq\mathcal Y$,
where the posterior on $\tilde Q$ is $\Pi(\mathrm dQ\mid Y_{1:n})
\propto \prod_{i=1}^{n-1}Q_{Y_i,Y_{i+1}}\,\Pi(\mathrm dQ)$, with
$Y_1$ taken as given so that $P(Y_1\mid\tilde Q)$ does not appear in the
likelihood. The Markov-exchangeable case is the special case of
\eqref{eq:icb-general} with latent task $\tilde Q$ and kernel $P_{\tilde
Q}(y\mid y_{1:n})=\tilde Q_{y_n,y}$: the kernel depends on the history only
through its last token. Our Markov chains family \citep{edelman2024the,Park2025}
is one instance of this class: under our prior $\Pi$, the rows of $\tilde Q$
are independent, and each row is Dirichlet-distributed,
$(\tilde Q_{a,b})_{b\in\mathcal Y}\sim\operatorname{Dirichlet}(1,\dots,1)$ for
each $a\in\mathcal Y$. Any other prior on $\mathcal Q$ would define a different
$1$-Markov exchangeable family.

\begin{table}[!htbp]
  \centering
  \caption{\textbf{Synthetic task families used in this paper.} The first two rows are exchangeable ($0$-Markov) families; the third row instantiates $1$-Markov exchangeability, following \citet{Park2025}.}
  \label{tab:families}
  \footnotesize
  \renewcommand{\arraystretch}{1.15}
  \setlength{\tabcolsep}{6pt}
  \begin{tabularx}{\linewidth}{@{}>{\raggedright\arraybackslash}m{2.5cm} l >{\raggedright\arraybackslash}X >{\raggedright\arraybackslash}X@{}}
    \toprule
    Synthetic task family                                       & Latent task $\tilde\theta$                 & Likelihood                                                                                       & Prior                                                         \\
    \midrule
    Balls-and-urns\newline (App.~\ref{app:urn-baselines})       & $\tilde\theta = \tilde p \in \Delta^{|\mathcal Y|-1}$ & $Y_i\mid \tilde p \sim \operatorname{Categorical}(\tilde p)$                                     & $\Pi(\tilde p) = \operatorname{Dirichlet}(1,\dots,1)$          \\
    \midrule
    Linear regression\newline (App.~\ref{app:linear-baselines}) & $\tilde\theta = \tilde w \in \mathbb R^d$  & $X_i \sim \mathcal N(0,I_d)$, $Y_i\mid X_i,\tilde w \sim \mathcal N(\tilde w^\top X_i,\sigma^2)$ & $\Pi(\tilde w) = \mathcal N(0,I_d)$                            \\
    \midrule
    Markov chains\newline (App.~\ref{app:markov-baselines})     & $\tilde\theta = \tilde Q \in \mathcal Q$   & $Y_i\mid Y_{1:i-1},\tilde Q \sim \tilde Q(\cdot\mid Y_{i-1})$                                    & $\Pi(\tilde Q) =$ row-wise $\operatorname{Dirichlet}(1,\dots,1)$ \\
    \bottomrule
  \end{tabularx}
\end{table}

\section{Meta-learning on Bayes-filtered data}
\label{sec:framework}
This section presents the meta-learning framework, building toward a
definition for the \emph{Bayes-filtered transformer}
(Definition~\ref{def:bft}). We develop it in the unsupervised setting (token
sequences $Y_{1:n}$ on $\mathcal Y$, covering balls-and-urns and Markov chains);
linear regression is recovered by replacing $Y_i$ with the pair $(X_i, Y_i)$,
with $X_i\in\mathcal X$ an exogenous covariate.

\noindent\textbf{Pretraining distribution.} The meta-learning setup has two ingredients, both from Section~\ref{sec:in-context-kme}: a prior $\Pi$ on the latent task $\tilde\theta\in\Theta$ and, for each $\theta\in\Theta$, a probability kernel $P_\theta$ from $\mathcal Y^*$ to $\mathcal Y$. A pretraining sequence is drawn by sampling $\tilde\theta\sim \Pi$ and then generating $Y_{1:n}$ token by token from $P_{\tilde\theta}$, so the joint law of a pretraining sequence is exactly $\mathbb P$ of \eqref{eq:icb-joint}: the product inside the expectation there is the chain-rule factorization of the law of $Y_{1:n}$ conditional on $\tilde\theta$.

\noindent\textbf{From Bayesian prediction to optimization.} Let $T_\phi\colon \mathcal Y^*\to \Delta(\mathcal Y)$ be a sequence model, e.g. a transformer, where $\Delta(\mathcal Y)$ denotes the set of probability distributions on $\mathcal Y$.
By the Ionescu--Tulcea theorem \citep{Kallenberg2021}, $T_\phi$ induces a unique joint law $\mathbb P_\phi$ on $\mathcal Y^\infty$ whose one-step predictives are $T_\phi$ at every history, so $\mathbb P_\phi(Y_{1:n}=y_{1:n})=\prod_{i=1}^n T_\phi(y_i\mid y_{1:i-1})$.
We fit $T_\phi$ to $\mathbb P$ by minimizing the KL between the two joint laws:
\begin{equation}
\arg\min_\phi\; \mathrm{KL}\bigl(\mathbb P(Y_{1:n})\,\|\,\mathbb P_\phi(Y_{1:n})\bigr)
\;=\;
\arg\min_\phi\; \mathbb E_{\mathbb P}\!\left[-\sum_{i=1}^n \log T_\phi(Y_i\mid Y_{1:i-1})\right],
\label{eq:pop-risk}
\end{equation}
where the equality uses the factorization of $\mathbb P_\phi$ and drops the entropy of $\mathbb P$ (independent of $\phi$).
The result is the standard autoregressive log loss, which converts the computation of the Bayesian PPD of the pretraining model $(\Pi, \{P_\theta\}_{\theta\in\Theta})$, typically intractable, into an optimization problem.
For linear regression, $T_\phi$ conditions on the augmented prefix $(X_{1:i}, Y_{1:i-1})$ at position $i$ and predicts $Y_i$. The loss is otherwise unchanged.

\noindent\textbf{Per-token targets are Bayes-filtered.} At every position $i$, the objective \eqref{eq:pop-risk} fits $T_\phi$ to predict $Y_i$ given the realized prefix $Y_{1:i-1}$. Under the pretraining distribution, the conditional law of $Y_i$ given $Y_{1:i-1}$ is the Bayesian PPD,
\begin{equation}
\mathbb P(Y_i\in A\mid Y_{1:i-1})
=
\mathbb E_{\tilde\theta\sim \Pi(\cdot\mid Y_{1:i-1})}\!\left[P_{\tilde\theta}(A\mid Y_{1:i-1})\right],
\qquad A\subseteq\mathcal Y,
\label{eq:bayes-filtered}
\end{equation}
with $\Pi(\cdot\mid Y_{1:i-1})$ the posterior induced by $\Pi$ and the observed prefix. Each pretraining target has thus been \emph{filtered} through Bayes' rule against the latent task, and \citet{Ortega2019} call data with this structure \emph{Bayes-filtered}. 

\noindent\textbf{Meta-learning \citep{Ortega2019}.} \emph{Meta-learning} is empirical risk minimization of \eqref{eq:pop-risk} on a pretraining set $\mathcal D$ of sequences drawn from $\mathbb P$:
\begin{equation}
\arg\min_\phi\;\frac{1}{|\mathcal D|}\sum_{y_{1:n}\in \mathcal D}\Bigl[-\sum_{i=1}^n \log T_\phi(y_i\mid y_{1:i-1})\Bigr].
\label{eq:emp-risk}
\end{equation}
In our experiments, as in the source papers, every training step draws a fresh batch of sequences from $\mathbb P$, so no sequence is ever seen twice during training. In the memorizing setting of Section~\ref{sec:experiments}, what is finite is the set of latent tasks supporting the prior. We name a model trained this way after the structure of its per-token targets:

\begin{definition}[Bayes-filtered transformer]\label{def:bft}
  A \emph{Bayes-filtered transformer} (BFT) is a transformer $T_\phi$ fit by the meta-learning objective \eqref{eq:emp-risk} on sequences drawn from the pretraining joint law $\mathbb P$ of \eqref{eq:icb-joint}, for some choice of kernel family $\{P_\theta\}_{\theta\in\Theta}$ and prior $\Pi$.
\end{definition}

In the idealized limit of infinite pretraining data and a sufficiently
expressive $T_\phi$, the minimizer of \eqref{eq:emp-risk} reproduces the
Bayesian PPD \eqref{eq:bayes-filtered} at every position and prefix: the
optimal BFT \emph{is} the PPD of the pretraining model. In practice these
conditions fail, so any trained BFT is only an approximation to that ideal
object. PMC (Section~\ref{sec:pmc}) is designed to answer the interpretive
question of what prior and posterior over $\theta$ the trained BFT has actually
internalized.

The five source papers \citep{Raventos2023,Carroll2025,NEURIPS2025_abb22b1e,edelman2024the,Park2025} each train a BFT in the sense of Definition~\ref{def:bft}, instantiated with one of the families of Section~\ref{sec:in-context-kme}. Our implementation departs from the linear regression papers in the output head, which we discuss in Appendix~\ref{app:experiments}.

\section{Predictive Monte Carlo (PMC)}
\label{sec:pmc}

PMC \citep{Fortini2020} samples from the conditional law of the latent task
$\tilde\theta$ given a (possibly empty) prompt $y_{1:\promptlen}$. Applied to a
trained BFT, the underlying law is $\mathbb P_\phi$, the joint law that
$T_\phi$ induces on $\mathcal Y^\infty$ (Section~\ref{sec:framework}). When
$\mathbb P_\phi$ is exchangeable (balls-and-urns, linear regression),
de Finetti's theorem applies; when it is $1$-Markov exchangeable and
additionally recurrent (Markov chains), the Diaconis--Freedman theorem
applies. In either case the representation makes $\tilde\theta$ a function
of the infinite sample path $y_{1:\infty}$ (for linear regression, the pair
sequence $z_{1:\infty}$ with $z_t=(x_t,y_t)$). The explicit recovery in each family is:
\begin{itemize}[leftmargin=1.5em,itemsep=2pt,topsep=2pt]
  \item \textbf{Balls-and-urns.} $\tilde p_a = \lim_{n\to\infty}\frac{1}{n}\sum_{i=1}^n \mathbf 1\{y_i=a\}$ for each $a\in\mathcal Y$, $\mathbb P_\phi$-almost surely: the empirical distribution of the path.
  \item \textbf{Linear regression.} $\tilde w = \lim_{n\to\infty}\bigl(\sum_{i=1}^n x_i x_i^\top\bigr)^{-1}\sum_{i=1}^n x_i y_i$, $\mathbb P_\phi$-almost surely: the OLS estimator on the pair sequence.
  \item \textbf{Markov chains.} $\tilde Q$ is the row-wise limit of the matrix of normalized transition counts (i.e. the empirical transition matrix), with entries $\tilde Q_{a,b} = \lim_{n\to\infty} N_{a,b}(y_{1:n})\big/\sum_{b'\in\mathcal Y} N_{a,b'}(y_{1:n})$, $\mathbb P_\phi$-almost surely on the event that every state is visited infinitely often.
\end{itemize}
We construct a \textbf{single sample} of $\tilde\theta$ given $y_{1:\promptlen}$ in two steps:
\begin{enumerate}[label=(\roman*),leftmargin=1.8em,itemsep=2pt,topsep=2pt]
  \item \textbf{Rollout}: extend $y_{1:\promptlen}$ by iteratively drawing $Y_{\promptlen+1}\sim T_\phi(\cdot\mid y_{1:\promptlen})$, $Y_{\promptlen+2}\sim T_\phi(\cdot\mid y_{1:\promptlen+1})$, and so on. By construction, the completed path is distributed according to $\mathbb P_\phi$ conditioned on the path beginning with $y_{1:\promptlen}$.
  \item \textbf{Evaluate}: compute $\tilde\theta$ on the completed path. The resulting draw then has the conditional law of $\tilde\theta$ given $y_{1:\promptlen}$ under $\mathbb P_\phi$ (the BFT's implicit prior when $y_{1:\promptlen}$ is empty, the BFT's implicit posterior otherwise).
\end{enumerate}

Step~(i) cannot be carried out to infinity, so each rollout is truncated at
finite length $\rolllen$, and the infinite-path limits above are replaced by
finite-rollout estimators: empirical token frequencies for balls-and-urns
(Appendix~\ref{app:urn-baselines}), ordinary least squares on the pair sequence
for linear regression (Appendix~\ref{app:linear-baselines}), and
Laplace-smoothed empirical transition counts for Markov chains, where the
smoothing keeps every row of the estimated matrix well-defined when a state is
rarely visited within the truncated rollout
(Appendix~\ref{app:markov-baselines}). To produce $\pmcsamples$ approximate
samples of $\tilde\theta$ given $y_{1:\promptlen}$, we repeat the
rollout-and-evaluate construction $\pmcsamples$ times independently, at
$\pmcsamples\cdot\rolllen$ forward passes through $T_\phi$ in total.

\noindent\textbf{Validity.} The recovery of $\tilde\theta$ above rests on
$\mathbb P_\phi$ satisfying the hypotheses of the relevant representation
theorem: exchangeability for balls-and-urns and linear regression, $1$-Markov
exchangeability together with recurrence for Markov chains. If $T_\phi$ agrees
with the one-step predictives of the pretraining law $\mathbb P$ at every
history, then $\mathbb P_\phi=\mathbb P$ by uniqueness of the Ionescu--Tulcea
extension, and the hypotheses hold because $\mathbb P$ satisfies them by
construction. A trained BFT only approximates those predictives, so the
hypotheses must be established for $\mathbb P_\phi$ separately. The sufficient
conditions for PMC validity in the Bayesian predictive inference literature are
sufficient conditions for the process $(Y_n)$ to be \emph{asymptotically
exchangeable} under $\mathbb P_\phi$: the law of the shifted sequence
$(Y_{n+1}, Y_{n+2}, \dots)$ converges, as $n\to\infty$, to the law of an
exchangeable process \citep{Fortini2020,Fortini2025}. Appendix~\ref{app:pmc-diagnostics} reviews
these conditions in the exchangeable case, summarizes what is available for the
Markov-exchangeable case, and surveys recent empirical attempts at verifying
them in TabPFN, a foundation BFT.

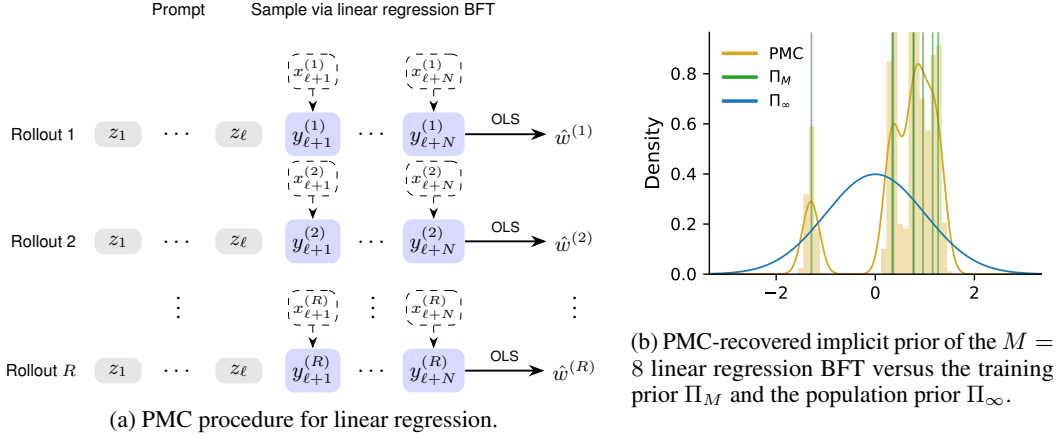
\begin{figure}[!htbp]
  \centering
  \begin{subfigure}[c]{0.58\linewidth}
    \centering
    \resizebox{\linewidth}{!}{%
      \begin{tikzpicture}[
          xscale=1.0, yscale=1.1,
          rollout/.style={font=\small},
          shared/.style={text=black, fill=gray!20, rounded corners, inner sep=3pt, minimum width=0.7cm, align=center},
          suffix/.style={text=black, fill=blue!15, rounded corners, inner sep=3pt, minimum width=0.7cm, align=center},
          input/.style={text=black, font=\scriptsize\sffamily, dashed, draw, rounded corners, inner sep=2pt},
          arrow/.style={->, thick, >=Stealth},
          header/.style={align=center, font=\scriptsize\sffamily, anchor=south}
        ]

        \node[header] (h1) at (0.9, 1.45) {Prompt};
        \node[header] (h2) at (3.8, 1.45) {Sample via linear regression BFT};

        \node[font=\sffamily\scriptsize, anchor=east] at (-0.5, 0) {Rollout 1};
        \node[rollout, shared] (z1_1) at (0, 0) {$z_1$};
        \node (z1_d) at (0.9, 0) {$\dots$};
        \node[rollout, shared] (z1_n) at (1.8, 0) {$z_{\promptlen}$};
        \node[input] (x1_1) at (2.9, 0.85) {$x_{\promptlen+1}^{(1)}$};
        \node[input] (x1_N) at (4.7, 0.85) {$x_{\promptlen+N}^{(1)}$};
        \node[rollout, suffix] (s1_1) at (2.9, 0) {$y_{\promptlen+1}^{(1)}$};
        \node (s1_d) at (3.8, 0) {$\dots$};
        \node[rollout, suffix] (s1_N) at (4.7, 0) {$y_{\promptlen+N}^{(1)}$};
        \draw[arrow, dashed, thin] (x1_1) -- (s1_1);
        \draw[arrow, dashed, thin] (x1_N) -- (s1_N);
        \node[font=\small] (theta1) at (6.8, 0) {$\hat w^{(1)}$};
        \draw[arrow] (s1_N) -- (theta1) node[midway, above=0.01cm, font=\tiny\sffamily] {OLS};

        \node[font=\sffamily\scriptsize, anchor=east] at (-0.5, -1.45) {Rollout 2};
        \node[rollout, shared] (z2_1) at (0, -1.45) {$z_1$};
        \node (z2_d) at (0.9, -1.45) {$\dots$};
        \node[rollout, shared] (z2_n) at (1.8, -1.45) {$z_{\promptlen}$};
        \node[input] (x2_1) at (2.9, -0.6) {$x_{\promptlen+1}^{(2)}$};
        \node[input] (x2_N) at (4.7, -0.6) {$x_{\promptlen+N}^{(2)}$};
        \node[rollout, suffix] (s2_1) at (2.9, -1.45) {$y_{\promptlen+1}^{(2)}$};
        \node (s2_d) at (3.8, -1.45) {$\dots$};
        \node[rollout, suffix] (s2_N) at (4.7, -1.45) {$y_{\promptlen+N}^{(2)}$};
        \draw[arrow, dashed, thin] (x2_1) -- (s2_1);
        \draw[arrow, dashed, thin] (x2_N) -- (s2_N);
        \node[font=\small] (theta2) at (6.8, -1.45) {$\hat w^{(2)}$};
        \draw[arrow] (s2_N) -- (theta2) node[midway, above=0.01cm, font=\tiny\sffamily] {OLS};

        \node at (0.9, -2.25) {$\vdots$};
        \node at (3.8, -2.25) {$\vdots$};
        \node at (6.8, -2.25) {$\vdots$};

        \node[font=\sffamily\scriptsize, anchor=east] at (-0.5, -3.2) {Rollout $R$};
        \node[rollout, shared] (zL_1) at (0, -3.2) {$z_1$};
        \node (zL_d) at (0.9, -3.2) {$\dots$};
        \node[rollout, shared] (zL_n) at (1.8, -3.2) {$z_{\promptlen}$};
        \node[input] (xL_1) at (2.9, -2.35) {$x_{\promptlen+1}^{(R)}$};
        \node[input] (xL_N) at (4.7, -2.35) {$x_{\promptlen+N}^{(R)}$};
        \node[rollout, suffix] (sL_1) at (2.9, -3.2) {$y_{\promptlen+1}^{(R)}$};
        \node (sL_d) at (3.8, -3.2) {$\dots$};
        \node[rollout, suffix] (sL_N) at (4.7, -3.2) {$y_{\promptlen+N}^{(R)}$};
        \draw[arrow, dashed, thin] (xL_1) -- (sL_1);
        \draw[arrow, dashed, thin] (xL_N) -- (sL_N);
        \node[font=\small] (thetaL) at (6.8, -3.2) {$\hat w^{(R)}$};
        \draw[arrow] (sL_N) -- (thetaL) node[midway, above=0.01cm, font=\tiny\sffamily] {OLS};

      \end{tikzpicture}
    }
    \caption{PMC procedure for linear regression.}
    \label{fig:pmc-schematic-lr}
  \end{subfigure}
  \hfill
  \begin{subfigure}[c]{0.40\linewidth}
    \centering
    \includegraphics[width=\linewidth]{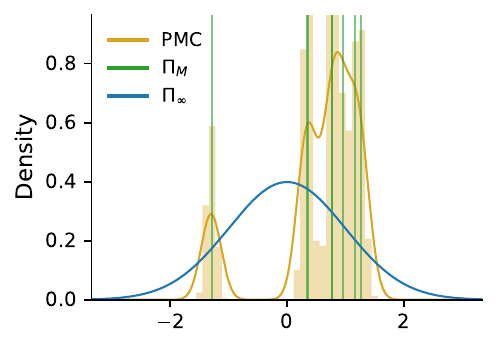}
    \caption{PMC-recovered implicit prior of the $\taskdiv=8$ linear regression BFT versus the training prior $\Pi_{\taskdiv}$ and the population prior $\Pi_\infty$.}
    \label{fig:linear-samples}
  \end{subfigure}
  \caption{\textbf{PMC for in-context linear regression.} (a) The latent task is the regression coefficient $w\in\mathbb R^d$. Unlike the Beta-Bernoulli setting of Figure~\ref{fig:pmc-schematic}, each rollout step pairs an exogenously sampled covariate $x_t\sim\mathcal N(0,I_d)$ with a transformer-supplied response $y_t\sim T_\phi(\cdot\mid x_t,z_{1:t-1})$, where $z_t=(x_t,y_t)$. The rollout estimator $\hat w^{(r)}$ is OLS on the completed sequence $z_{1:\promptlen+N}^{(r)}$. Panel~(a) is adapted from \citet{Ng2026}.
 (b) PMC applied with an empty prompt to the linear regression BFT trained at task diversity $\taskdiv=8$. The histogram shows the PMC samples of the first coordinate $w_0$ of $w$. Overlaid are the corresponding marginal densities of the training prior $\Pi_{\taskdiv}$ and of the population prior $\Pi_\infty=\mathcal N(0,I_d)$: the PMC samples concentrate near the former.
 Figure~\ref{fig:linear-samples-app} repeats this comparison for four coordinates of $w$, with density and CDF panels. The complete grid across all task diversities is in Appendix~\ref{app:stitched-marginals}.}
  \label{fig:lr-pmc-overview}
\end{figure}

\section{Experiments}
\label{sec:experiments}
This section trains BFTs in the three task families, applies PMC to each
trained model, and compares the recovered prior and posterior against two
reference posteriors, in both prediction space and latent space. For each
family we report three results: recovery of the implicit prior and posterior,
the task-diversity threshold, and transient generalization. Evaluation prompts
have length $\promptlen=16$ for balls-and-urns and $\promptlen=8$ for linear
regression and Markov chains. PMC settings (number of rollouts $\pmcsamples$
and rollout length $\rolllen$) are given per family in
Appendices~\ref{app:linear-training}, \ref{app:urn-baselines}, and
\ref{app:markov-baselines}.

\noindent\textbf{Training transformers across task families.} Architectural and training details are collected in Appendix~\ref{app:experiments}, with family-specific instantiations in Appendices~\ref{app:linear-training}, \ref{app:urn-baselines}, and \ref{app:markov-baselines}. Following \citet{Raventos2023}, we vary the \emph{task diversity} $\taskdiv$: the number of latent tasks in the training distribution. For each task family and each value of $\taskdiv$, we form the empirical prior $\Pi_{\taskdiv} := \mathrm{Unif}(\{\theta^{(1)}, \dots, \theta^{(\taskdiv)}\})$ by drawing a fresh set $\theta^{(1)}, \dots, \theta^{(\taskdiv)} \overset{\text{iid}}{\sim} \Pi_\infty$ once. The transformer is then trained on sequences drawn by sampling $\tilde\theta \sim \Pi_{\taskdiv}$ and then $Y_{1:n} \sim P_{\tilde\theta}$. We write $\mathbb{P}_{\taskdiv}$ for the joint law of these training sequences, which is \eqref{eq:icb-joint} with $\Pi_{\taskdiv}$ in place of $\Pi$.

\noindent\textbf{Memorizing and generalizing baselines.} The trained BFT is compared against two reference posteriors over $\theta$. Let $c = y_{1:\promptlen}$ be an observed prompt, and let $L_\theta(c) := \prod_{i=1}^{\promptlen} P_\theta(y_i \mid y_{1:i-1})$ be its likelihood under task $\theta$. For linear regression, the conditioning additionally includes the covariates, $L_w(c) = \prod_{i=1}^{\promptlen} P_w(y_i \mid x_i)$; the covariate density does not depend on the task, so it cancels in \eqref{eq:baseline-posts}. The \emph{memorizing} baseline uses the empirical prior $\Pi_{\taskdiv}$; the \emph{generalizing} baseline uses the population prior $\Pi_\infty$. The corresponding posteriors are
\begin{equation}
\Pi^{\mem}(\theta \mid c) \;\propto\; L_\theta(c)\, \Pi_{\taskdiv}(\theta),
\qquad
\Pi^{\gen}(\theta \mid c) \;\propto\; L_\theta(c)\, \Pi_\infty(\theta).
\label{eq:baseline-posts}
\end{equation}
Each induces a PPD for the next token. Family-specific derivations are in Appendices~\ref{app:linear-baselines}, \ref{app:urn-baselines}, and \ref{app:markov-baselines}.

\noindent\textbf{Prediction-space metrics.} The prediction-space metric is family-specific. We define the linear regression metric here; the metrics for balls-and-urns and Markov chains are defined in Appendix~\ref{app:pred-metrics}. A length-$\promptlen$ evaluation prompt is $c=((x_1,y_1),\dots,(x_{\promptlen},y_{\promptlen}))$. Let $\hat y_k(c)$ be $T_\phi$'s mean prediction of $y_k$ given the prefix $((x_1,y_1),\dots,(x_{k-1},y_{k-1}),x_k)$, and $\hat y_k^\mem(c), \hat y_k^\gen(c)$ the corresponding posterior predictive means under the priors $\Pi_\taskdiv$ and $\Pi_\infty$ respectively. Following \citet{Raventos2023}, we report the mean-squared gap, averaged over positions, between the transformer's predictions and a reference's predictions:
\begin{equation}
\Delta_{\mathrm{MSE}}
\;:=\;
\E_{c \sim P}\!\left[\, \frac{1}{\promptlen}\sum_{k=1}^{\promptlen} \bigl(\hat y_k(c)-\hat y_k^{\mathrm{ref}}(c)\bigr)^2 \,\right],
\label{eq:pred-delta-mse}
\end{equation}
where $\mathrm{ref}$ refers either to $\mem$ or $\gen$. We evaluate this gap on either in-distribution ($P = \mathbb{P}_M$) or out-of-distribution ($P = \mathbb{P}$) data. Throughout the experiments, $\mathbb P$ denotes the joint law \eqref{eq:icb-joint} under $\Pi = \Pi_\infty$. The empirical estimator of \eqref{eq:pred-delta-mse} is given in Appendix~\ref{app:pred-estimators}.

\noindent\textbf{Latent-space metrics.} Let $\widehat\Pi(\cdot\mid c)$ be the PMC-recovered distribution conditioned on prompt $c$, and let $\Pi^{\mathrm{ref}}(\cdot\mid c)$ denote either baseline from \eqref{eq:baseline-posts}, with $\mathrm{ref}$ standing for $\mem$ or $\gen$. Both live on $\mathbb R^D$, where $D$ is the dimension in which the latent task is represented: $D=|\mathcal Y|=12$ for balls-and-urns (the probability vector), $D=d=8$ for linear regression, and $D=|\mathcal Y|^2=100$ for Markov chains (the flattened transition matrix). Our first metric is the squared \emph{energy distance} \citep{Rizzo2016}. For $U,U'\overset{\mathrm{iid}}{\sim}\widehat\Pi(\cdot\mid c)$ and $V,V'\overset{\mathrm{iid}}{\sim}\Pi^{\mathrm{ref}}(\cdot\mid c)$, all four independent, it is
\begin{equation}
\mathcal{E}^{2}(U, V)
\;:=\;
2\,\E\|U-V\| \;-\; \E\|U-U'\| \;-\; \E\|V-V'\|.
\label{eq:energy-distance-population}
\end{equation}
Our second metric compares the same two distributions through their
one-dimensional projections. For the same $U$ and $V$, the \emph{sliced
Wasserstein-1 distance} \citep{Bonneel2015,Kolouri2019} is
\begin{equation}
\mathrm{SW}_{1}(U, V)
\;:=\;
\E_{\vartheta\sim\mathrm{Unif}(\mathbb S^{D-1})}\, W_{1}\bigl(\vartheta^\top U,\,\vartheta^\top V\bigr),
\label{eq:sliced-wasserstein-population}
\end{equation}
where $W_1$ denotes the $1$-Wasserstein distance between the laws of the
scalar projections $\vartheta^\top U$ and $\vartheta^\top V$. For the
posterior, we average each distance over prompts $c \sim P$, with prompts
originating either from in-distribution ($P = \mathbb{P}_M$) or
out-of-distribution ($P = \mathbb{P}$) data. For the prior, we set $c =
\varnothing$ and take the reference to be $\Pi_\taskdiv$ (memorizing) or
$\Pi_\infty$ (generalizing). There is a single prompt, so no averaging over
prompts is involved. In practice each expectation in
\eqref{eq:energy-distance-population} and
\eqref{eq:sliced-wasserstein-population} is estimated from the finite PMC and
reference samples. The estimators are written out in
Appendix~\ref{app:latent-distances}.

\noindent\textbf{Linear regression (0-Markov exchangeable).} PMC recovers the implicit prior and posterior at task diversity $\taskdiv=8$: the PMC prior marginals match the training prior $\Pi_{\taskdiv}$ (Figure~\ref{fig:linear-samples}), and conditioning on an in-distribution prompt of length $8$, the PMC posterior samples closely match the closed-form memorizing posterior. On an out-of-distribution prompt of the same length, the induced posterior samples match neither the memorizing nor the generalizing posterior (Figure~\ref{fig:linear-samples-app}). Two explanations are consistent with this mismatch: the trained model may genuinely depart from both reference posteriors on out-of-distribution prompts, or the sufficient conditions for PMC validity (Section~\ref{sec:pmc}) may fail for $\mathbb P_\phi$. We do not know which explanation is operative. Separating them requires verifying those conditions (Appendix~\ref{app:pmc-diagnostics}).

\begin{figure}[!htbp]
  \centering
  \begin{subfigure}[t]{0.49\linewidth}
    \centering
    \includegraphics[width=\linewidth]{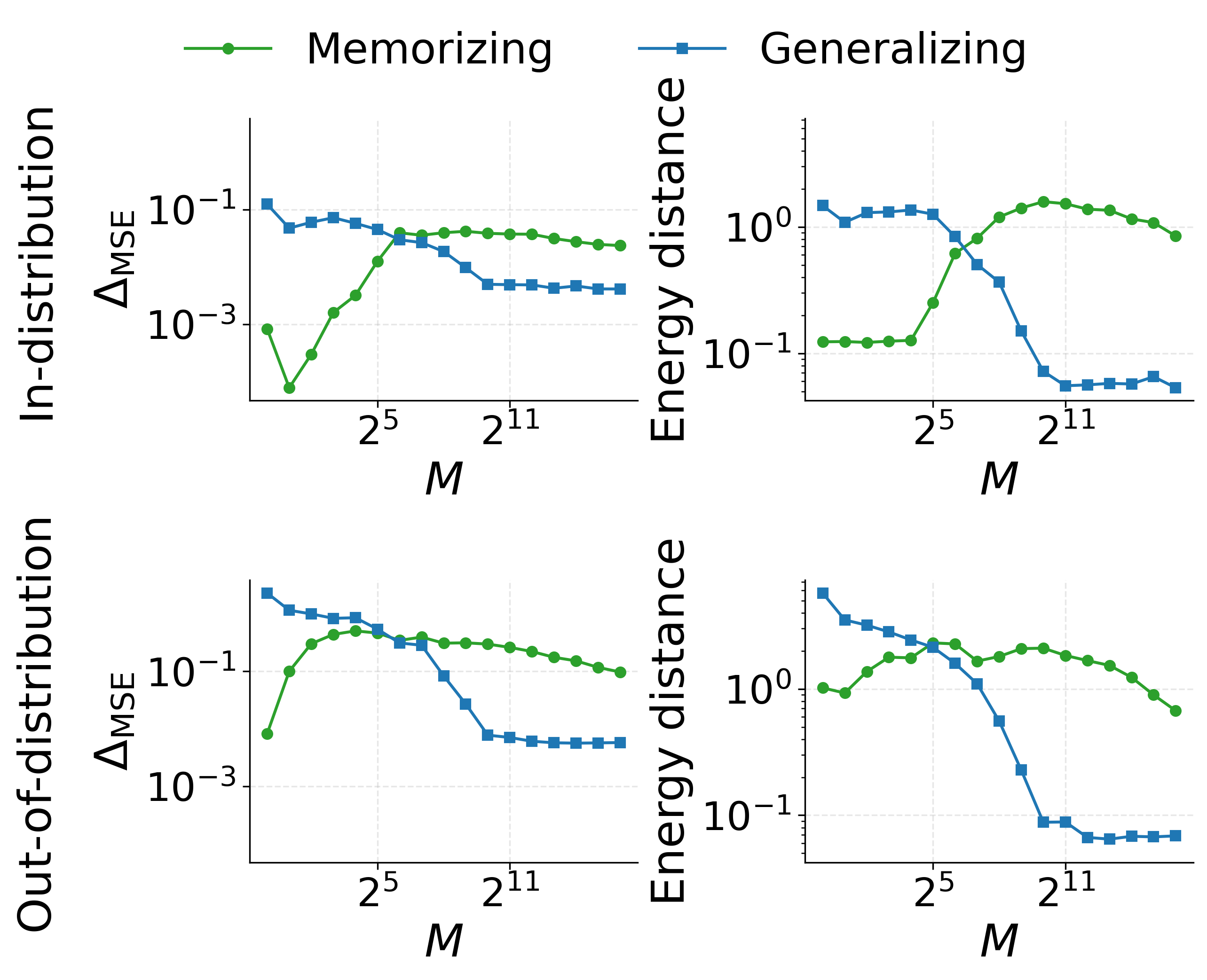}
    \caption{Task-diversity sweep at end of training.}
  \end{subfigure}
  \hfill
  \begin{subfigure}[t]{0.49\linewidth}
    \centering
    \includegraphics[width=\linewidth]{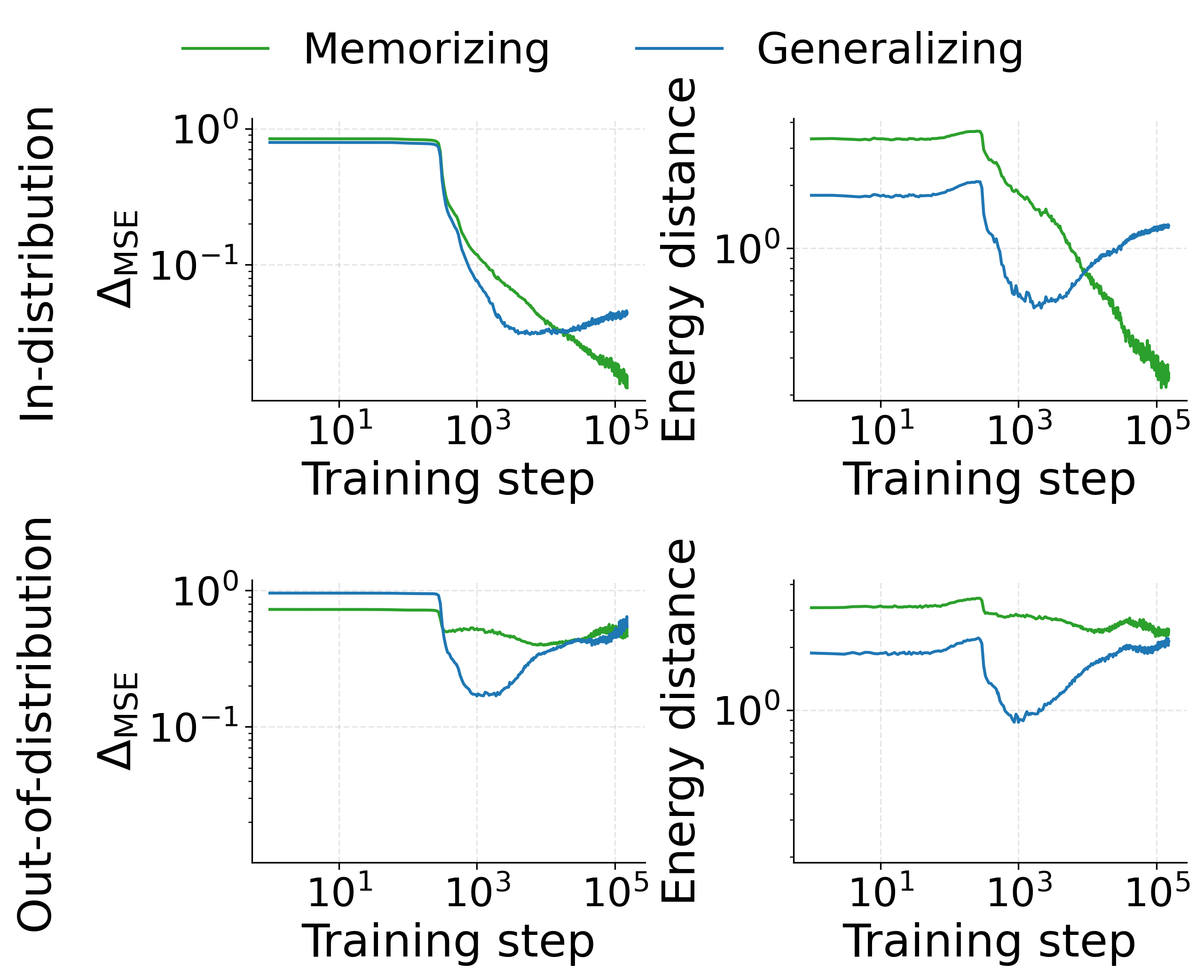}
    \caption{Training dynamics at $\taskdiv=32$.}
  \end{subfigure}
  \caption{\textbf{Linear regression: task-diversity threshold and transient generalization.} (a) As task diversity $\taskdiv$ grows, the PMC posterior moves from the memorizing baseline toward the generalizing baseline. (b) At intermediate task diversity $\taskdiv=32$, the BFT first approximates the generalizing baseline before specializing to the memorizing baseline. Rows: in-distribution prompts (top) and out-of-distribution prompts (bottom). Columns: distance to each baseline in prediction space ($\Delta_{\mathrm{MSE}}$) and in the posterior (energy distance). Sliced Wasserstein versions of the posterior column are in Appendix~\ref{app:linear-training}: Figure~\ref{fig:linear-diversity-app} for (a) and Figure~\ref{fig:linear-transient-app} for (b).}
  \label{fig:lr-diversity-transient}
\end{figure}

At low task diversity $\taskdiv$, the priors
$\Pi_\taskdiv$ and $\Pi_\infty$ are well separated, and the PMC prior matches
$\Pi_\taskdiv$. At high task diversity, $\Pi_\taskdiv$ approaches
$\Pi_\infty$, and the distances from the PMC prior to the two references
become harder to distinguish. In the posterior, the task-diversity threshold
of \citet{Raventos2023} appears on both in-distribution and
out-of-distribution prompts: as $\taskdiv$ increases, the PMC posterior moves
away from the memorizing posterior and toward the generalizing posterior
(Figure~\ref{fig:lr-diversity-transient}(a)). At intermediate $\taskdiv=32$, the BFT first approximates the
generalizing baseline and later moves toward the memorizing baseline
(Figure~\ref{fig:lr-diversity-transient}(b)), reproducing the transient
generalization of \citet{Carroll2025} in both prediction space (left column)
and in the posterior (right column).

\noindent\textbf{Balls-and-urns (0-Markov exchangeable).} PMC again recovers the implicit prior and posterior at task diversity $\taskdiv=8$: the PMC prior marginals match the training prior $\Pi_{\taskdiv}$ (Figure~\ref{fig:bau-samples}, Appendix~\ref{app:urn-baselines}), and conditioning on an in-distribution prompt of length $16$, the PMC posterior samples match the closed-form memorizing posterior. An out-of-distribution prompt of the same length gives a similar result.

\begin{figure}[!htbp]
  \centering
  \begin{subfigure}[t]{0.49\linewidth}
    \centering
    \includegraphics[width=\linewidth]{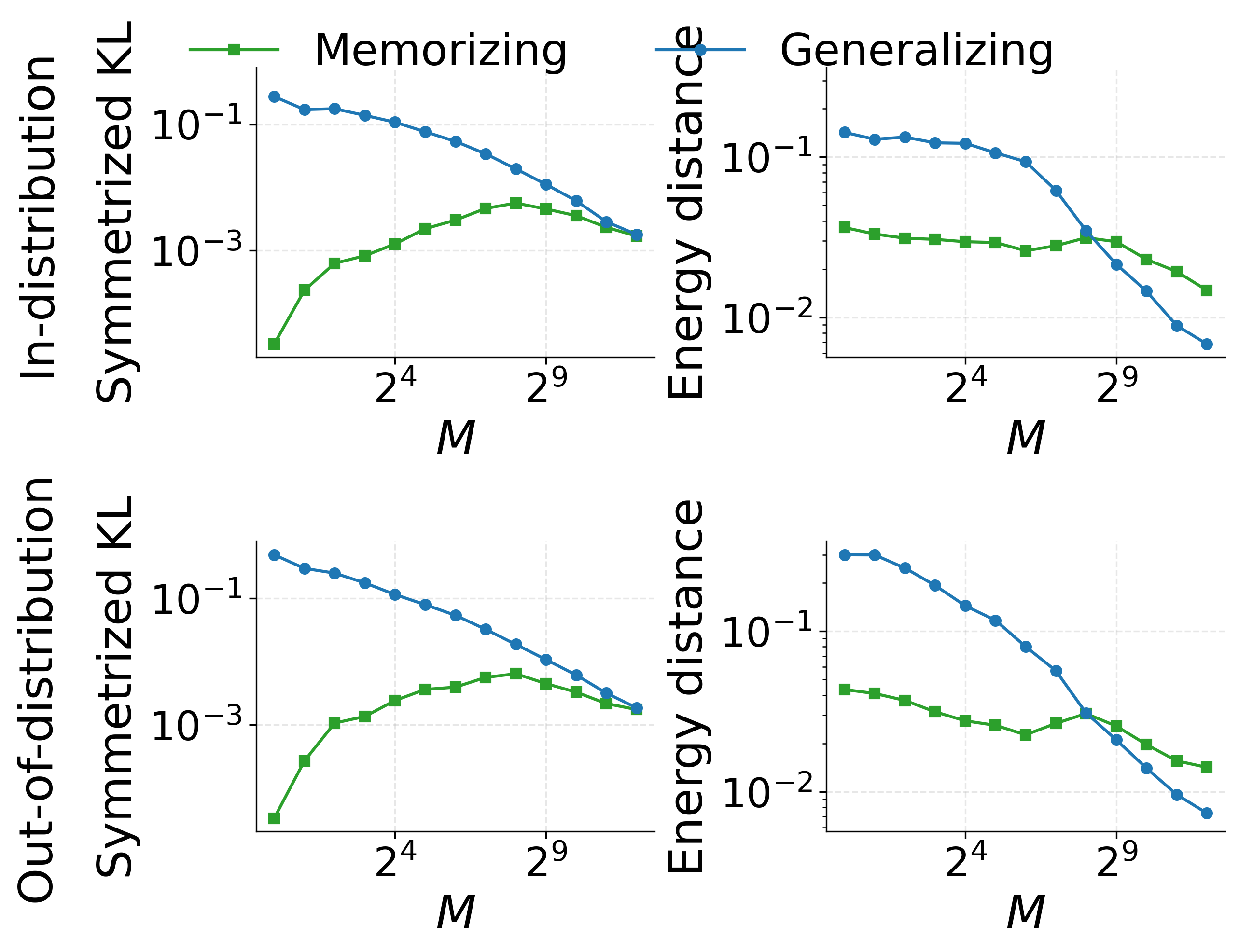}
    \caption{Task-diversity sweep at end of training.}
  \end{subfigure}
  \hfill
  \begin{subfigure}[t]{0.49\linewidth}
    \centering
    \includegraphics[width=\linewidth]{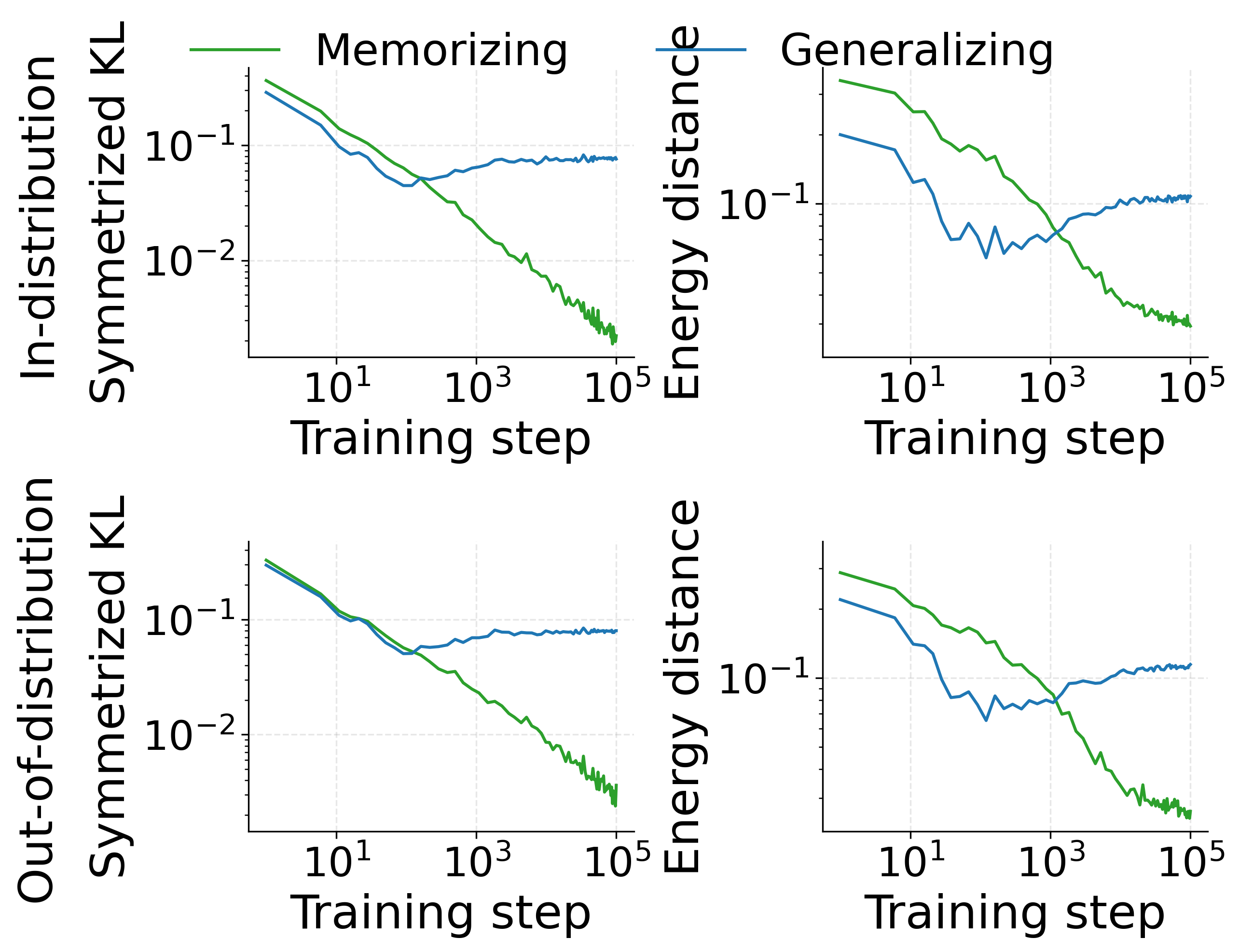}
    \caption{Training dynamics at $\taskdiv=32$.}
  \end{subfigure}
  \caption{\textbf{Balls-and-urns: task-diversity threshold and transient generalization.} (a) As task diversity $\taskdiv$ grows, the PMC posterior moves from the memorizing baseline toward the generalizing baseline. (b) At intermediate task diversity $\taskdiv=32$, the BFT first approximates the generalizing baseline before specializing to the memorizing baseline. Rows: in-distribution prompts (top) and out-of-distribution prompts (bottom). Columns: distance to each baseline in prediction space (symmetrized KL) and in the posterior (energy distance). Sliced Wasserstein versions of the posterior column are in Appendix~\ref{app:urn-baselines}: Figure~\ref{fig:bau-diversity} for (a) and Figure~\ref{fig:bau-transient} for (b).}
  \label{fig:bau-main}
\end{figure}

The task-diversity threshold appears in the posterior (Figure~\ref{fig:bau-main}(a)). At this prompt length, the prediction-space view does not exhibit the transition, as we discuss in Appendix~\ref{app:urn-baselines}. Transient generalization appears at task diversity $\taskdiv=32$: the posterior first approximates the generalizing baseline before specializing to the memorizing baseline (Figure~\ref{fig:bau-main}(b)).

\noindent\textbf{Markov chains (1-Markov exchangeable).} PMC recovers the implicit prior and posterior at task diversity $\taskdiv=4$: the PMC prior marginals match the training prior $\Pi_{\taskdiv}$ (Figure~\ref{fig:markov-samples}, Appendix~\ref{app:markov-baselines}), and conditioning on an in-distribution prompt of length $8$, the PMC posterior samples match the memorizing posterior. An out-of-distribution prompt of the same length gives a similar result. The match is less apparent than in the other task families, as we discuss in Appendix~\ref{app:markov-baselines}.

\begin{figure}[!htbp]
  \centering
  \begin{subfigure}[t]{0.49\linewidth}
    \centering
    \includegraphics[width=\linewidth]{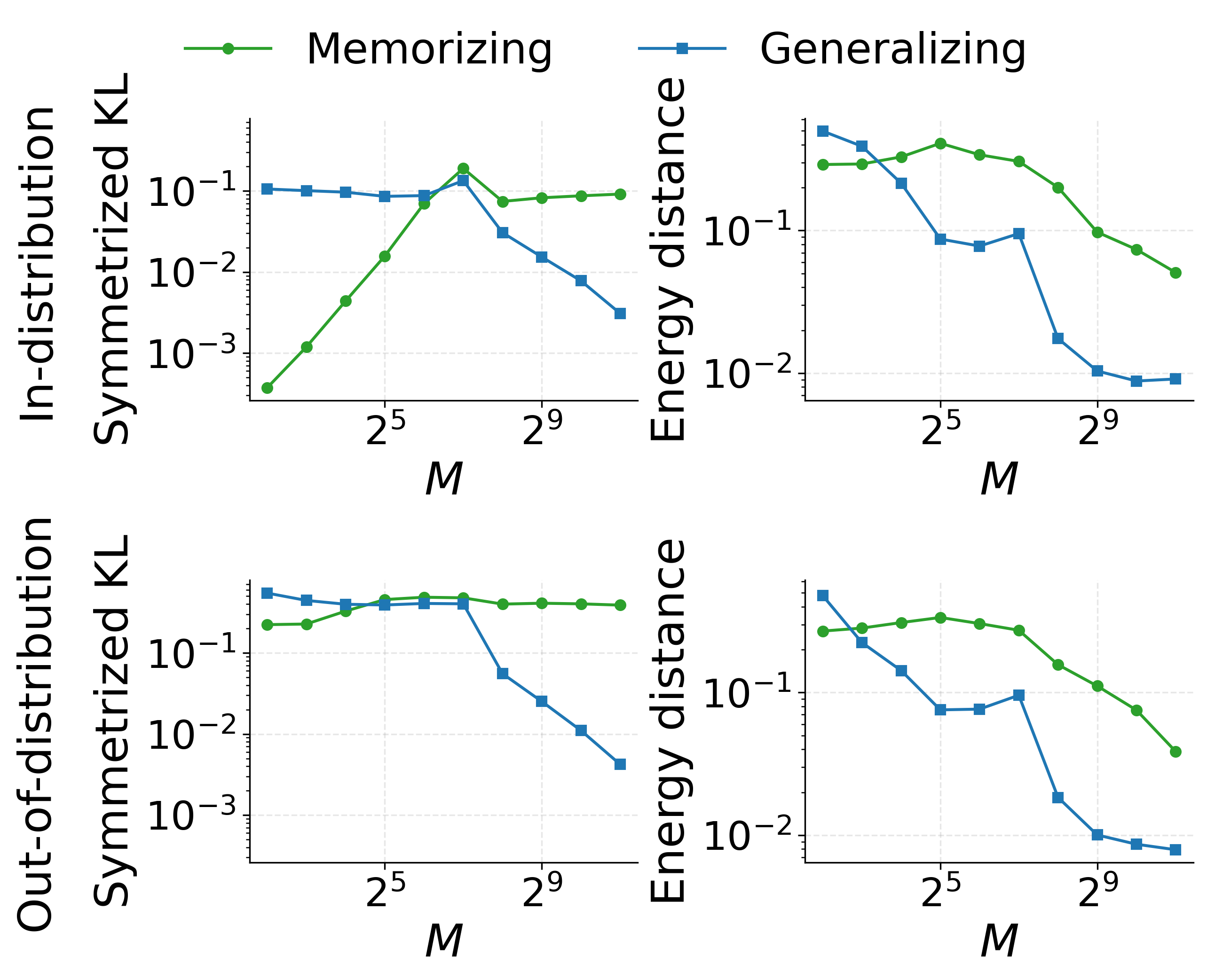}
    \caption{Task-diversity sweep at end of training.}
  \end{subfigure}
  \hfill
  \begin{subfigure}[t]{0.49\linewidth}
    \centering
    \includegraphics[width=\linewidth]{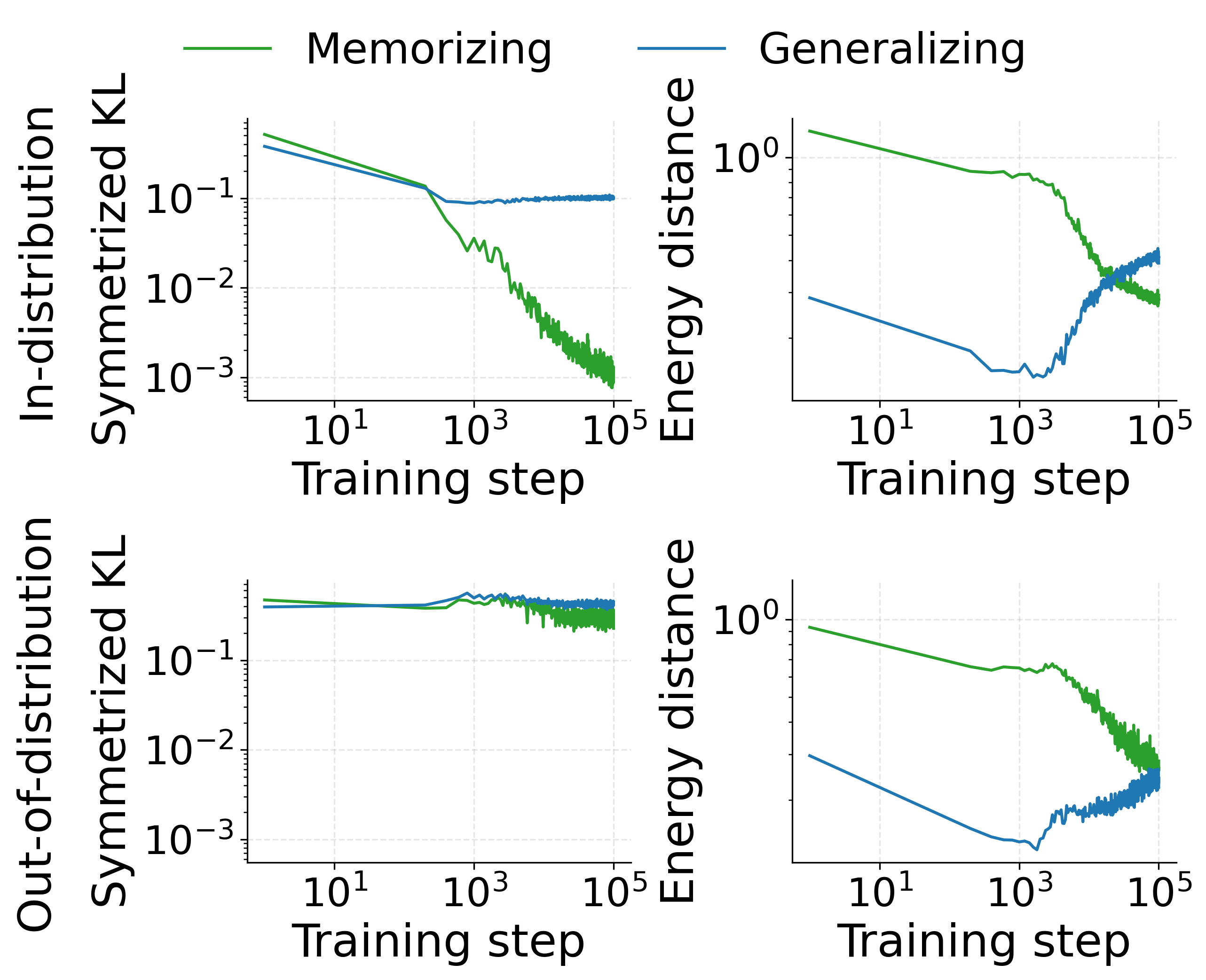}
    \caption{Training dynamics at $\taskdiv=8$.}
  \end{subfigure}
  \caption{\textbf{Markov chains: task-diversity threshold and transient generalization.} (a) As task diversity $\taskdiv$ grows, the PMC posterior moves from the memorizing baseline toward the generalizing baseline. (b) At task diversity $\taskdiv=8$, the BFT first approximates the generalizing baseline before specializing to the memorizing baseline. Rows: in-distribution prompts (top) and out-of-distribution prompts (bottom). Columns: distance to each baseline in prediction space (symmetrized KL) and in the posterior (energy distance). Sliced Wasserstein versions of the posterior column are in Appendix~\ref{app:markov-baselines}: Figure~\ref{fig:markov-diversity} for (a) and Figure~\ref{fig:markov-transient} for (b).}
  \label{fig:markov-main}
\end{figure}

The task-diversity threshold appears in the posterior (Figure~\ref{fig:markov-main}(a)). Transient generalization appears at task diversity $\taskdiv=8$: the posterior first approximates the generalizing baseline before specializing to the memorizing baseline (Figure~\ref{fig:markov-main}(b)).

\section{Related work}
\label{sec:related}

\noindent\textbf{The Bayesian view of ICL.} Explanations of ICL in LLMs \citep{NEURIPS2020_1457c0d6} include Bayesian accounts \citep{Xie2022}, induction heads \citep{olsson2022context}, and implicit gradient descent \citep{pmlr-v202-von-oswald23a,akyurek2023what}. We follow the Bayesian line that posits the transformer's next-token predictions arise from a posterior over latent task structure updated by the context. \citet{Xie2022} introduced this view as an explanation for ICL in pretrained LLMs, and \citet{falck2024is} tested it empirically on LLMs through the martingale property. Our work moves to the narrower setting of synthetic studies, where transformers are trained on hierarchical pretraining distributions.
Within this synthetic setting, \citet{NEURIPS2022_c529dba0} and \citet{panwar2024incontext} show that transformers can learn simple function classes in context. Follow-up work \citep{Raventos2023,Carroll2025,NEURIPS2025_abb22b1e,edelman2024the,Park2025} asks which Bayesian scheme the trained transformer adopts: a memorizing one (using the empirical pretraining prior) or a generalizing one (using the population prior).

\noindent\textbf{BFTs.} \citet{Ortega2019} established the equivalence between meta-learning on hierarchical data and Bayesian posterior prediction. We adopt this view to define the idealized BFT in Section~\ref{sec:framework}. \citet{Mikulik2020,Genewein2023,pmlr-v235-grau-moya24a} show empirically that trained sequence models approach the Bayesian PPD on controlled synthetic tasks. Prior-data fitted networks \citep[PFNs;][]{Muller2024} are BFTs that use a transformer and restrict attention to exchangeable data. \citet{Hollmann2023,Hollmann2025} scale this construction into TabPFN, a tabular foundation model.

\noindent\textbf{BPI.}
\citet{Fortini2020} introduced PMC, which uses a sequence's predictive distribution to draw samples from the prior or posterior over its directing random measure. \citet{Fortini2025} survey the surrounding predictive framework. \citet{Fong2024} recast it as \emph{predictive resampling} under the \emph{martingale posterior} framework, a generalized Bayes setting in which the target functional need not index the probability model. \citet{Battiston2025} and \citet{fortini2026principledframeworkuncertaintydecomposition} progressively weaken the sufficient conditions under which BPI applies to general sequence models, and both \citet{Ng2026}, who adapt the martingale posterior to TabPFN (TabMGP), and \citet{fortini2026principledframeworkuncertaintydecomposition}, who develop an asymptotic uncertainty decomposition for TabPFN under a quasi-martingale condition, apply BPI to transformer foundation models. A subsequent version of the latter \citep{fortini2026uncertaintydecompositionbft} relaxes the quasi-martingale condition to a weaker \emph{signed} condition.

\noindent\textbf{Related interpretability work.} \citet{NEURIPS2024_8936fa16} study transformers trained on sequences from a hidden Markov model (HMM) data generating process and show that the Bayesian posterior over the HMM's hidden state is represented in the residual stream, by training a linear probe against the ground-truth posterior. Their setup shows the posterior is encoded within the model but does not provide a way to recover it without access to the ground truth. We instead use PMC to read beliefs off the trained transformer itself, without inspecting model internals. Whether PMC samples relate to the belief-state geometry \citep{NEURIPS2024_8936fa16} is a direction for future work.

\section{Conclusion}

\textbf{Summary.} This paper repurposes PMC as a black-box interpretability tool for BFTs: using only next-token generation from the trained model, PMC returns samples from the prior and posterior the model has internalized over its latent task variable. Identifiability motivates moving the evaluation of ICL from prediction space to latent space: distinct posteriors can produce identical predictions, so prediction-space diagnostics cannot tell them apart. We demonstrate the methodology on the three task families (balls-and-urns, linear regression, and Markov chains) and reproduce the task-diversity threshold and transient generalization phenomena directly in latent space.

\textbf{Limitations.} PMC validity rests on sufficient conditions that are hard to verify for BFTs, both theoretically and empirically. Theoretically, we cannot currently show anything substantive about a real trained transformer, and the sufficient conditions for PMC validity from the BPI literature are no exception. Empirically, the brute-force quasi-martingale tests of \citet{fortini2026principledframeworkuncertaintydecomposition} on TabPFN, a foundation BFT, demand Monte Carlo budgets so large that verification has so far been inconclusive. Closing the gap calls for advances along two orthogonal axes: weaker sufficient conditions for PMC validity, and empirical diagnostics with tractable sample complexity. Beyond these verification concerns, whether PMC transfers directly to LLMs is unknown. BFTs relate to Solomonoff induction, the idealized predictor that mixes over all computable generative programs \citep{pmlr-v235-grau-moya24a}, and LLMs have themselves been linked to Solomonoff induction \citep{deletang2024language}, so the transfer is worth investigating.

\textbf{Outlook.} Even absent direct transfer to frontier LLMs, BFTs are appealing as a model system for phenomena that in LLMs are entangled with many confounding factors. Emergent misalignment \citep{pmlr-v267-betley25a} is a concrete candidate: narrow finetuning on misbehaved examples induces broadly misaligned behavior, a mechanism difficult to isolate at scale and already suspected to operate through shifts in a latent persona representation \citep{pmlr-v267-wang25ak}. A BFT whose prior predictive encodes a mixture of behavioral personas would offer a controlled setting in which PMC can track the posterior over personas before, during, and after exposure to adversarial examples. The same template extends to other LLM-scale phenomena whose underlying belief dynamics are suspected but not directly observable.

\section*{Acknowledgements}
This work was supported by The Alignment Project (grant AP-S2-100069), funded by the UK AI Security Institute, Department for Science, Innovation and Technology. This research was supported in part by the MASSIVE HPC facility (www.massive.org.au).

\bibliographystyle{plainnat}
\bibliography{references}

\clearpage
\appendix
\raggedbottom

\section*{Appendix contents}
\begin{itemize}[leftmargin=2.5em, itemsep=0pt]
  \item[\textbf{\ref*{app:notation}}] Notation summary \dotfill \pageref{app:notation}
  \item[\textbf{\ref*{app:kmarkov}}] General $k$-Markov exchangeability \dotfill \pageref{app:kmarkov}
  \item[\textbf{\ref*{app:pmc-diagnostics}}] Sufficient conditions for PMC \dotfill \pageref{app:pmc-diagnostics}
  \item[\textbf{\ref*{app:experiments}}] Experimental setup \dotfill \pageref{app:experiments}
  \item[\textbf{\ref*{app:beta-bernoulli}}] Beta-Bernoulli \dotfill \pageref{app:beta-bernoulli}
  \item[\textbf{\ref*{app:linear-training}}] Linear regression \dotfill \pageref{app:linear-training}
  \item[\textbf{\ref*{app:urn-baselines}}] Balls-and-urns \dotfill \pageref{app:urn-baselines}
  \item[\textbf{\ref*{app:markov-baselines}}] Markov chains \dotfill \pageref{app:markov-baselines}
  \item[\textbf{\ref*{app:stitched-marginals}}] Marginal densities and CDFs from PMC \dotfill \pageref{app:stitched-marginals}
\end{itemize}

\clearpage

\section{Notation summary}
\label{app:notation}

Table~\ref{tab:notation-summary} records the main notation used throughout the
manuscript. Symbols that appear in only one family appendix are defined where
they are used.

\begin{table}[H]
  \centering
  \caption{Core notation used throughout the paper.}
  \label{tab:notation-summary}
  \begin{tabular}{ll}
    \toprule
    Symbol                     & Meaning                                                                                           \\
    \midrule
    $\mathcal Y$               & finite token alphabet                                                                             \\
    $\theta\in\Theta$          & latent task parameter and its space (e.g.\ regression coefficient $w$, urn probability vector $p$, transition matrix $Q$) \\
    $\tilde\theta$             & the latent task as a random variable, $\tilde\theta\sim\Pi$; plain $\theta$ denotes values       \\
    $\tilde P$                 & directing random measure of an exchangeable process (Section~\ref{sec:exch})                     \\
    $\Pi$                      & prior on the latent task space (instantiated as $\Pi_\infty$ or $\Pi_{\taskdiv}$)                \\
    $\Pi_\infty$               & population prior on the latent task space                                                        \\
    $\Pi_{\taskdiv}$           & empirical prior formed from $\taskdiv$ sampled tasks                                             \\
    $\taskdiv$                 & task diversity, i.e.\ number of latent tasks in the empirical prior                              \\
    $P_\theta$                 & probability kernel from $\mathcal Y^*$ to $\mathcal Y$: after history $h$, the next token is drawn from $P_\theta(\cdot\mid h)$ \\
    $\mathbb P$                & joint law on $\mathcal Y^\infty$ of \eqref{eq:icb-joint}, induced by $\Pi$; in the experiments, $\Pi=\Pi_\infty$ \\
    $\mathbb P_{\taskdiv}$     & joint law on $\mathcal Y^\infty$ induced by $\Pi_{\taskdiv}$                                      \\
    $T_\phi$                   & sequence model $\mathcal Y^*\to\Delta(\mathcal Y)$, e.g.\ a transformer                           \\
    $\mathbb P_\phi$           & joint law on $\mathcal Y^\infty$ induced by $T_\phi$                                              \\
    $P_n$                      & one-step predictive of $\mathbb P$, see~\eqref{eq:predictive-def}                                 \\
    $P^\phi_n$                 & one-step predictive of $\mathbb P_\phi$ (Appendix~\ref{app:pmc-diagnostics})                      \\
    $\metatrainlen$            & pretraining sequence length                                                                       \\
    $c$                        & evaluation prompt, $c = y_{1:\promptlen}$                                                         \\
    $\promptlen$               & evaluation prompt length                                                                          \\
    $\rolllen$                 & PMC rollout length                                                                                \\
    $\pmcsamples$              & number of independent rollouts in PMC                                                             \\
    $\Pi^{\mem}(\cdot\mid c)$  & memorizing posterior, under the prior $\Pi_{\taskdiv}$, see~\eqref{eq:baseline-posts}             \\
    $\Pi^{\gen}(\cdot\mid c)$  & generalizing posterior, under the prior $\Pi_\infty$, see~\eqref{eq:baseline-posts}               \\
    $\widehat\Pi(\cdot\mid c)$ & PMC empirical law over the latent task conditional on prompt $c$                                  \\
    \bottomrule
  \end{tabular}
\end{table}

\section{General $k$-Markov exchangeability}
\label{app:kmarkov}

For completeness, we record the generalization of
Section~\ref{sec:1markov-exch} to higher Markov orders. Our experiments use
only $k=0$ and $k=1$, but the higher-order case covers the $n$-gram
extensions studied by \citet{edelman2024the}, where the next-token
distribution is drawn afresh for each length-$(n-1)$ context: their $n$-gram
model is the case $k = n-1$ in the notation of this appendix.

Fix $k\ge 1$. We call $(Y_i)$ \emph{$k$-Markov exchangeable} if the sequence
of overlapping $k$-blocks
\[
(Y_{1:k}),\ (Y_{2:k+1}),\ (Y_{3:k+2}),\ \dots
\]
is Markov exchangeable in the sense of Section~\ref{sec:1markov-exch} when
viewed as a process on $\mathcal Y^k$.

Under the corresponding recurrence condition, the Diaconis--Freedman
representation yields a mixture of $1$-step Markov chains on $\mathcal Y^k$.
Because consecutive blocks of a path overlap in $k-1$ coordinates, the mixing
kernel is supported on transitions between overlapping blocks, and a $1$-step
chain on $\mathcal Y^k$ with that support is exactly a $k$th-order Markov
chain on $\mathcal Y$.

For finite $\mathcal Y$, a $k$th-order transition matrix is a row-stochastic
array
\[
Q=(Q_{u,y})_{u\in\mathcal Y^k,\,y\in\mathcal Y},
\qquad
\sum_{y\in\mathcal Y} Q_{u,y}=1 \quad\text{for every } u\in\mathcal Y^k.
\]
Writing $\mathcal Q_k$ for the set of such arrays, \eqref{eq:icb-general}
holds with $\Pi$ supported on $\mathcal Q_k$: the latent kernel depends on the
history only through the last $k$ tokens. A population prior on
$\mathcal Q_k$ is typically implemented as independent Dirichlet priors over
the transition probabilities indexed by length-$k$ contexts.

\section{Sufficient conditions for PMC}
\label{app:pmc-diagnostics}

PMC is applied to the trained BFT $T_\phi$, so the conditions below are
conditions on the model's induced joint $\mathbb P_\phi$
(Section~\ref{sec:framework}). What PMC needs from $\mathbb P_\phi$ depends on
which representation theorem governs the task family: de Finetti for
balls-and-urns and linear regression, Diaconis--Freedman for Markov chains. We
treat the two cases separately. For the exchangeable case, the BPI literature
provides several sufficient conditions for asymptotic exchangeability, ordered
below from strongest to weakest. For the Markov-exchangeable case, no
comparable set of conditions has been articulated.

\noindent\textbf{Sufficient conditions for asymptotic exchangeability.}
In the evaluate step of Section~\ref{sec:pmc}, the latent task is read off
the limit of an infinite rollout. For that limit to exist, the one-step
predictives of the model's law, which we write
$P^\phi_n := \mathbb P_\phi(Y_{n+1}\in\cdot\mid Y_{1:n})$ to distinguish
them from the predictives $P_n$ of $\mathbb P$ in
\eqref{eq:predictive-def}, must converge weakly to a random probability
measure $\tilde P$ on $\mathcal Y$, $\mathbb P_\phi$-almost surely. The
convergence is path by path: for $\mathbb P_\phi$-almost every sample path
$\omega$, the sequence of predictive distributions $P^\phi_n(\omega)$ has a
weak limit $\tilde P(\omega)$, so the limit is a random measure whose
realization depends on the path. Given this convergence, PMC's two-step
construction returns a sample from the posterior of $\tilde P$ given
$y_{1:\promptlen}$. In the parameterized families, $\tilde P$ ranges over
the token distributions indexed by the latent task, for instance
$\tilde P = \operatorname{Categorical}(\tilde p)$ for balls-and-urns, so a
sample of $\tilde P$ given $y_{1:\promptlen}$ is a sample of $\tilde\theta$
given $y_{1:\promptlen}$.

The BPI literature provides several sufficient conditions on
$\mathbb P_\phi$ for this convergence. In the list below, (i) implies (ii),
and (ii) implies both (iii) and (iv): a martingale has zero conditional
drift, so it satisfies each relaxation trivially. The literature has not
compared (iii) and (iv) to each other; the example after the list shows they
are formally distinct.
\begin{enumerate}[label=(\roman*),leftmargin=1.8em,itemsep=4pt,topsep=2pt]
  \item $\mathbb P_\phi$ is exchangeable. The predictive distributions of an exchangeable process converge weakly, almost surely, to its directing random measure \citep{Fortini2025}, so $P^\phi_n \Rightarrow \tilde P$, $\mathbb P_\phi$-almost surely.
  \item $(P^\phi_n)$ is a measure-valued martingale, equivalently $\mathbb P_\phi$ is conditionally identically distributed (c.i.d.); martingale convergence then supplies the almost-sure limit \citep{Fortini2025}. The martingale property has been checked empirically on trained models. \citet{falck2024is} show it fails for in-context predictives of large language models. \citet{NaglerRugamer2025} find that TabPFN fails it as well, and for that reason they abandon TabPFN as the predictive rule after the first step, switching to a copula update that preserves the martingale property.
  \item $\mathbb P_\phi$ is asymptotically c.i.d.\ (a.c.i.d.), introduced by \citet{Battiston2025}: the conditional drift of the predictive is bounded by an almost-surely vanishing sequence. \citet{Ng2026} check this condition empirically for TabPFN and find that TabPFN departs from it, yet the resulting martingale posteriors remain stable with near-nominal frequentist coverage, suggesting the condition is stricter than necessary.
  \item $(P^\phi_n)$ is a quasi-martingale, setwise: the predictive update $\Delta_n(A) := P^\phi_n(A) - P^\phi_{n-1}(A)$ satisfies, for every $A\subseteq\mathcal Y$,
        \begin{equation}
        \sum_{n\ge 1}\mathbb E_{\mathbb P_\phi}\bigl[\bigl|\mathbb E_{\mathbb P_\phi}[\Delta_n(A) \mid Y_{1:n-1}]\bigr|\bigr] < \infty.
        \label{eq:qm}
        \end{equation}
        In contrast with (iii), the conditional drift here must be summable in expectation. \citet{fortini2026principledframeworkuncertaintydecomposition} check this condition empirically for TabPFN; the Monte Carlo budgets required are so large that the verification has so far been inconclusive. A subsequent version \citep{fortini2026uncertaintydecompositionbft} relaxes this quasi-martingale condition to a weaker \emph{signed} condition, requiring only that the signed tail sums of the conditional drift vanish, and reports positive diagnostics on a small Beta-Bernoulli BFT while the TabPFN diagnostics remain inconclusive.
\end{enumerate}
A c.i.d.\ sequence satisfies both (iii) and (iv) because its conditional drift is identically zero. Conditions (iii) and (iv) themselves are close in spirit but formally distinct: a drift bound that vanishes without being summable satisfies the former and can violate the latter.

\noindent\textbf{The Markov-exchangeable case.} Here $P^\phi_n$ depends on the most recent state $Y_n$ and does not converge pointwise to a single random measure on $\mathcal Y$. The relevant random object is instead the random transition matrix $\tilde Q\in\mathcal Q$ on $\mathcal Y\times\mathcal Y$ from Section~\ref{sec:1markov-exch}, obtained as the row-wise a.s.\ limit of the matrix of normalized transition counts. The exact condition on $\mathbb P_\phi$ that delivers this convergence is the Diaconis--Freedman setting: $\mathbb P_\phi$ is Markov exchangeable and the chain is recurrent under $\mathbb P_\phi$ (every state in $\mathcal Y$ is visited infinitely often, $\mathbb P_\phi$-almost surely) \citep{Fortini2025}.

Beyond this exact condition, no analogous hierarchy of relaxed sufficient
conditions for the Markov-exchangeable case has been articulated in the BPI
literature, to our knowledge. The c.i.d./a.c.i.d./quasi-martingale machinery
underlying conditions (ii)--(iv) above gives asymptotic exchangeability
\citep{Fortini2025,Battiston2025,fortini2026principledframeworkuncertaintydecomposition};
the partially c.i.d.\ extension of \citet{Fortini2025} addresses partial
exchangeability across multiple parallel sequences (the multi-experiment
setting), which is a different form of partial exchangeability from Markov
exchangeability. Natural Markov analogs of (ii)--(iv) (a Markov c.i.d.\
condition; an asymptotic Markov exchangeability relaxation; a quasi-martingale
condition for the matrix of transition counts) and their corresponding
empirical diagnostics on trained foundation BFTs remain open.

\noindent\textbf{Verification in this paper.} We do not verify any of these conditions for the BFTs studied here. In the controlled families of Section~\ref{sec:experiments} the reference prior and posteriors over $\theta$ are available either in closed form or by direct Monte Carlo from the prior predictive, so PMC outputs $\widehat\Pi(\cdot\mid c)$ can be compared against them directly. In our experiments this match occurs in two regimes. First, for every family, the PMC prior matches the training prior $\Pi_{\taskdiv}$. Second, on in-distribution prompts, the PMC posterior matches the memorizing posterior $\Pi^{\mem}(\cdot\mid c)$. In these regimes the agreement itself certifies the PMC output: whatever the status of the sufficient conditions above, the recovered distribution coincides with a known reference.
Where the output matches no reference, as for out-of-distribution prompts in linear regression (Section~\ref{sec:experiments}), the comparison cannot separate two explanations: the trained BFT may genuinely depart from both reference posteriors, or $\mathbb P_\phi$ may violate the conditions above, in which case PMC samples lose their interpretation as draws from an implicit posterior. 
No affordable empirical test of the conditions above yet exists for trained BFTs. Developing one would separate the two explanations.

\section{Experimental setup}
\label{app:experiments}

This appendix records the training protocol, architectural choices, and
reporting conventions shared across task families. Family-specific
instantiations (state-space size, model width, PMC rollout length,
task-diversity grid) are in Appendices~\ref{app:linear-training},
\ref{app:urn-baselines}, and \ref{app:markov-baselines}.

\noindent\textbf{Training protocol.} For each task family $f$ and each chosen task-diversity level $\taskdiv$, we train a separate transformer on sequences drawn from the empirical mixture law $\mathbb P_{\taskdiv}$, equivalently from the empirical task prior $\Pi_{\taskdiv}$. We write $\metatrainlen$ for the meta-training sequence length and use one fixed value of $\metatrainlen$ per task family in the main text. At evaluation time we draw prompts from the length-$\promptlen$ marginals of $\mathbb P_{\taskdiv}$ and $\mathbb P$. Prompt length $\promptlen$ may vary within a family, and checkpoint sweeps may be included when studying transient behavior, but neither prompt source nor checkpoint index is treated as a primary axis of the experimental design.

\noindent\textbf{Architecture.} Table~\ref{tab:architecture} summarizes the architectural choices across task families and source papers, alongside our own choices. Entries marked with ${}^\dagger$ are not explicitly stated in the source paper but inferred from the named architecture (GPT-2 implies learned absolute positional embeddings; GPT-NeoX implies RoPE). The main difference from the source papers that study linear regression is the output head: where \citet{Raventos2023} and \citet{NEURIPS2025_abb22b1e} use a scalar output head trained under squared-error loss, we discretize the response variable and use a categorical output head trained under log loss, so that the transformer returns the full next-token distribution needed for PMC rollouts. Squared-error training coincides with the meta-learning objective \eqref{eq:emp-risk} under a Gaussian output head of fixed variance and therefore fits only the mean of the PPD. Using a single categorical head across all three families keeps the architecture and loss uniform, at the price of not exploiting the closed form of the linear regression PPD. For linear regression, we use a different attention masking scheme than causal masking to enforce covariates to attend to covariate-response pairs. We refer to this masking scheme as `Autoregressive-PFN' and refer to Appendix~\ref{app:linear-training} for details.

For the two discrete families (balls-and-urns and Markov chains), a BOS
token is required so that the BFT can express a marginal predictive
$P^\phi_0$ for the first observation, matching $P_0$ in the ideal limit. For linear regression, the exogenous covariate
$x_1$ seeds the first prediction and no BOS token is needed. We refer to
Appendices~\ref{app:linear-training}, \ref{app:urn-baselines}, and
\ref{app:markov-baselines} for particular details on the transformer
architecture used in each task family. We include a summary of training
hyperparameters in the main text in Table~\ref{tab:training-hyperparameters}.

\begin{table}[H]
  \centering
  \caption{\textbf{Transformer architecture across task families and source papers.} Entries labeled with ${}^\dagger$ are inferred from the named architecture and not explicitly stated in the text. See Table~\ref{tab:training-hyperparameters} for specific task-family training hyperparameters.}
  \label{tab:architecture}
  \smallskip
  \renewcommand{\arraystretch}{1.3}
  \resizebox{\textwidth}{!}{%
    \begin{tabular}{@{}lllllll@{}}
      \toprule
      \textbf{Task family} & \textbf{Source}              & \textbf{Attention} & \textbf{Pos.\ encoding}      & \textbf{Output head}   & \textbf{Sequence length} & \textbf{BOS} \\
      \midrule
      \multirow{3}{*}{Linear regression}
                           & \citet{Raventos2023}         & Causal             & Learned absolute${}^\dagger$ & Scalar (MSE)           & 16                       & No           \\
                           & \citet{NEURIPS2025_abb22b1e} & Causal${}^\dagger$ & RoPE${}^\dagger$             & Scalar (MSE)           & \{16, 32, 64\}           & No           \\
                           & Ours                         & Autoregressive-PFN & Learned absolute             & Categorical (log loss) & 64                       & No           \\
      \midrule
      \multirow{2}{*}{Balls-and-urns}
                           & \citet{NEURIPS2025_abb22b1e} & Causal${}^\dagger$ & RoPE${}^\dagger$             & Categorical (log loss) & \{128, 256, 320\}        & N/A          \\
                           & Ours                         & Causal             & Learned absolute             & Categorical (log loss) & 256                      & Yes          \\
      \midrule
      \multirow{2}{*}{\shortstack[l]{Markov chains                                                                                                                                    \\(row-wise Dirichlet)}}
                           & \citet{Park2025}             & Causal             & RoPE                         & Categorical (log loss) & 512                      & No           \\
                           & Ours                         & Causal             & RoPE                         & Categorical (log loss) & 1024                     & Yes          \\
      \bottomrule
    \end{tabular}}% end resizebox
\end{table}

\begin{table}[H]
  \centering
  \caption{\textbf{Training hyperparameters.}}
  \label{tab:training-hyperparameters}
  \smallskip
  \renewcommand{\arraystretch}{1.2}
  \setlength{\tabcolsep}{6pt}
  \small
  \resizebox{\textwidth}{!}{%
    \begin{tabular}{@{}llccc@{}}
      \toprule
      \textbf{Category} & \textbf{Hyperparameter}                 & \textbf{Linear regression}            & \textbf{Balls-and-urns}      & \textbf{Markov chains}       \\
      \midrule
      Data              & Covariate dimension ($d$)               & $8$                                   & N/A                          & N/A                          \\
                        & State-space size ($|\mathcal Y|$)       & N/A                                   & $12$                         & $10$                         \\
                        & Observation noise variance ($\sigma^2$) & $0.25$                                & N/A                          & N/A                          \\
                        & Training sequence length                & $64$ pairs                            & $255$ observations (+ BOS)   & $1023$ observations (+ BOS)  \\
      \midrule
      Model             & Number of layers                        & $2$                                   & $2$                          & $2$                          \\
                        & Number of attention heads               & $4$                                   & $1$                          & $4$                          \\
                        & Embedding dimension                     & $512$                                 & $64$                         & $64$                         \\
                        & Attention head dimension                & $128$                                 & $64$                         & $16$                         \\
                        & MLP hidden dimension                    & $512$                                 & $256$                        & $256$                        \\
                        & Activation                              & GELU                                  & GELU                         & ReLU                         \\
                        & Layer normalization                     & Pre                                   & Pre                          & Pre                          \\
                        & Positional encoding                     & Learned absolute                      & Learned absolute             & RoPE                         \\
                        & Attention masking                       & Autoregressive-PFN                    & Causal                       & Causal                       \\
                        & Output head                             & Categorical, $256$ bins on $[-10,10]$ & Categorical, vocabulary $12$ & Categorical, vocabulary $11$ \\
                        & BOS token                               & No                                    & Yes                          & Yes                          \\
                        & Max context length                      & $128$                                 & $256$                        & $1024$                       \\
                        & Total parameters                        & $\approx 3.5$M                        & $\approx 118$K               & $\approx 100$K               \\
      \midrule
      Optimizer         & Type                                    & AdamW                                 & AdamW                        & AdamW                        \\
                        & Weight decay                            & $10^{-5}$                             & $10^{-5}$                    & $0.01$                       \\
                        & Gradient clipping (norm)                & $1$                                   & $1$                          & N/A                          \\
      \midrule
      LR schedule       & Shape                                   & Linear warmup, then constant          & Constant (no warmup)         & Constant (no warmup)         \\
                        & Peak learning rate                      & $10^{-4}$                             & $5\times 10^{-4}$            & $6\times 10^{-4}$            \\
                        & Warmup steps                            & $1{,}000$                             & $0$                          & $0$                          \\
      \midrule
      Experiments       & Batch size                              & $256$                                 & $64$                         & $128$                        \\
                        & Training steps                          & $150{,}000$                           & $100{,}000$                  & $100{,}000$                  \\
                        & Random seed                             & $42$                                  & $42$                         & $42$                         \\
                        & Task-diversity grid                     & $\{2^j : 0 \le j \le 16\}$            & $\{2^j : 0 \le j \le 12\}$   & $\{2^j : 2 \le j \le 11\}$   \\
      \bottomrule
    \end{tabular}}% end resizebox
\end{table}

\noindent\textbf{Prediction-space metrics beyond linear regression.}
\label{app:pred-metrics}
For balls-and-urns, let $p_k(c)$ denote the BFT's next-token distribution at position $k$ of prompt $c=(y_1,\dots,y_\promptlen)$, and $p_k^{\mathrm{ref}}(c)$ the corresponding next-token distribution of the reference Bayes predictor from \eqref{eq:baseline-posts}, with $\mathrm{ref}$ standing for $\mem$ or $\gen$ as in \eqref{eq:pred-delta-mse}. The symmetrized-KL prompt gap of \citet{NEURIPS2025_abb22b1e} is
\begin{equation}
\Delta_{\mathrm{sKL}}
\;:=\;
\E_{c \sim P}\!\left[\, \frac{1}{\promptlen}\sum_{k=1}^{\promptlen} \tfrac{1}{2}\bigl( \mathrm{KL}(p_k(c) \| p_k^{\mathrm{ref}}(c)) + \mathrm{KL}(p_k^{\mathrm{ref}}(c) \| p_k(c)) \bigr) \,\right],
\label{eq:pred-skl}
\end{equation}
with $P$ as in \eqref{eq:pred-delta-mse}.

For the Markov chains family we use the same $\Delta_{\mathrm{sKL}}$, with the
reference taken as each of the memorizing and generalizing posterior
predictives of \eqref{eq:baseline-posts}. We note that \citet{Park2025} report a
sequence-level KL metric of the transformer's empirical transition matrix
against the data-generating kernel. As we need a way to compare the
transformer's predictions against the memorizing and generalizing baselines, we
use the $\Delta_{\mathrm{sKL}}$ instead.

\noindent\textbf{Empirical estimators of the latent-space metrics.}
\label{app:latent-distances}
We compute the latent-space metrics \eqref{eq:energy-distance-population} and \eqref{eq:sliced-wasserstein-population} with the estimators below. For the prior, the reported quantities are
\begin{equation}
\mathcal{E}^2(U, V) \quad\text{and}\quad \mathrm{SW}_1(U, V), \qquad U \sim \widehat\Pi(\cdot\mid \varnothing),\quad V \sim \Pi^* \in \{\Pi_{\taskdiv},\,\Pi_\infty\},
\label{eq:latent-prior-gap}
\end{equation}
and for the posterior they are averaged over prompts,
\begin{equation}
\begin{aligned}
   & \E_{c \sim P}\!\left[\,\mathcal{E}^2(U, V)\right] \quad\text{and}\quad \E_{c \sim P}\!\left[\,\mathrm{SW}_1(U, V)\right], \\
   & U \sim \widehat\Pi(\cdot\mid c),\quad V \sim \Pi^*(\cdot\mid c)\in\{\Pi^\mem(\cdot\mid c),\,\Pi^\gen(\cdot\mid c)\}.
\end{aligned}
\label{eq:latent-post-gap}
\end{equation}
We estimate each of these distances from two empirical samples of size $\pmcsamples$: a PMC sample $U=\{u_i\}_{i=1}^{\pmcsamples}$ and a baseline sample $V=\{v_j\}_{j=1}^{\pmcsamples}$.
For the prior, we draw $U$ from the transformer's implicit prior $\widehat\Pi(\cdot\mid\varnothing)$ via PMC, and we draw $V$ from either the training prior $\Pi_\taskdiv$ or the population prior $\Pi_\infty$.
For the posterior, we draw $U$ from the transformer's implicit posterior $\widehat\Pi(\cdot\mid c)$ via PMC conditioned on $c$, and we draw $V$ from either the memorizing posterior $\Pi^\mem(\cdot\mid c)$ or the generalizing posterior $\Pi^\gen(\cdot\mid c)$.

We estimate the energy distance \eqref{eq:energy-distance-population} with the
unbiased U-statistic
\begin{equation}
\widehat{\mathcal{E}}^{2}(U,V)
\;=\;
\frac{2}{\pmcsamples^2}\sum_{i,j}\|u_i-v_j\|
\;-\;\frac{1}{\pmcsamples(\pmcsamples-1)}\sum_{i\ne i'}\|u_i-u_{i'}\|
\;-\;\frac{1}{\pmcsamples(\pmcsamples-1)}\sum_{j\ne j'}\|v_j-v_{j'}\|.
\label{eq:energy-distance}
\end{equation}
We report $\widehat{\mathcal{E}}^{2}$ directly rather than its square root.

For the sliced Wasserstein-1 distance \eqref{eq:sliced-wasserstein-population},
given empirical samples $U,V$ of size $\pmcsamples$ and directions
$\vartheta_1,\dots,\vartheta_L$ drawn iid from $\mathrm{Unif}(\mathbb
  S^{D-1})$, the estimator is
\begin{equation}
\widehat{\mathrm{SW}}_{1}(U,V)
\;=\;
\frac{1}{L}\sum_{s=1}^{L}\frac{1}{\pmcsamples}\sum_{k=1}^{\pmcsamples}
\bigl|u^{(k)}_{s}-v^{(k)}_{s}\bigr|,
\label{eq:sliced-wasserstein}
\end{equation}
with $u^{(k)}_{s}, v^{(k)}_{s}$ the $k$-th order statistics of $\{\vartheta_s^\top u_i\}_i$ and $\{\vartheta_s^\top v_j\}_j$. We take $L=100$.

\noindent\textbf{Empirical estimators of the prediction-space metrics.}
\label{app:pred-estimators} In \eqref{eq:pred-delta-mse} and \eqref{eq:pred-skl}, every quantity inside the prompt expectation is computed exactly: $\hat y_k(c)$ is the mean of the transformer's categorical output distribution, and the reference predictions and next-token distributions are computed in closed form (Appendices~\ref{app:linear-baselines}, \ref{app:urn-baselines}, and \ref{app:markov-baselines}). Only the prompt expectation $\E_{c\sim P}$ requires estimation. Given evaluation prompts $c_1,\dots,c_{n_P}$ drawn once from $P$, the estimator of \eqref{eq:pred-delta-mse} is
\begin{equation}
\widehat\Delta_{\mathrm{MSE}}
\;=\;
\frac{1}{n_P}\sum_{m=1}^{n_P}\, \frac{1}{\promptlen}\sum_{k=1}^{\promptlen}\bigl(\hat y_k(c_m)-\hat y_k^{\mathrm{ref}}(c_m)\bigr)^2,
\label{eq:pred-delta-mse-hat}
\end{equation}
and the estimator $\widehat\Delta_{\mathrm{sKL}}$ of \eqref{eq:pred-skl} replaces the squared difference with the symmetrized KL summand. The same prompt sets are used to estimate the prompt expectations in the posterior latent-space metrics \eqref{eq:latent-post-gap}. The number $n_P$ and length $\promptlen$ of evaluation prompts for each family are recorded in Appendices~\ref{app:linear-training}, \ref{app:urn-baselines}, and \ref{app:markov-baselines}.

\noindent\textbf{Reported figures and 1-D slices.} For each task family, three sets of figures are reported: PMC prior and posterior samples; task diversity in the prior and posterior; and transient generalization in the posterior. Posterior-space sweeps and dynamics appear in Section~\ref{sec:experiments}; prior-space sweeps and samples appear in the family-specific appendices. Because the full $(\taskdiv,\text{training step})$ grid is large, each of the latter two figures is a 1-D slice. For the \emph{task-diversity sweep}, we sweep all $\taskdiv$ at the final checkpoint. For the \emph{transient-generalization experiment}, we fix an intermediate task diversity $\taskdiv$ where transient generalization is most pronounced and sweep all saved training checkpoints. Table~\ref{tab:slices} records the specific choices for each family.

\begin{table}[H]
  \centering
  \caption{\textbf{Experimental slices.} For each task family and phenomenon we report results along a 1-D slice of the task-diversity $\times$ training-step space, sweeping one axis and fixing the other.}
  \label{tab:slices}
  \smallskip
  \begin{tabular}{@{}lll@{}}
    \toprule
    \textbf{Phenomenon (swept axis)} & \textbf{Task family} & \textbf{Fixed axis} \\
    \midrule
    \multirow{3}{*}{Task-diversity threshold (all $\taskdiv$)}
                                     & Linear regression    & final checkpoint    \\
                                     & Balls-and-urns       & final checkpoint    \\
                                     & Markov chains        & final checkpoint    \\
    \midrule
    \multirow{3}{*}{Transient generalization (all checkpoints)}
                                     & Linear regression    & $\taskdiv=32$       \\
                                     & Balls-and-urns       & $\taskdiv=32$       \\
                                     & Markov chains        & $\taskdiv=8$        \\
    \bottomrule
  \end{tabular}
\end{table}

\FloatBarrier

\clearpage
\section{Beta-Bernoulli}
\label{app:beta-bernoulli}

\noindent\textbf{Data-generating setup.}
Training sequences were drawn by first drawing a ground truth probability
$\theta^\star$ from a $\operatorname{Beta}(1,1)$ prior, then drawing independent
Bernoulli observations with probability $\theta^\star$.

\noindent\textbf{Transformer architecture.}
The Beta-Bernoulli BFT was trained with $2$ layers with embedding
dimension $d_\mathrm{model}=128$, $2$ attention heads with dimension
$d_\mathrm{head} = 32$, MLP hidden dimension $d_\mathrm{mlp} = 512$ and a context length of $512$ tokens (a BOS token followed by $511$ observations). The BOS token allows the model to represent an unconditioned
predictive distribution.

\noindent\textbf{Training details.}
Training was performed over $100{,}000$ steps with a batch size of $128$ and seed $0$. We use a linear learning rate schedule with $500$ warmup steps, staying constant at $10^{-4}$ thereafter. The model is trained autoregressively with next-token cross-entropy loss.

\begin{figure}[H]
  \centering
  \includegraphics[
    width=\linewidth,
    height=0.9\textheight,
    keepaspectratio
  ]{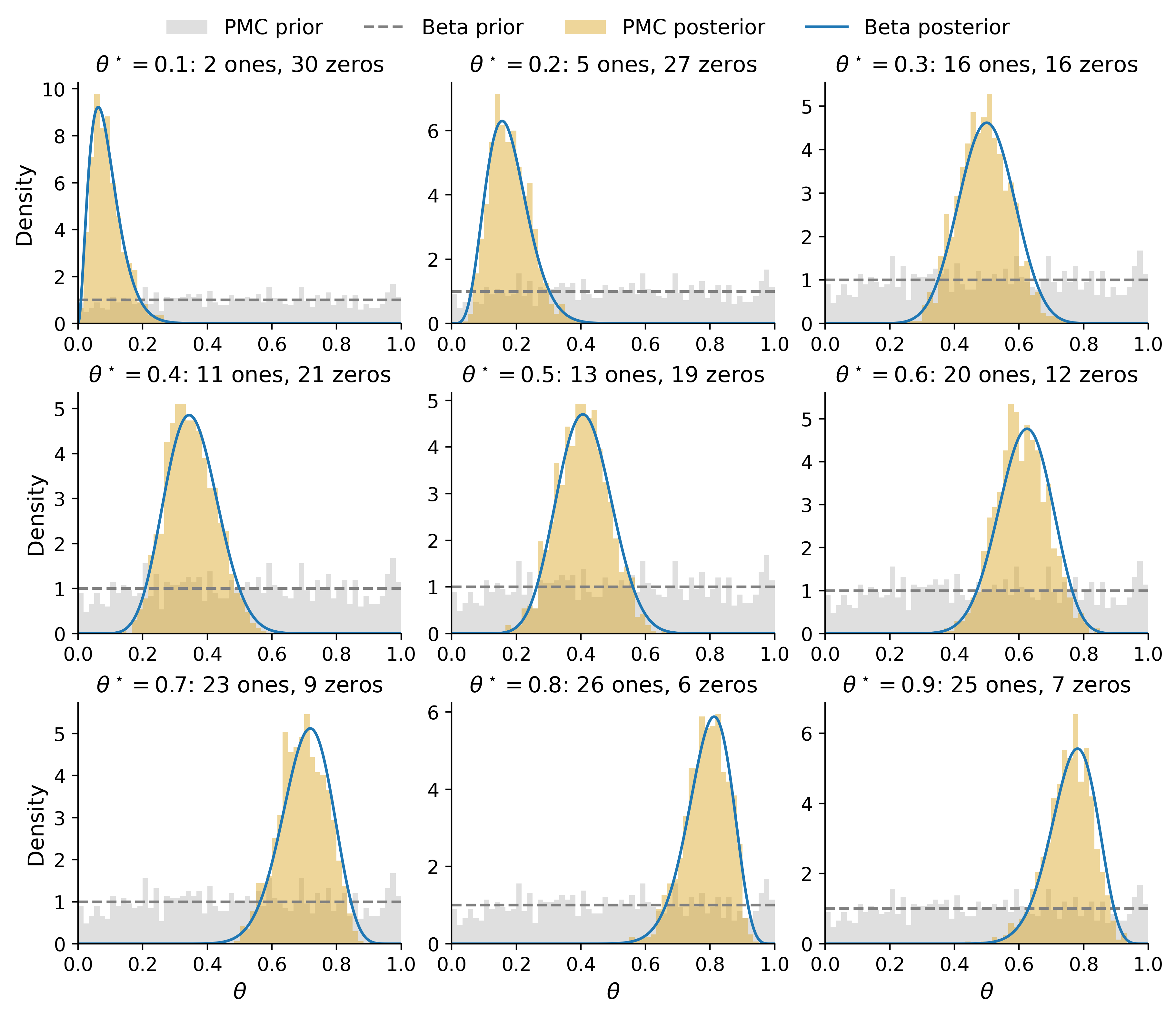}
  \caption{\textbf{Beta-Bernoulli: PMC prior and posterior across prompts.}
    PMC-recovered prior and posterior over $\theta$ versus the analytic
    $\operatorname{Beta}(1,1)$ prior and conjugate posterior.}
  \label{fig:grid-beta-bernoulli}
\end{figure}

\noindent\textbf{PMC.}
For a prompt $y_{1:n}$, let $S_n=\sum_{i=1}^n y_i$. The conjugate posterior is
\[
  \theta\mid y_{1:n} \sim \operatorname{Beta}(1+S_n,1+n-S_n).
\]
Rollouts extend the prompt by $\rolllen=450$ predictive steps, and every panel uses $\pmcsamples=1000$ rollouts (the empty-prompt prior is shared across panels). Figure~\ref{fig:pmc-schematic}(b) is the $\theta^\star=0.2$ panel of the grid in Figure~\ref{fig:grid-beta-bernoulli}, shown on its own for the main text.
Figure~\ref{fig:grid-beta-bernoulli} compares the PMC-recovered prior and
posterior over $\theta$ to the analytic prior and posterior. Each panel
conditions on a different prompt $y_{1:n}$ of length $32$. Reading the panels
left-to-right, top-to-bottom, the underlying ground truth probability takes
values $\theta^\star = 0.1, 0.2, \dots, 0.9$.

\FloatBarrier

\clearpage
\section{Linear regression}
\label{app:linear-training}
\label{app:linear-baselines}

\noindent\textbf{Data-generating setup.}
We study in-context linear regression with input dimension $d=8$ and observation model
\[
  y_i = w^\top x_i + \varepsilon_i,\qquad x_i\sim \mathcal N(0,I_d),\qquad \varepsilon_i\sim \mathcal N(0,\sigma^2),
\]
with $\sigma^2 = 0.25$. The population prior is $\Pi_\infty = \mathcal
  N(0,I_d)$. At task diversity $M$, training-sequence generation follows
Section~\ref{sec:framework}: sample $M$ regression vectors from $\Pi_\infty$ to
form the training prior, then generate training sequences from that discrete
prior. Evaluation prompts are written as
\[
  c=((x_1,y_1),\dots,(x_\ell,y_\ell)),
\]
with the latent regression vector drawn either from the training prior
$\Pi_{\taskdiv}$ (in-distribution) or from $\Pi_\infty$ (out-of-distribution).

\noindent\textbf{Baselines.}
Let the support points of the empirical training prior at diversity $M$ be $w^{(1)},\dots,w^{(M)}$. For a prompt $c=((x_1,y_1),\dots,(x_\ell,y_\ell))$, write $X=(x_1,\dots,x_\ell)^\top\in\mathbb R^{\ell\times d}$ and $\mathbf y=(y_1,\dots,y_\ell)^\top$. The memorizing baseline uses the empirical training prior, so its posterior is discrete:
\[
  \Pi^{\mem}(w^{(m)}\mid c)
  = \frac{\exp\!\left(-\frac{1}{2\sigma^2}\lVert \mathbf y - X w^{(m)}\rVert^2\right)}{\sum_{j=1}^{M}\exp\!\left(-\frac{1}{2\sigma^2}\lVert \mathbf y - X w^{(j)}\rVert^2\right)}.
\]
Its Bayes predictor at a query covariate $x_*$ is the prediction under the posterior mean, called dMMSE (discrete minimum mean squared error) by \citet{Raventos2023}:
\[
  \hat w_{\mathrm{dMMSE}} = \sum_{m=1}^{M} \Pi^{\mem}(w^{(m)}\mid c)\, w^{(m)},
  \qquad \hat y_{\mathrm{dMMSE}} = x_*^\top \hat w_{\mathrm{dMMSE}}.
\]

The Gaussian generalizing baseline uses
\[
  w \sim \mathcal N(0,I_d),
\]
which yields the conjugate posterior
\[
  w\mid X,\mathbf y \sim \mathcal N(\mu_n,\Sigma_n),
  \qquad \Sigma_n=(X^\top X/\sigma^2 + I_d)^{-1},
  \qquad \mu_n = \Sigma_n X^\top \mathbf y / \sigma^2,
\]
and the ridge-regression Bayes predictor
\[
  \hat w_{\mathrm{ridge}} = (X^\top X + \sigma^2 I_d)^{-1}X^\top \mathbf y.
\]

\noindent\textbf{Transformer architecture.}
The transformer architectures for linear regression in \citet{Raventos2023}, \citet{Carroll2025} and \citet{NEURIPS2025_abb22b1e} are all decoder-only, sequential residual stream, autoregressive transformers. \citet{NEURIPS2025_abb22b1e} uses the GPT-NeoX architecture \citep{Black2022} from Huggingface, which implies the use of rotary positional encodings (RoPE) instead of learned absolute positional embeddings. Following \citet{Raventos2023}, we train transformers with learned absolute positional embeddings. We use the HookedTransformer in TransformerLens \citep{nanda2022transformerlens} as the base implementation for our transformers, and follow the configuration of \citet{Carroll2025}: $2$ layers with embedding dimension $d_\mathrm{model} = 512$, $4$ attention heads with dimension $d_{\mathrm{head}} = 128$, MLP hidden dimension $d_\mathrm{mlp} = 512$, and GELU activations. This comes to approximately $3.5$M parameters per model. We train on sequences of length $64$. As the covariate and response are modeled separately, this means we train on a context length of $128$.

The transformers in \citet{Raventos2023} and \citet{NEURIPS2025_abb22b1e}
utilize a point prediction head, where the transformer outputs a point
prediction based on the context, and is trained on autoregressive mean-squared
error loss. To perform PMC, we require our transformers to model a predictive
distribution. To do so, we follow the setup of \citet{Muller2024}: we discretize the output space into categorical buckets and train on
(autoregressive) negative log-likelihood. This allows us to sample from the
transformer's predictive distribution, and allows for the autoregressive
generation of sequences from the model needed for PMC. We discretize $y$ into
$256$ uniform-width bins covering $[-10,10]$, with half-normal tails at the two
edge bins to have an unbounded support. For prediction space metrics, we obtain
a point prediction by taking the expected value of the transformer's predictive
distribution. We use inverse transform sampling to sample from the
transformer's predictive distribution for PMC rollout.

\begin{figure}[H]
  \centering
  \resizebox{0.6\linewidth}{!}{\input{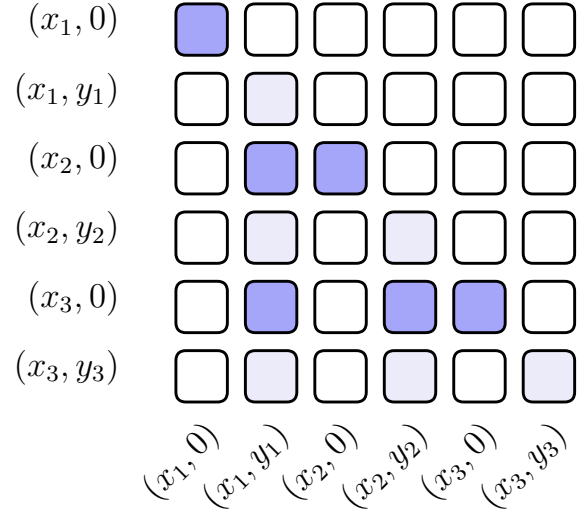}}
  \caption{\textbf{Autoregressive-PFN attention mask.} Even-position tokens embed the covariate $x_i$ and odd-position tokens embed the pair $(x_i, y_i)$. Each token attends to itself (darker cells) and to all earlier odd-position (full-pair) tokens (lighter cells).}
  \label{fig:ar-pfn-mask}
\end{figure}

We also alter the attention masking scheme. We construct the sequence
embeddings such that even-position (under $0$-indexing) tokens contain the
embedding for only the covariates $x_i$. The odd-position tokens contain the
sum of the embeddings of the covariate $x_i$ and the corresponding response
$y_i$. Instead of a traditional causal mask, we instead enforce tokens to
attend \emph{only} to themselves and previous \emph{odd-position} tokens in the
sequence. This attention masking scheme enforces the relationship that query
tokens only attend to themselves and previous complete pairs instead of
previous query tokens. Without enforcing this structure, the model would have
to learn the relationship between token positions and the corresponding
covariate-response pair. Figure~\ref{fig:ar-pfn-mask} illustrates the attention
mask. We note that the transformers are only trained on outputs from the
even-position (query) tokens.

We do not use a beginning-of-sequence (BOS) token in linear regression. The
discrete task family transformers (balls-and-urns and Markov chains) use a BOS token
for two reasons: it lets the BFT express a marginal predictive $P^\phi_0$ for the
first observation, and it supplies the starting context for PMC rollouts from
the prior $\widehat\Pi(\cdot\mid\varnothing)$. As our sequences are initialized
by an initial covariate $x_1$, this covariate can act in place of the BOS
token. To initialize a PMC rollout from the prior, we sample $x_1 \sim \mathcal
  N(0, I_d)$ and feed it in as the first token.

\noindent\textbf{Training details.}
We optimize with AdamW, warming the learning rate up linearly over $1{,}000$ steps to $10^{-4}$ and holding it constant thereafter. We train with weight decay $10^{-5}$, batch size $256$, gradient clipping at norm $1$, and $150{,}000$ training steps on a single NVIDIA RTX 5090 GPU with seed $42$. Each run takes approximately 90 minutes, and we parallelize the task-diversity sweep across an $8\times$RTX 5090 machine. Each training step draws a fresh batch of regression vectors from the empirical training prior $\Pi_{\taskdiv}$ together with fresh observations conditional on those vectors, and we minimize the cross-entropy of the bin label of each $y_i$. We save checkpoints on the union of $200$ logarithmically-spaced steps and a linear schedule every $500$ steps. We train one model per task diversity $M \in \{2^0, 2^1, \dots, 2^{16}\}$.

\noindent\textbf{PMC.}
To roll out a sequence from a prompt, we autoregressively sample a fresh covariate $x_i \sim \mathcal{N}(0, I_d)$, append it to the context, query the transformer for its categorical predictive over $y_i$, and draw $y_i$ from the transformer's predictive distribution through inverse-transform sampling. We append $(x_i, y_i)$ and iterate for $\rolllen$ steps. Prior rollouts have no observed prompt and initialize the sequence with a single $x_1 \sim \mathcal{N}(0, I_d)$. Posterior rollouts initialize the sequence with an observed prompt $(x_{1:\promptlen}, y_{1:\promptlen})$ drawn from the prompt source under evaluation.

\begin{figure}[H]
  \centering
  \begin{subfigure}[t]{\linewidth}
    \centering
    \includegraphics[width=0.9\linewidth]{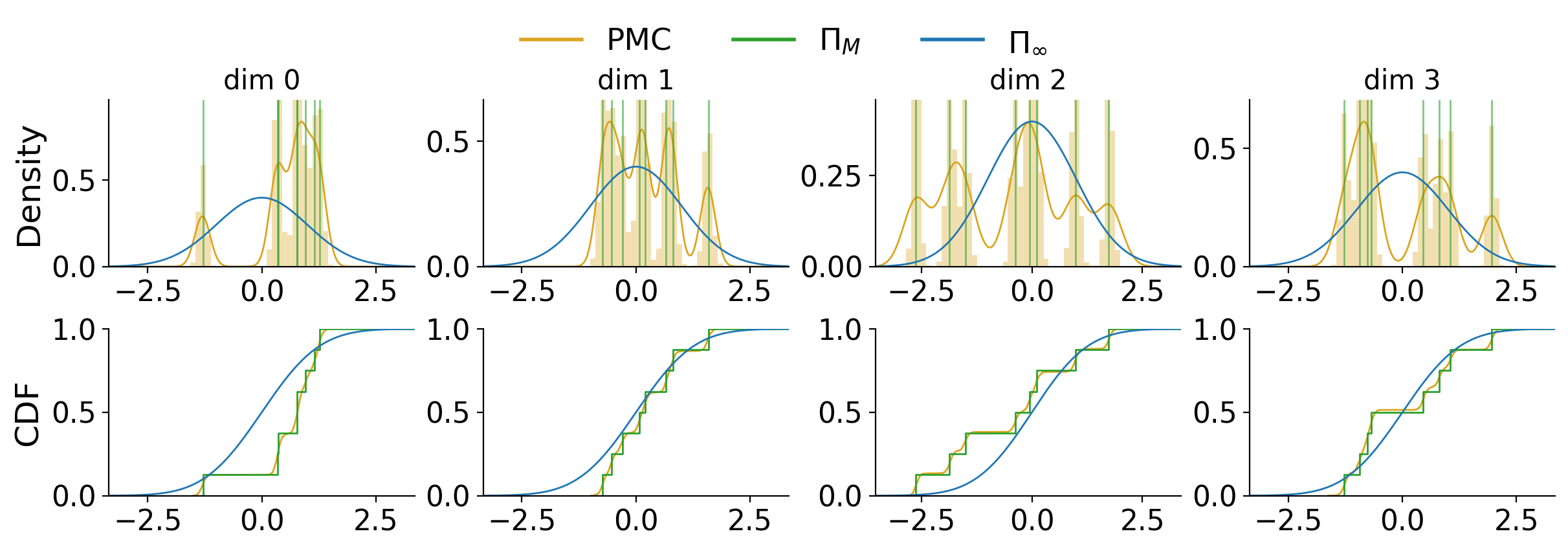}
    \caption{Prior samples.}
  \end{subfigure}
  \vspace{0.5em}
  \begin{subfigure}[t]{\linewidth}
    \centering
    \includegraphics[width=0.9\linewidth]{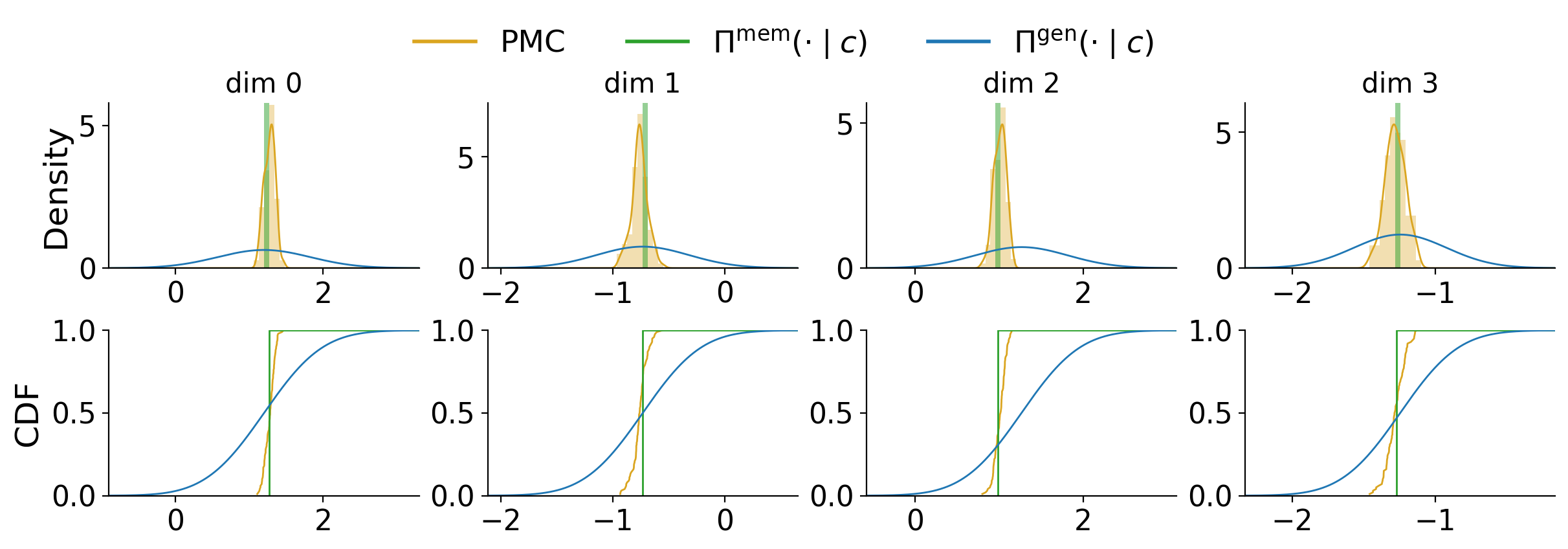}
    \caption{Posterior samples, conditioned on an in-distribution prompt of length $8$.}
  \end{subfigure}
  \vspace{0.5em}
  \begin{subfigure}[t]{\linewidth}
    \centering
    \includegraphics[width=0.9\linewidth]{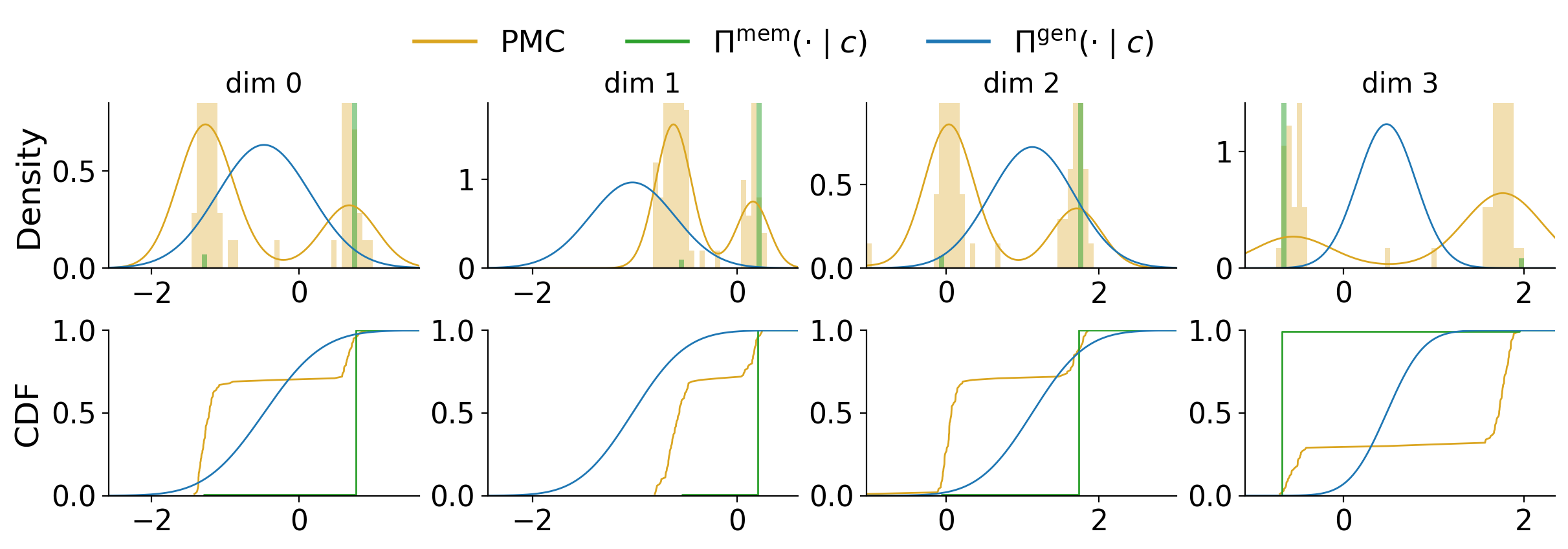}
    \caption{Posterior samples, conditioned on an out-of-distribution prompt of length $8$.}
  \end{subfigure}
  \caption{\textbf{Linear regression: PMC prior and posterior samples.} Marginal densities and CDFs over the first four dimensions of $w$ at task diversity $M=8$, against the training prior $\Pi_{\taskdiv}$, the population prior $\Pi_\infty$, and their corresponding posteriors.}
  \label{fig:linear-samples-app}
\end{figure}

\begin{figure}[H]
  \centering
  \begin{subfigure}[t]{\linewidth}
    \centering
    \includegraphics[width=0.95\linewidth]{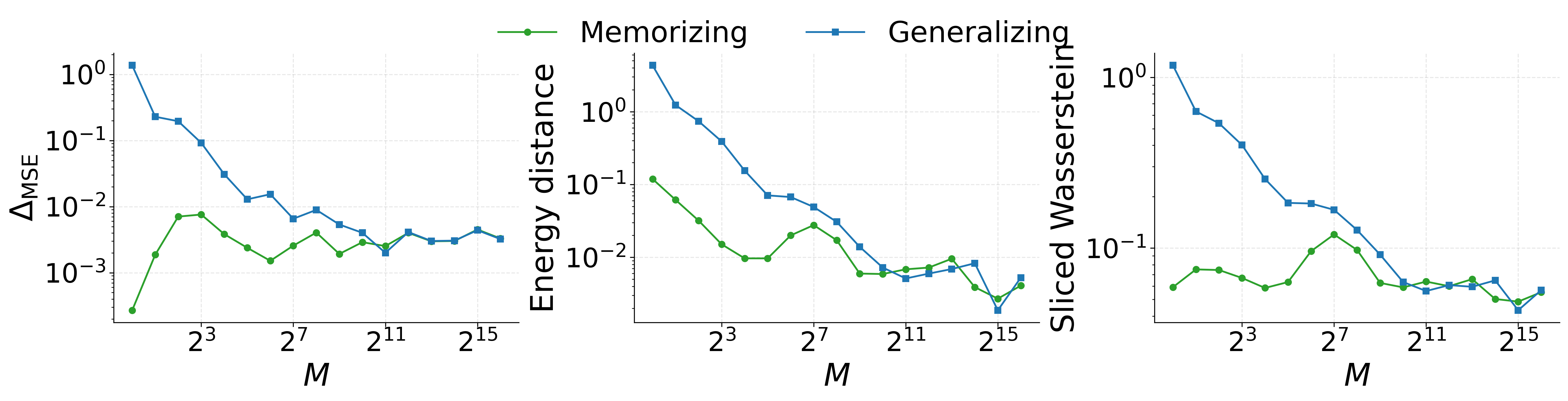}
    \caption{Prior space. Prediction-space $\Delta_{\mathrm{MSE}}$ (left),
      energy distance between PMC prior samples and $\Pi_{\taskdiv}$ or
      $\Pi_\infty$ (middle), and sliced Wasserstein distance to
      $\Pi_{\taskdiv}$ or $\Pi_\infty$ (right), as task diversity varies.}
  \end{subfigure}
  \vspace{0.5em}
  \begin{subfigure}[t]{\linewidth}
    \centering
    \includegraphics[width=0.95\linewidth]{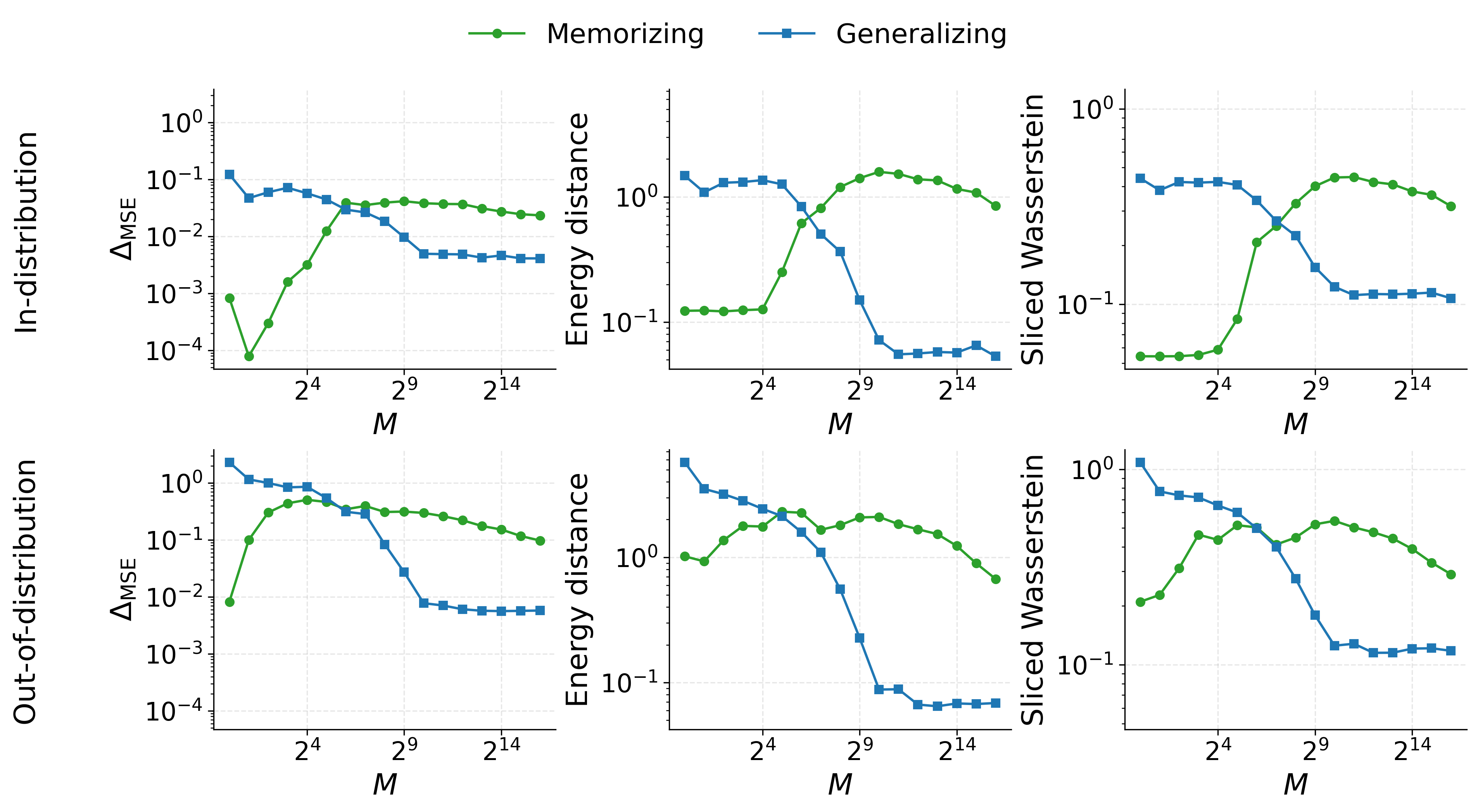}
    \caption{Posterior space. For in-distribution prompts (top) and
      out-of-distribution prompts (bottom), panels show prediction-space
      $\Delta_{\mathrm{MSE}}$ (left), energy distance to the memorizing or
      generalizing posterior (middle), and sliced Wasserstein distance to the
      memorizing or generalizing posterior (right), as task diversity varies.}
  \end{subfigure}
  \caption{\textbf{Linear regression: task diversity in the prior and posterior.} As task diversity grows, the PMC-recovered prior and posterior move from the memorizing baseline toward the generalizing baseline. The transition is sharper in the posterior.}
  \label{fig:linear-diversity-app}
\end{figure}

\begin{figure}[H]
  \centering
  \includegraphics[width=0.95\linewidth]{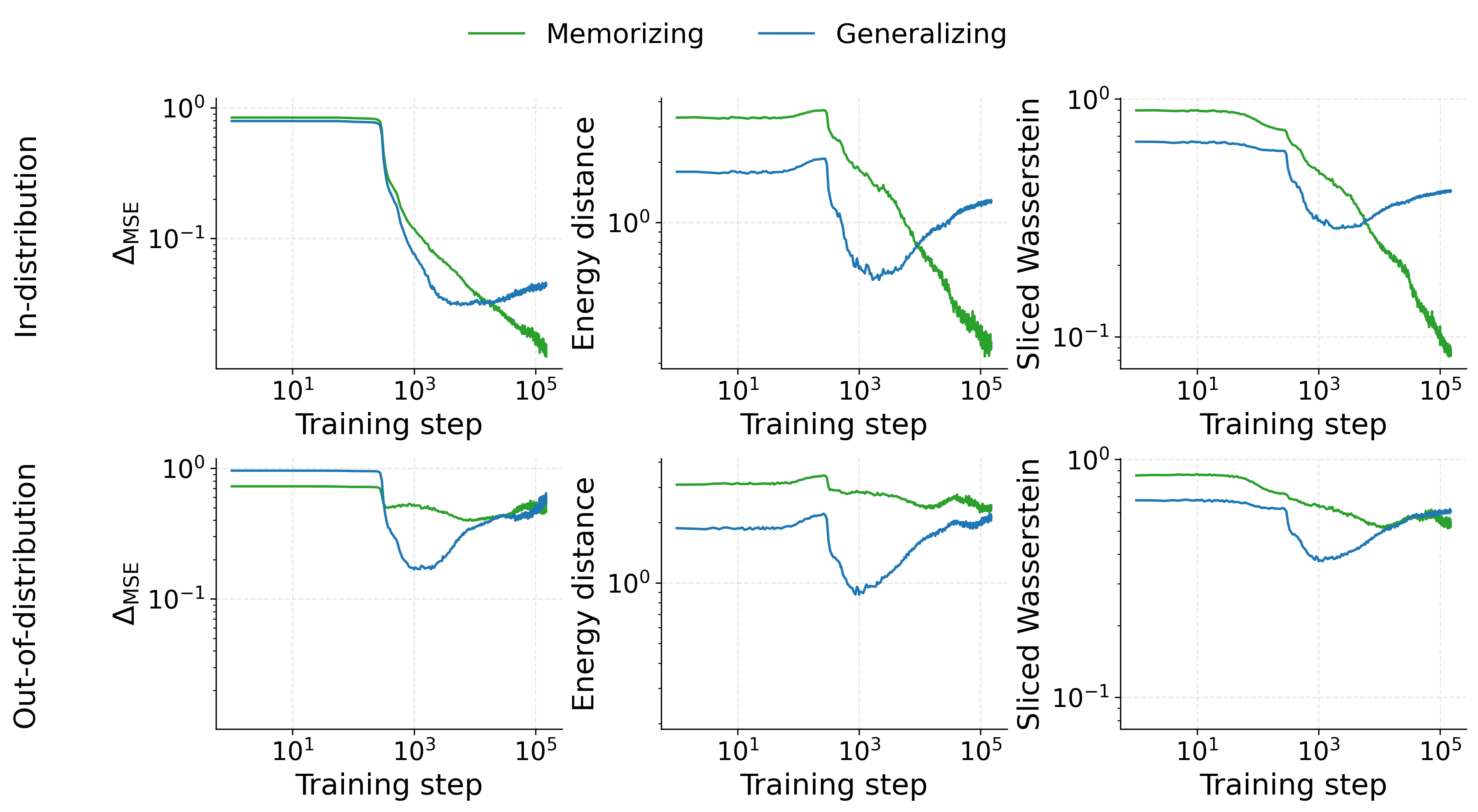}
  \caption{\textbf{Linear regression: transient generalization.} At intermediate
    task diversity $M=32$, the transformer first approximates the generalizing
    baseline before specializing to the memorizing baseline. For
    in-distribution prompts (top) and out-of-distribution prompts (bottom),
    columns show prediction-space $\Delta_{\text{MSE}}$ (left), energy distance
    to each baseline (middle), and sliced Wasserstein distance to each baseline
    (right).}
  \label{fig:linear-transient-app}
\end{figure}

For the task-diversity sweep (Figures~\ref{fig:linear-samples-app} and
\ref{fig:linear-diversity-app}) we set the PMC rollout length to $\rolllen=56$
predictive steps. The posterior is recovered from $\pmcsamples=100$ rollouts
per evaluation prompt. We use $50$ evaluation prompts at length $\promptlen=8$
from each prompt source. In-distribution prompts are obtained by sampling a
regression vector from the empirical training prior $\Pi_{\taskdiv}$ and
generating an $8$-pair sequence. Out-of-distribution prompts sample the
regression vector from the population prior $\Pi_\infty=\mathcal N(0,I_d)$
instead. The prior is recovered from $\pmcsamples=3000$ rollouts. The
transient-generalization experiment at $\taskdiv=32$
(Figure~\ref{fig:linear-transient-app}) uses the same rollout settings and the
same evaluation prompts.

\noindent\textbf{Prior in prediction space.} Figure~\ref{fig:linear-diversity-app}(a), leftmost panel, reports the $\Delta_{\mathrm{MSE}}$ for the transformer's prior. To compute this, we sample $4096$ query covariates $x \sim \mathcal{N}(0, I_d)$ and record the unconditioned prediction $\hat y(x)$ from the transformer. $\Delta_{\mathrm{MSE}}$ is the mean squared difference between $\hat y(x)$ and the prior predictor of each baseline at the same $x$. The generalizing predictor is $0$, and the memorizing predictor is $\tfrac{1}{M}\sum_{m=1}^{M}(w^{(m)})^\top x$.

\FloatBarrier

\clearpage
\section{Balls-and-urns}
\label{app:urn-baselines}

\noindent\textbf{Data-generating setup.}
The balls-and-urns family generalizes the Beta-Bernoulli setup of
Appendix~\ref{app:beta-bernoulli} from two outcomes to $|\mathcal Y|=12$ categorical
outcomes.
Let $p\in\Delta^{|\mathcal Y|-1}$ denote the urn proportions, with population prior
$\Pi_\infty = \operatorname{Dirichlet}(1,\dots,1)$.
At task diversity $M$, training-sequence generation again follows
Section~\ref{sec:framework}: sample $M$ urns from $\Pi_\infty$ to form the empirical
training prior, then generate training sequences from that discrete prior.
Evaluation prompts are written as $c=(y_1,\dots,y_\ell)$, with the latent urn drawn either from the empirical training prior $\Pi_{\taskdiv}$ (in-distribution) or from $\Pi_\infty$ (out-of-distribution).

\noindent\textbf{Baselines.} Let the support points of the empirical training prior at diversity $M$ be $p^{(1)},\dots,p^{(M)}$. For a prompt $c=(y_1,\dots,y_\ell)$, the memorizing baseline uses that empirical training prior, so the posterior is the discrete law
\[
  \Pi^{\mem}(p^{(m)}\mid c)
  =
  \frac{\prod_{i=1}^\ell p^{(m)}_{y_i}}{\sum_{j=1}^M \prod_{i=1}^\ell p^{(j)}_{y_i}}.
\]
The generalizing baseline uses the Dirichlet population prior, giving the
conjugate posterior $p\mid c \sim
  \operatorname{Dirichlet}(1+N_1(c),\dots,1+N_{|\mathcal Y|}(c))$, where
$N_a(c)=\sum_{i=1}^\ell \mathbf 1\{y_i=a\}$.

\noindent\textbf{Transformer architecture.}
We follow the architecture of \citet{NEURIPS2025_abb22b1e}, who train decoder-only autoregressive transformers on sequences of categorical observations. We use the HookedTransformer in TransformerLens \citep{nanda2022transformerlens} as the base implementation for our transformers, and train with learned absolute position embeddings. We use $2$ layers with embedding dimension $d_\mathrm{model} = 64$, a single attention head with dimension $d_{\mathrm{head}} = 64$, MLP hidden dimension $d_\mathrm{mlp} = 256$, and GELU activations. This comes to approximately $118$K parameters per model. We train with maximum context length $256$, consisting of the BOS token followed by $255$ categorical observations.

We use the standard left-to-right causal attention mask and train on
autoregressive cross-entropy loss. To model the unconditioned prior predictive
distribution, we prepend a beginning-of-sequence (BOS) token to each sequence,
which lets the BFT express a marginal predictive $P^\phi_0$ for the first
observation and supplies the starting context for PMC rollouts from the prior
$\widehat\Pi(\cdot\mid\varnothing)$. We do not include the BOS token in the
output vocabulary space. That is, we have the input vocabulary dimension
$d_{\mathrm{vocab\_in}} = 13$ and output vocabulary $d_{\mathrm{vocab\_out}} =
  12$.

\noindent\textbf{Training details.}
We optimize with AdamW at learning rate $5\times 10^{-4}$, weight decay $10^{-5}$, batch size $64$, gradient clipping at norm $1$, and $100{,}000$ training steps on a single NVIDIA RTX 5090 GPU with seed $42$. Each run takes approximately 85 minutes, and we parallelize the task-diversity sweep across an $8\times$RTX 5090 machine. We use no warmup or learning-rate decay. Each training step draws a fresh batch of urn proportions from the empirical training prior $\Pi_{\taskdiv}$ together with fresh balls conditional on those urns, and we minimize the cross-entropy of the categorical prediction at each position. We save checkpoints on the union of $200$ logarithmically-spaced steps and a linear schedule every $500$ steps. We train one model per task diversity $M \in \{2^0, 2^1, \dots, 2^{12}\}$.

\begin{figure}[H]
  \centering
  \begin{subfigure}[t]{\linewidth}
    \centering
    \includegraphics[width=0.9\linewidth]{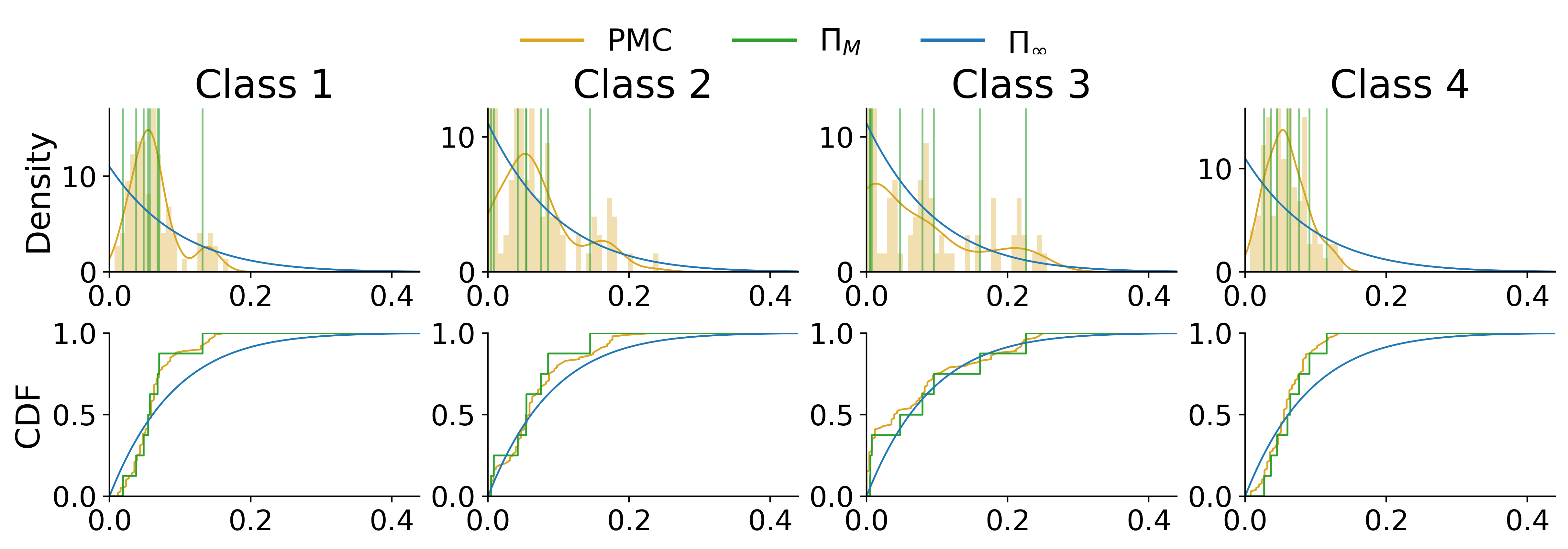}
    \caption{Prior samples.}
    \label{fig:bau-samples-prior}
  \end{subfigure}
  \vspace{0.5em}
  \begin{subfigure}[t]{\linewidth}
    \centering
    \includegraphics[width=0.9\linewidth]{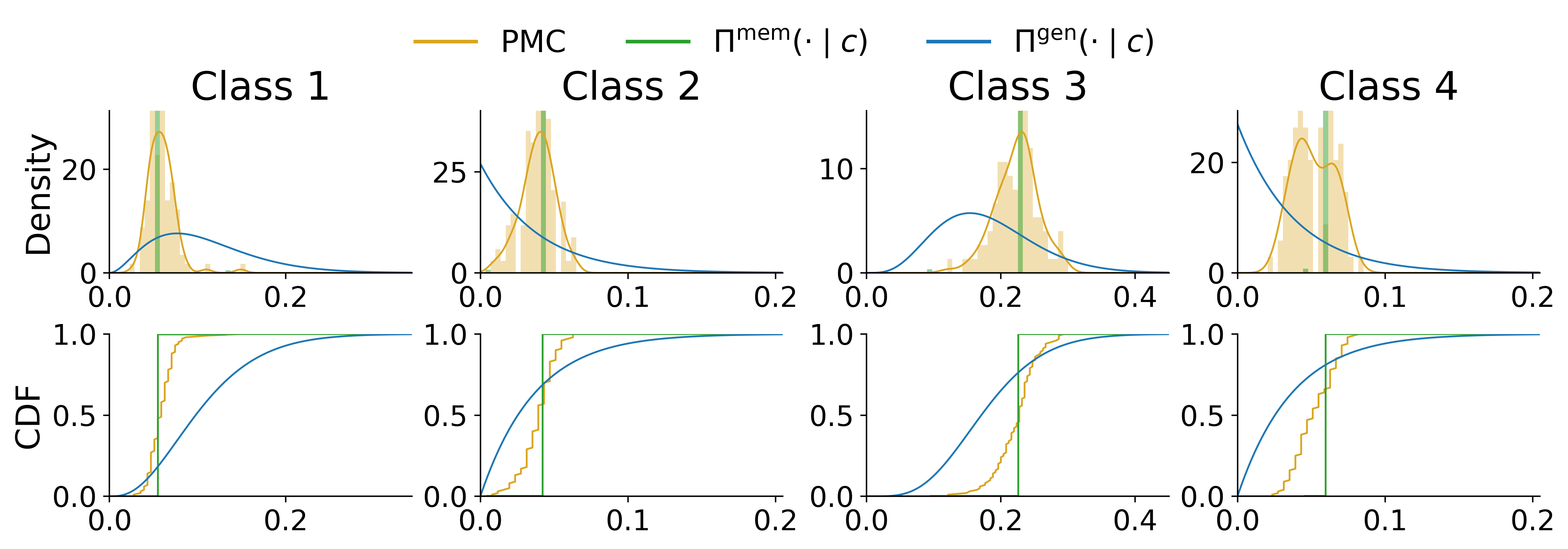}
    \caption{Posterior samples, conditioned on an in-distribution prompt of length $16$.}
    \label{fig:bau-samples-mem}
  \end{subfigure}
  \vspace{0.5em}
  \begin{subfigure}[t]{\linewidth}
    \centering
    \includegraphics[width=0.9\linewidth]{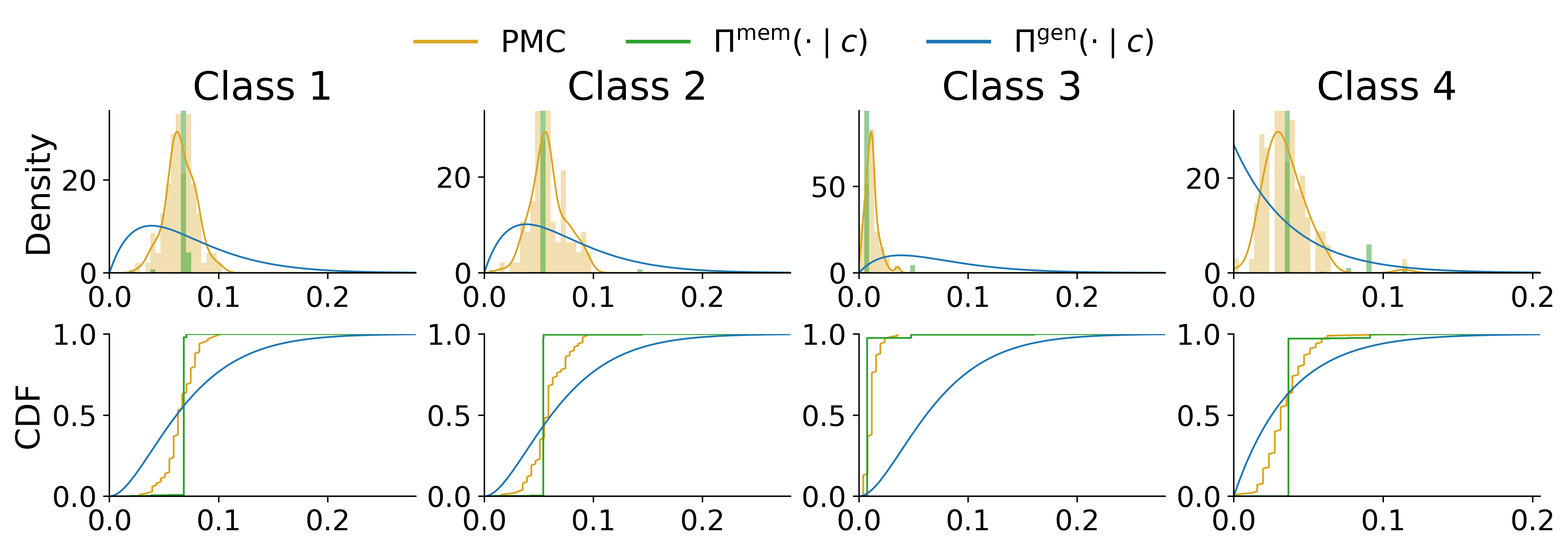}
    \caption{Posterior samples, conditioned on an out-of-distribution prompt of length $16$.}
    \label{fig:bau-samples-gen}
  \end{subfigure}
  \caption{\textbf{Balls-and-urns: PMC prior and posterior samples.} Marginal densities and CDFs over the first four classes of $p$ at task diversity $M=8$, against the training prior $\Pi_{\taskdiv}$, the population prior $\Pi_\infty$, and their corresponding posteriors.}
  \label{fig:bau-samples}
\end{figure}

\begin{figure}[H]
  \centering
  \begin{subfigure}[t]{\linewidth}
    \centering
    \includegraphics[width=0.95\linewidth]{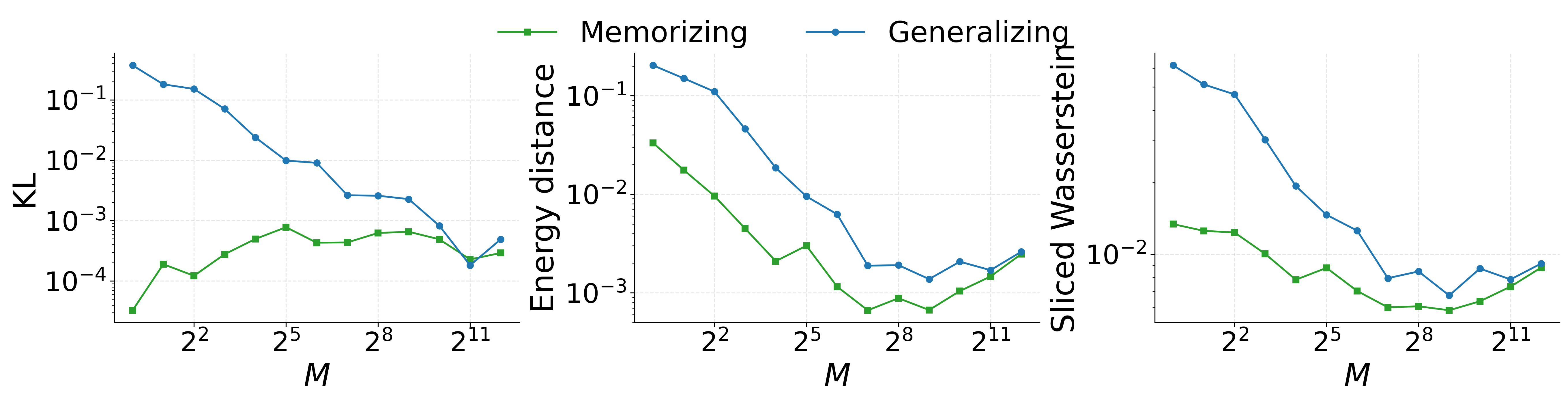}
    \caption{Prior space.}
  \end{subfigure}
  \vspace{0.5em}
  \begin{subfigure}[t]{\linewidth}
    \centering
    \includegraphics[width=\linewidth]{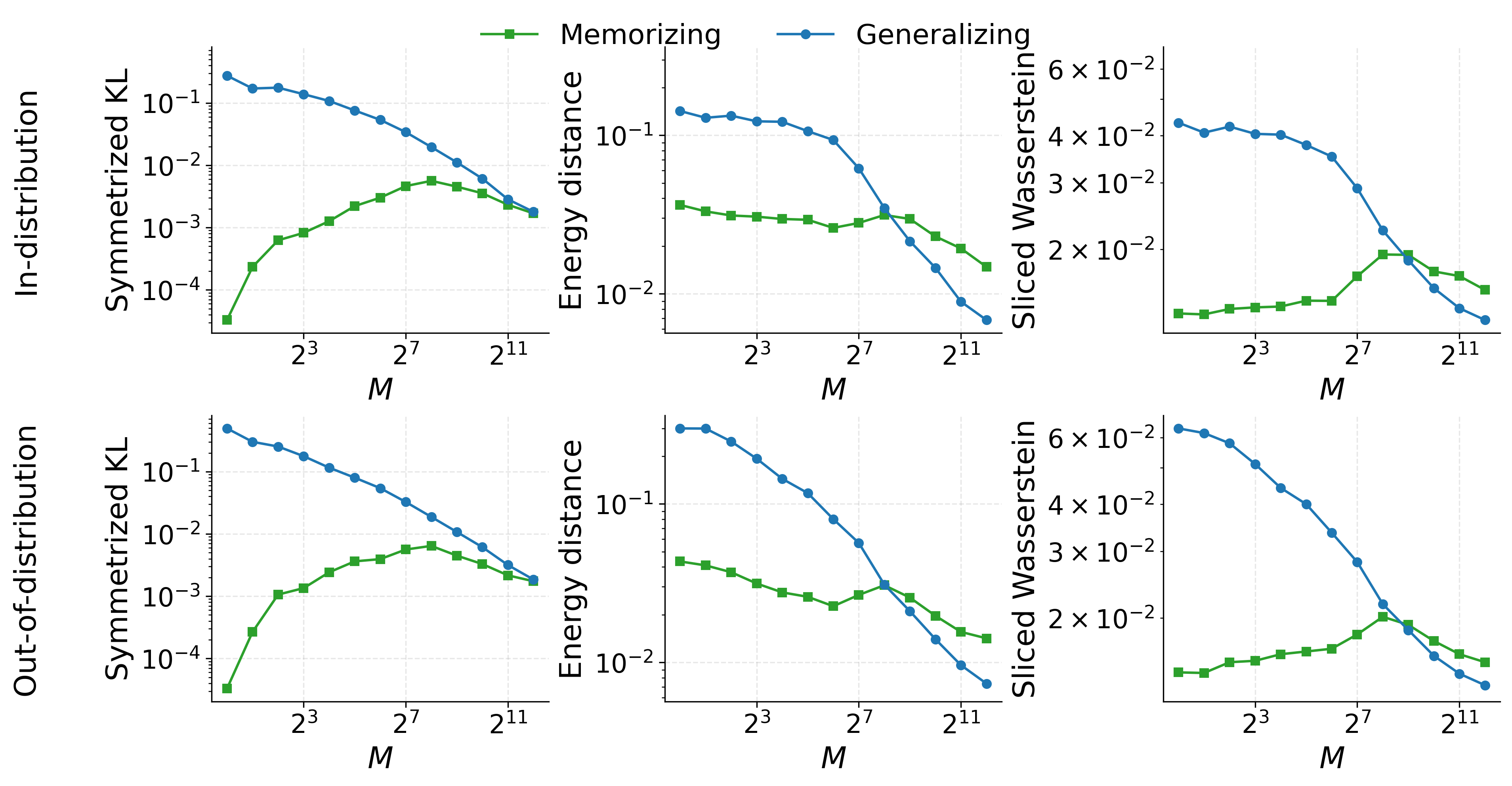}
    \caption{Posterior space.}
  \end{subfigure}
  \caption{\textbf{Balls-and-urns: task diversity in the prior and posterior, all three metrics.} Prediction-space symmetrized KL gap (left), energy distance (middle), and sliced Wasserstein distance (right) to each baseline, as task diversity varies. The main-text view is Figure~\ref{fig:bau-main}(a).}
  \label{fig:bau-diversity}
\end{figure}

\noindent\textbf{PMC.}
For the task-diversity sweep (Figures~\ref{fig:bau-samples}, \ref{fig:bau-main}(a), and \ref{fig:bau-diversity}) we roll out $\rolllen=255-\promptlen$ predictive steps, so every completed sequence contains $255$ observations. The posterior is recovered from $\pmcsamples=100$ rollouts per evaluation prompt. We use $128$ evaluation prompts at length $\promptlen=16$ from each prompt source. In-distribution prompts are obtained by sampling an urn $p$ from the empirical training prior $\Pi_{\taskdiv}$ and then drawing a length-$16$ sequence $y_{1:16}\overset{\mathrm{iid}}{\sim}\operatorname{Categorical}(p)$. Out-of-distribution prompts sample $p$ from the population prior $\Pi_\infty=\operatorname{Dirichlet}(1,\dots,1)$ instead. The prior is recovered from $\pmcsamples=1024$ rollouts of length $255$ from the BOS-only prompt. After each rollout we apply the empirical-frequency estimator $\hat p_a = (\promptlen+\rolllen)^{-1}\sum_{i=1}^{\promptlen+\rolllen}\mathbf 1\{y_i=a\}$ to obtain a PMC sample of $\hat p$. The transient-generalization experiment at $\taskdiv=32$ (Figure~\ref{fig:bau-main}(b)) uses the same rollout settings and the same evaluation prompts, and evaluates every fifth saved checkpoint from the training run to reduce analysis cost.

Figure~\ref{fig:bau-main}(a) shows the task-diversity transition from
memorizing to generalizing in latent space but not in prediction space. The
prediction-space metric is computed at prompt length $\promptlen=16$, much
shorter than the $255$ sequence length in training. In early positions in the
sequence, the transformer behaves closer to the memorizing predictor. The
latent-space metrics, evaluated on the same prompts, separate the
two baselines.

\noindent\textbf{Prior in prediction space.} Figure~\ref{fig:bau-diversity}(a), leftmost panel, reports the symmetrized KL divergence for the transformer's prior. To compute this, we record the unconditioned next-token predictive $\hat p$ from the transformer on the BOS token alone. The symmetrized KL divergence is between $\hat p$ and the prior predictive of each baseline over the first observation. The generalizing predictive is uniform on $\mathcal Y$, and the memorizing predictive is $\tfrac{1}{M}\sum_{m=1}^{M} p^{(m)}$.

\begin{figure}[H]
  \centering
  \includegraphics[width=\linewidth]{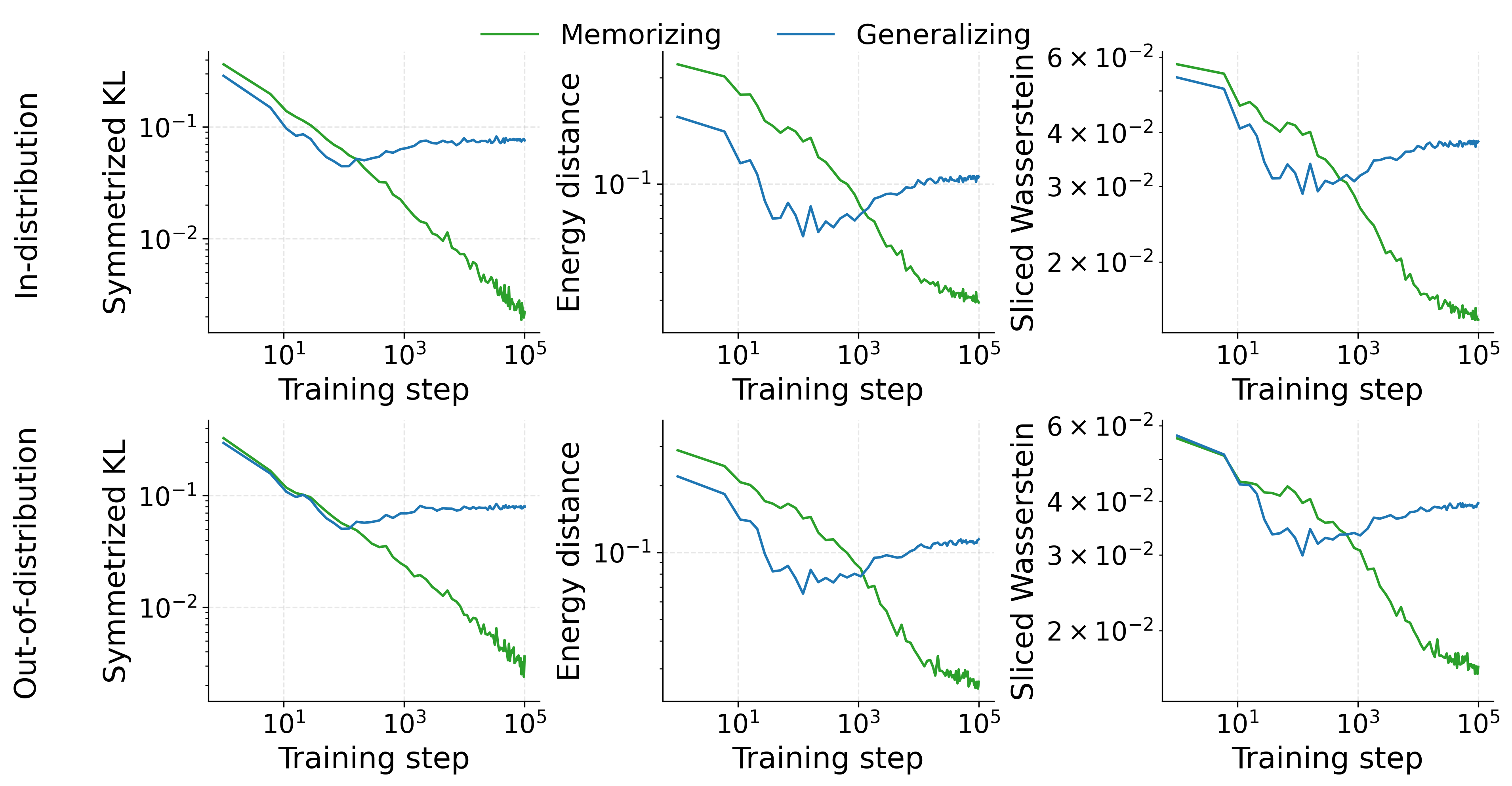}
  \caption{\textbf{Balls-and-urns: transient generalization, all three metrics.} At intermediate task diversity $M=32$. The main-text view is Figure~\ref{fig:bau-main}(b).}
  \label{fig:bau-transient}
\end{figure}

\FloatBarrier

\clearpage
\section{Markov chains}
\label{app:markov-baselines}

\noindent\textbf{Data-generating setup.} Let $Q$ be a transition matrix on the state space $\mathcal Y$. The population prior $\Pi_\infty$ draws each row of $Q$ independently from $\operatorname{Dirichlet}(1,\dots,1)$. At task diversity $M$, training-sequence generation follows Section~\ref{sec:framework}: sample $M$ kernels from $\Pi_\infty$ to form the empirical training prior, then generate training sequences from that discrete prior. Each sequence drawn from $Q$ is initialized with $y_1$ from the stationary distribution of $Q$. For the closed-form baselines we drop the $P(y_1\mid Q)$ factor from the likelihood, leaving the $\ell-1$ transitions $y_{i-1}\to y_i$ as the contributing terms. This ensures the posteriors are conjugate, and matches the convention of Section~\ref{sec:1markov-exch}, where $Y_1$ is taken as given. The omission is not innocuous at our prompt length: the sequences are generated with $y_1$ drawn from the stationary distribution of $Q$, so the exact posterior under the generative process carries a stationary-probability factor that the baselines omit, one of $\ell=8$ likelihood contributions. This is one candidate contribution to the looser match reported below. Evaluation prompts are written as $c=(y_1,\dots,y_\ell)$, with the latent kernel drawn either from the empirical training prior $\Pi_{\taskdiv}$ (in-distribution) or from $\Pi_\infty$ (out-of-distribution).

\noindent\textbf{Baselines.} The generalizing baseline uses $\Pi_\infty$ as the prior. Its posterior is
\[
  \Pi^{\gen}(Q\mid c)
  \propto
  \prod_{i=2}^\ell Q(y_i\mid y_{i-1})\,\Pi_\infty(Q).
\]
Let the support points of the empirical training prior at diversity $M$ be
$Q^{(1)},\dots,Q^{(M)}$. For a prompt $c=(y_1,\dots,y_\ell)$, the memorizing
baseline uses the empirical training prior. The posterior is the discrete law
\[
  p(Q=Q^{(m)}\mid c)
  =
  \frac{\prod_{i=2}^\ell Q^{(m)}(y_i\mid y_{i-1})}{\sum_{j=1}^M \prod_{i=2}^\ell Q^{(j)}(y_i\mid y_{i-1})}.
\]
The construction mirrors the memorizing and generalizing baselines used for
balls-and-urns and linear regression.

\noindent\textbf{Transformer architecture.}
For the Markov chains family, we use the same architecture as in \citet{Park2025}, namely a decoder-only causal transformer. We implement the model directly in PyTorch. We use $2$ pre-norm decoder blocks, multi-head causal self-attention with $4$ heads, model hidden dimension $d_\mathrm{model} = 64$, MLP hidden dimension $d_\mathrm{mlp} = 256$, and ReLU activations. Positional information is encoded with rotary embeddings (RoPE) at base $10^4$. Each training sequence consists of $1024$ sampled states; the model input is the BOS token followed by the first $1023$ states (an input window of $1024$ tokens), with all $1024$ states as prediction targets. Unlike the linear regression models, which predict a continuous output distribution, the Markov chains model operates on discrete state tokens with $|\mathcal Y|=10$ states. The input vocabulary has size $|\mathcal Y|+1=11$ to accommodate a BOS token, and the linear output head produces a matching $11$ logits. At inference we discard the BOS logit so the model only predicts states in $\mathcal Y$. Training cross-entropy is computed over all $11$ logits, with targets always in $\mathcal Y$.

\noindent\textbf{Training details.}
We optimize with AdamW at constant learning rate $6\times 10^{-4}$, weight decay $0.01$, batch size $128$, and $100{,}000$ training steps on a single NVIDIA RTX 5090 GPU with seed $42$. Each run takes approximately $150$ minutes, and we parallelize the task-diversity sweep across an $8\times$RTX 5090 machine. We use no warmup, no learning-rate decay, and no gradient clipping. Each training step draws a fresh batch of transition matrices from the empirical training prior $\Pi_{\taskdiv}$ together with fresh sequences conditional on those kernels, and we minimize the cross-entropy of the categorical prediction at each position. We save checkpoints every $200$ training steps. We train one model per task diversity $\taskdiv \in \{2^2, 2^3, \dots, 2^{11}\}$. Two evaluations run during training. The first, a training monitor in the style of \citet{Park2025} that does not enter any reported figure, computes a stationary-weighted KL between the transformer's implied transition matrix and the data-generating kernel every $200$ steps: each row $\hat Q(\cdot\mid a)$ is read off the model's next-token predictive after a length-$400$ context ending in state $a$, and the divergence is averaged over $10$ evaluation chains drawn either from the training pool (in-distribution) or afresh from $\Pi_\infty$ (out-of-distribution). The second computes the prediction-space $\Delta_{\mathrm{sKL}}$ of Appendix~\ref{app:experiments} against the memorizing and generalizing baselines on $16$ full-length sequences per evaluation point; the batch is kept small because the memorizing baseline's likelihood computation scales with $\taskdiv$.

\noindent\textbf{PMC.}
For the task-diversity sweep and the transient-generalization experiment, we set the PMC rollout length to $\rolllen=400$ predictive steps. The posterior is recovered from $\pmcsamples=128$ rollouts per evaluation prompt. We use $16$ evaluation prompts at length $\promptlen=8$ from each prompt source. In-distribution prompts are obtained by sampling a transition matrix from the training pool and generating an $8$-state sequence from it. Out-of-distribution prompts are obtained by sampling a fresh transition matrix whose rows are drawn independently from $\operatorname{Dirichlet}(1,\dots,1)$ and then generating a sequence of length $8$ from the drawn matrix. The prior is recovered from $\pmcsamples=1024$ rollouts with the BOS-only prompt and rollout length $\rolllen$. After each rollout, we Laplace-smooth the empirical transition counts of the full state sequence. For prior PMC this is the $\rolllen$ generated tokens (excluding the BOS token). For posterior PMC it is the $\promptlen$ prompt tokens followed by the $\rolllen$ generated tokens. For each pair $(a,b)\in\mathcal Y\times\mathcal Y$,
\[
  \hat Q(b\mid a) = \frac{1 + N_{a,b}}{|\mathcal Y| + N_{a,\cdot}},
\]
where $N_{a,b}$ counts $a\to b$ transitions across this full sequence and
$N_{a,\cdot} = \sum_{b'\in\mathcal Y} N_{a,b'}$ counts visits to state $a$.

\begin{figure}[H]
  \centering
  \begin{subfigure}[t]{\linewidth}
    \centering
    \includegraphics[width=0.45\linewidth]{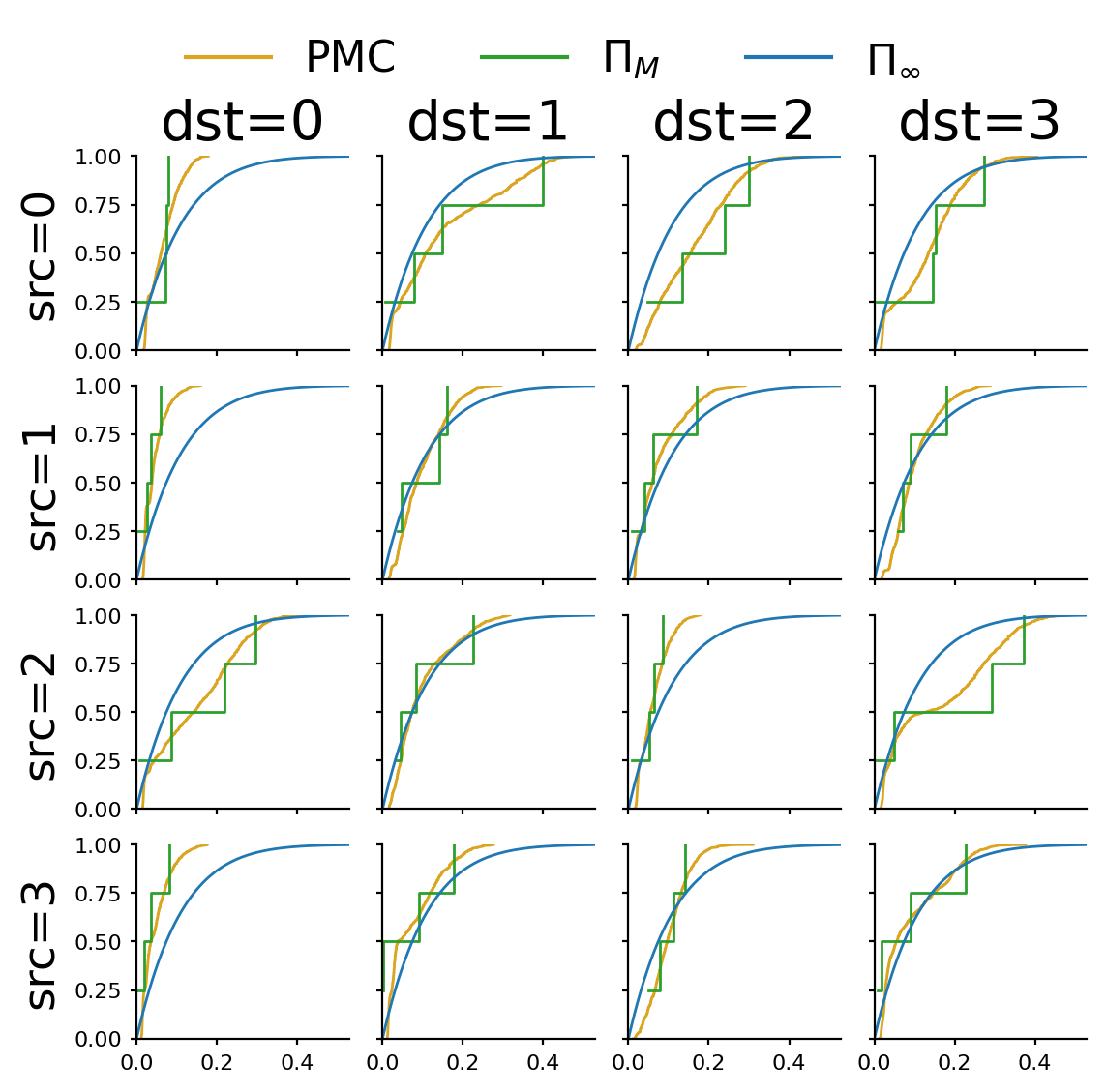}
    \caption{Prior samples.}
    \label{fig:markov-samples-prior}
  \end{subfigure}
  \vspace{0.5em}
  \begin{subfigure}[t]{\linewidth}
    \centering
    \includegraphics[width=0.45\linewidth]{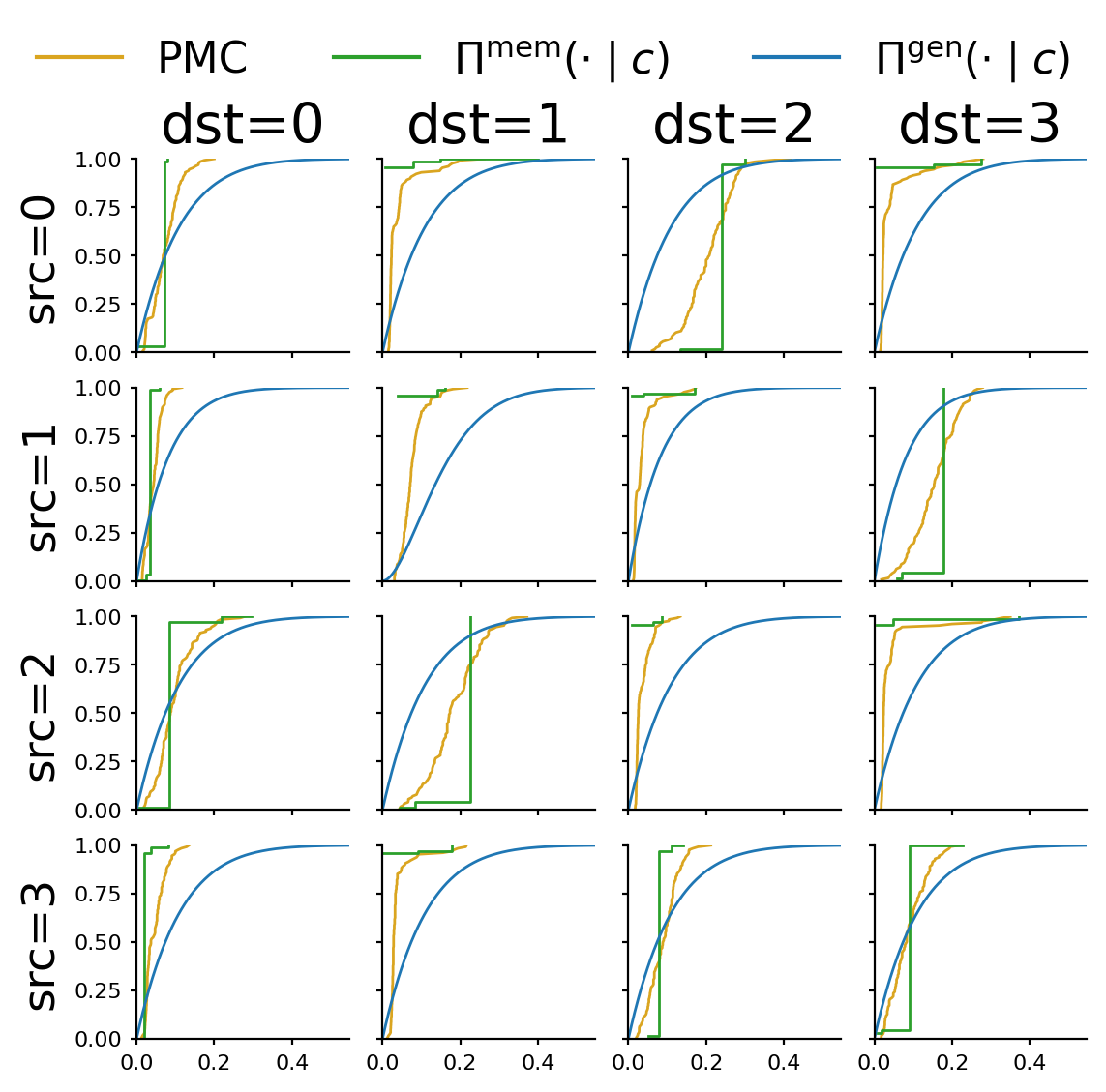}
    \caption{Posterior samples, conditioned on an in-distribution prompt of length 8.}
    \label{fig:markov-samples-mem}
  \end{subfigure}
  \vspace{0.5em}
  \begin{subfigure}[t]{\linewidth}
    \centering
    \includegraphics[width=0.45\linewidth]{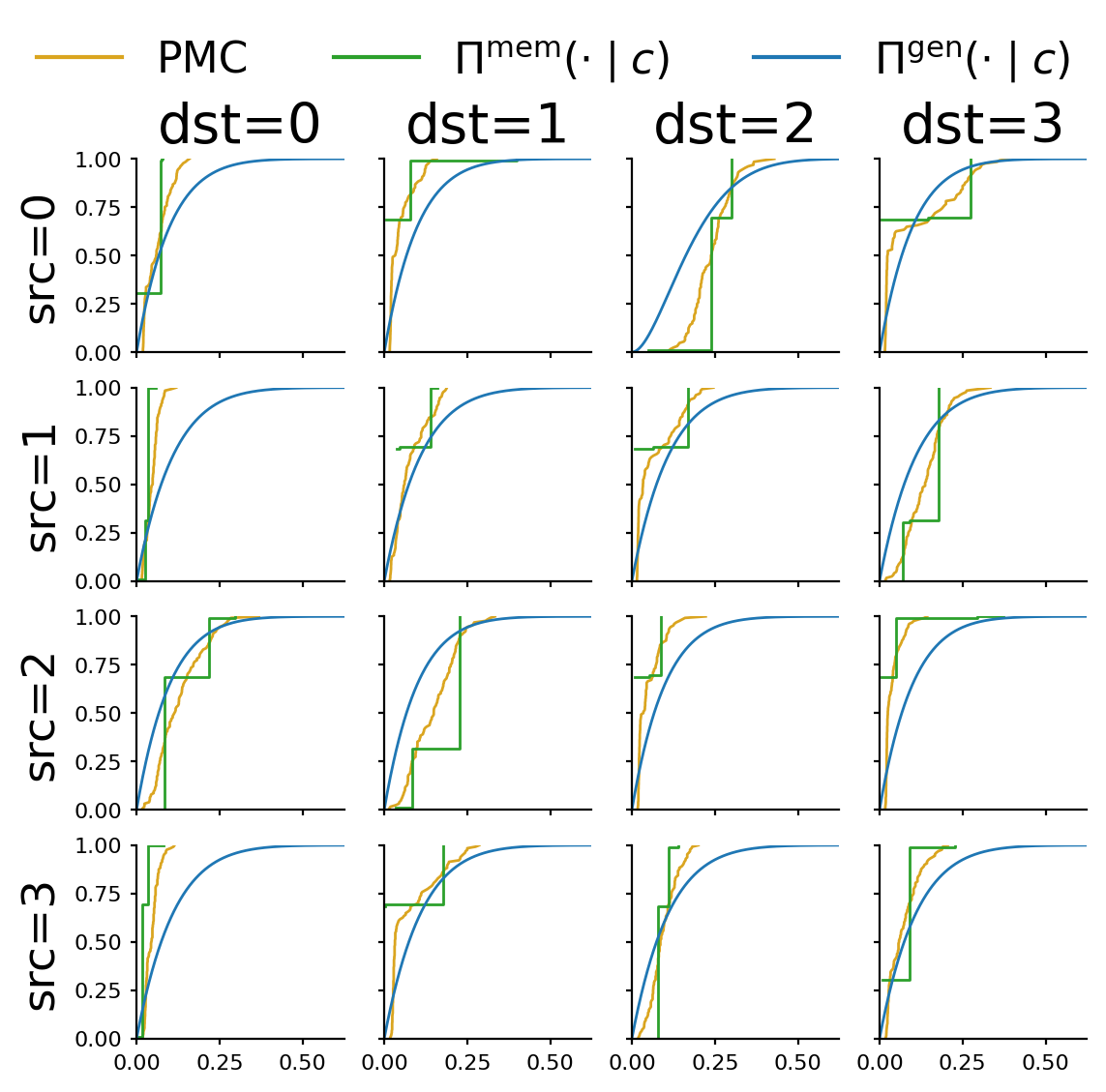}
    \caption{Posterior samples, conditioned on an out-of-distribution prompt of length 8.}
    \label{fig:markov-samples-gen}
  \end{subfigure}
  \caption{\textbf{Markov chains: PMC prior and posterior samples.} Each figure is a $4\times 4$ matrix of panels mirroring the top-left block of the transition matrix: the panel in row $\mathrm{src}=a$ and column $\mathrm{dst}=b$ shows the marginal CDF of the transition probability $Q_{a,b}$ from state $a$ to state $b$, at task diversity $M = 4$, against the training prior $\Pi_{\taskdiv}$, the population prior $\Pi_\infty$, and their corresponding posteriors.}
  \label{fig:markov-samples}
\end{figure}

PMC samples track the reference distributions less tightly in the Markov
chains family than in linear regression or balls-and-urns
(Figure~\ref{fig:markov-samples}). We propose two reasons for this. First, the
transition matrix $Q$ has $|\mathcal{Y}|(|\mathcal{Y}|-1) = 90$ free
parameters, considerably more than the latents in the other two families, so
PMC has a higher-dimensional posterior to recover. Second, relative to its latent task the Markov
chains BFT is the most heavily loaded model in our study: it must track a
$90$-parameter latent with roughly the same parameter budget ($\approx 10^5$)
that the balls-and-urns BFT devotes to an $11$-parameter latent.

\begin{figure}[H]
  \centering
  \begin{subfigure}[t]{\linewidth}
    \centering
    \includegraphics[width=0.95\linewidth]{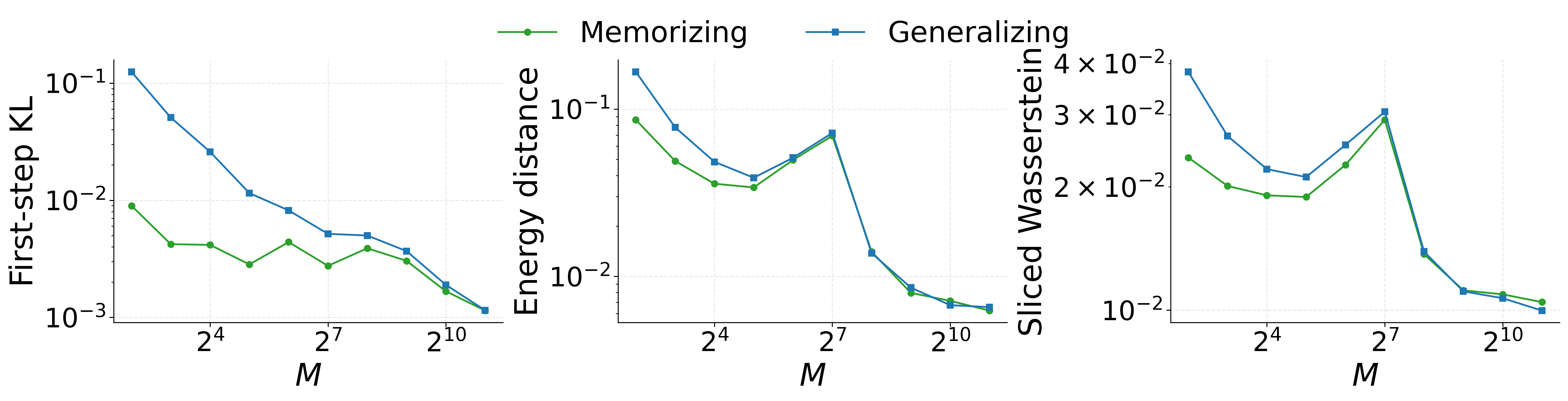}
    \caption{Prior space.}
  \end{subfigure}
  \vspace{0.5em}
  \begin{subfigure}[t]{\linewidth}
    \centering
    \includegraphics[width=0.95\linewidth]{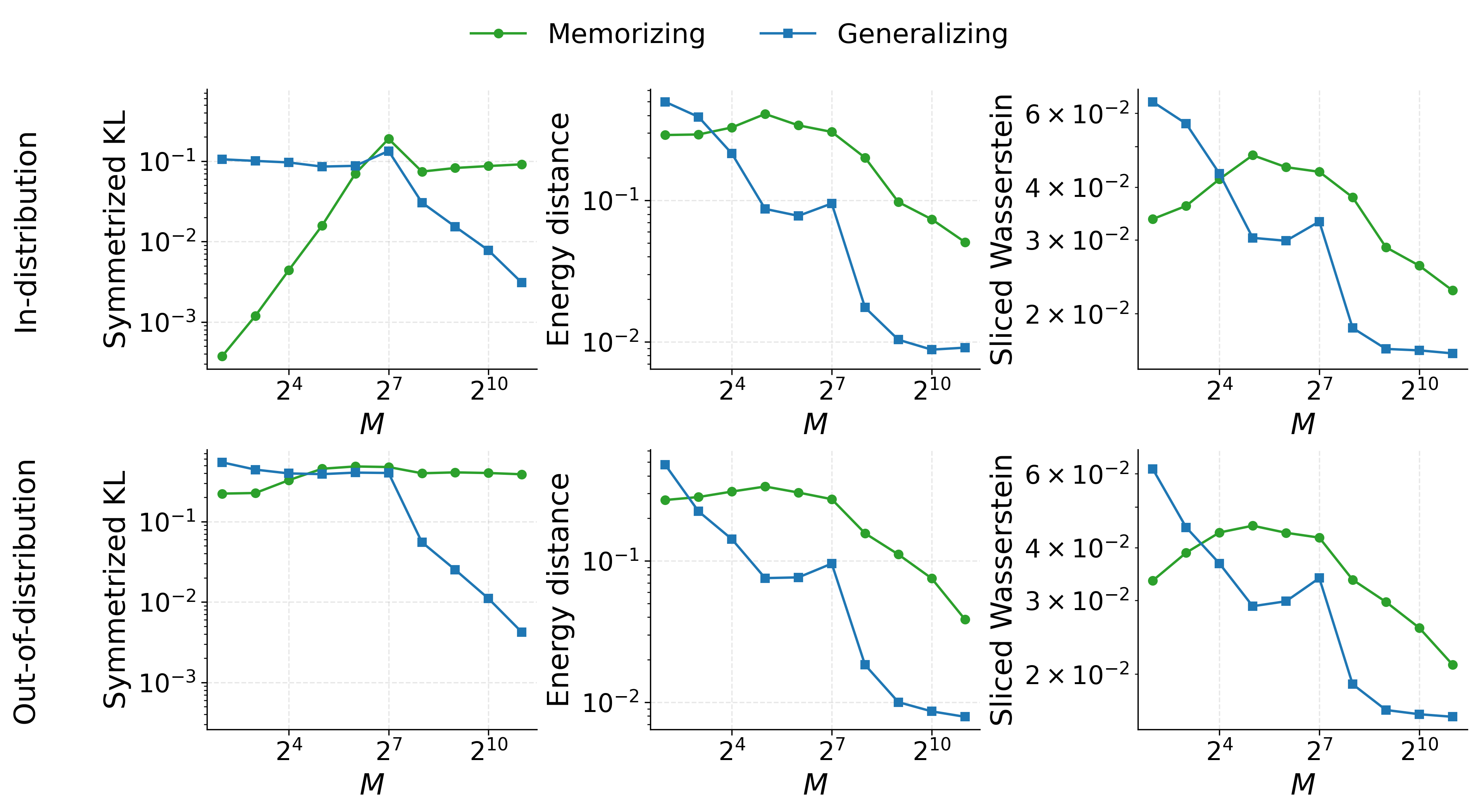}
    \caption{Posterior space.}
  \end{subfigure}
  \caption{\textbf{Markov chains: task diversity in the prior and posterior, all three metrics.} Prediction-space symmetrized KL (left), energy distance (middle), and sliced Wasserstein distance (right) to each baseline, as task diversity varies. The main-text view is Figure~\ref{fig:markov-main}(a).}
  \label{fig:markov-diversity}
\end{figure}

\noindent\textbf{Prior in prediction space.} Figure~\ref{fig:markov-diversity}(a), leftmost panel, reports the mean symmetrized KL for the transformer's prior. To compute this, for each state $a\in\mathcal Y$, we take the transformer's next-token predictive at the prompt $(\mathrm{BOS}, a)$, with no further context, as the first-step transition row $\hat Q(\cdot\mid a)$. The mean symmetrized KL is the average over $a\in\mathcal Y$ of the symmetrized KL between $\hat Q(\cdot\mid a)$ and the corresponding row of the prior predictive transition matrix of each baseline. The generalizing kernel is uniform over the rows, and row $a$ of the memorizing kernel is $\tfrac{1}{M}\sum_{m=1}^{M} Q^{(m)}(\cdot\mid a)$.

\begin{figure}[H]
  \centering
  \includegraphics[width=0.95\linewidth]{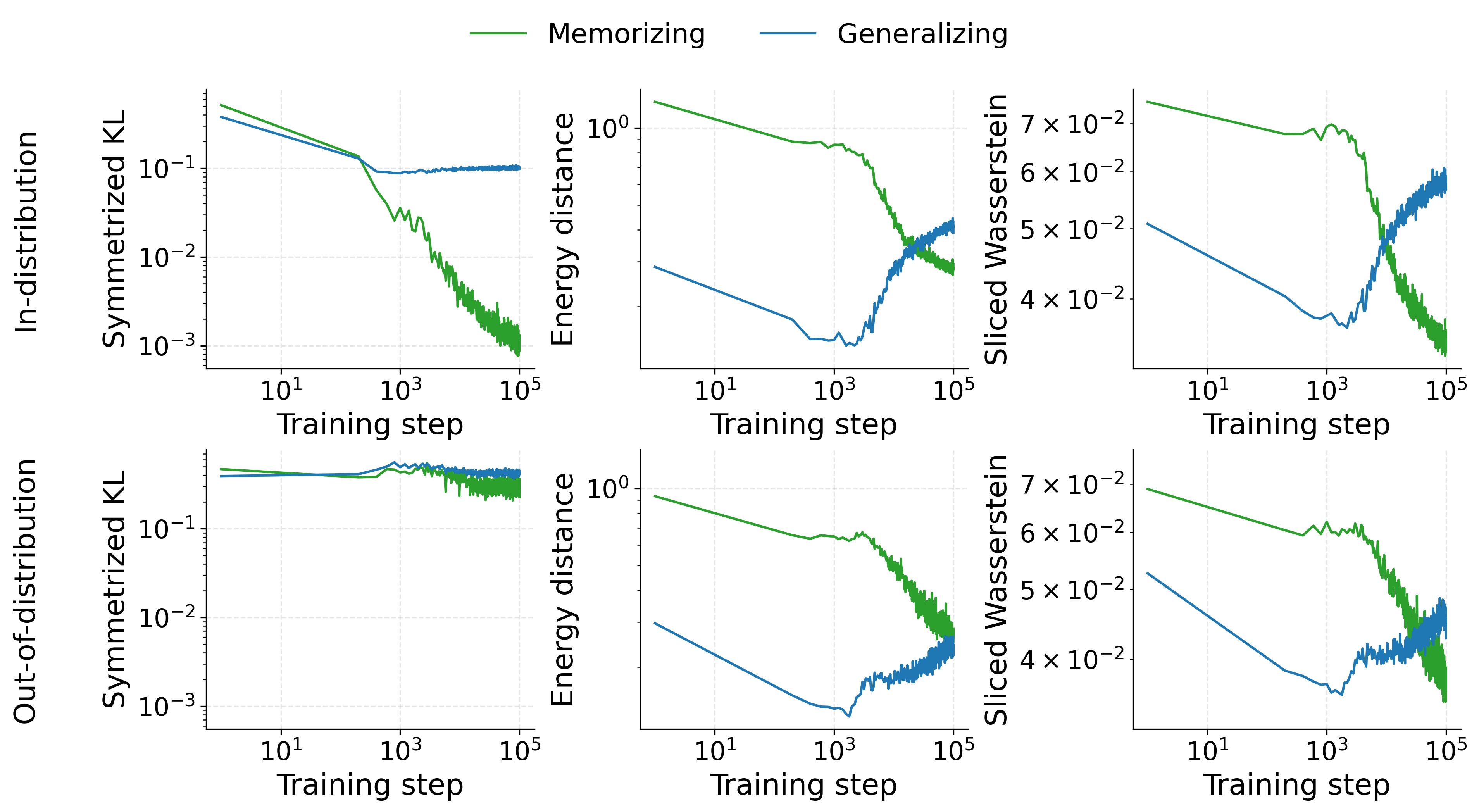}
  \caption{\textbf{Markov chains: transient generalization, all three metrics.} At task diversity $M=8$. The main-text view is Figure~\ref{fig:markov-main}(b).}
  \label{fig:markov-transient}
\end{figure}

\FloatBarrier

\clearpage

\section{Marginal densities and CDFs from PMC}
\label{app:stitched-marginals}

The complete set of PMC-recovered marginal densities and CDFs, spanning all
three task families across every task diversity, is provided as
supplementary material (the file \texttt{icl2\_supplementary.pdf}) accompanying the
code at \url{https://github.com/afiq-aswadi/bft-pmc}. Representative examples appear
in the main text and in the family appendices above.

\end{document}